\documentclass[10pt,twocolumn,letterpaper]{article}

\usepackage{iccv}
\usepackage{times}
\usepackage{epsfig}
\usepackage{graphicx}
\usepackage{amsmath}
\usepackage{amssymb}

\usepackage{booktabs}
\usepackage{caption}
\usepackage{subcaption}
\usepackage{wrapfig}
\usepackage{float}
\usepackage{rotating}
\usepackage{etoolbox,lipsum}
\usepackage{multirow}

\usepackage[pagebackref=true,breaklinks=true,letterpaper=true,colorlinks,bookmarks=false]{hyperref}

\iccvfinalcopy % *** Uncomment this line for the final submission

\def\iccvPaperID{7699} % *** Enter the ICCV Paper ID here

\ificcvfinal\fi

\begin{document}

%%%%%%%%% TITLE
\title{Scaling-up Disentanglement for Image Translation}

\author{Aviv Gabbay  \qquad Yedid Hoshen \\
School of Computer Science and Engineering\\
The Hebrew University of Jerusalem, Israel\\~\\
Project webpage: \textcolor{blue}{http://www.vision.huji.ac.il/overlord} \\
}

\newcommand{\todo}[1]{\textcolor{red}{[#1]}}

\maketitle
% Remove page # from the first page of camera-ready.
\ificcvfinal\thispagestyle{empty}\fi

\begin{abstract}
Image translation methods typically aim to manipulate a set of labeled attributes (given as supervision at training time e.g. domain label) while leaving the unlabeled attributes intact. Current methods achieve either: (i) disentanglement, which exhibits low visual fidelity and can only be satisfied where the attributes are perfectly uncorrelated. (ii) visually-plausible translations, which are clearly not disentangled. In this work, we propose OverLORD, a single framework for disentangling labeled and unlabeled attributes as well as synthesizing high-fidelity images, which is composed of two stages; (i) \textbf{Disentanglement}: Learning disentangled representations with latent optimization. Differently from previous approaches, we do not rely on adversarial training or any architectural biases. (ii) \textbf{Synthesis}: Training feed-forward encoders for inferring the learned attributes and tuning the generator in an adversarial manner to increase the perceptual quality. When the labeled and unlabeled attributes are correlated, we model an additional representation that accounts for the correlated attributes and improves disentanglement. We highlight that our flexible framework covers multiple settings as disentangling labeled attributes, pose and appearance, localized concepts, and shape and texture. We present significantly better disentanglement with higher translation quality and greater output diversity than state-of-the-art methods.
\end{abstract}

\begin{figure*}[t]
\begin{center}
\begin{subfigure}{0.49\textwidth}
\begin{tabular}{c@{\hskip1pt}c@{\hskip1pt}c}
\textbf{Younger} & Original & \textbf{Older} \\
\includegraphics[width=0.32\linewidth]{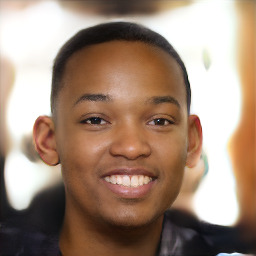} &
\includegraphics[width=0.32\linewidth]{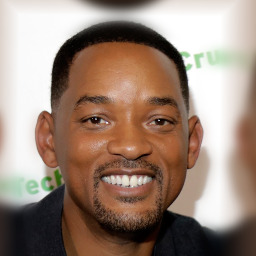} &
\includegraphics[width=0.32\linewidth]{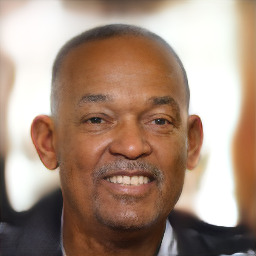}
\end{tabular}
\caption{Labeled Attribute (Age)}
\vspace{0.5em}
\end{subfigure}
\begin{subfigure}{0.49\textwidth}
\begin{tabular}{c@{\hskip1pt}c@{\hskip1pt}c}
Pose & Appearance & \textbf{Translation} \\
\includegraphics[width=0.32\linewidth]{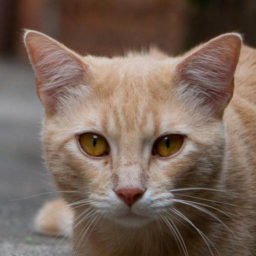} &
\includegraphics[width=0.32\linewidth]{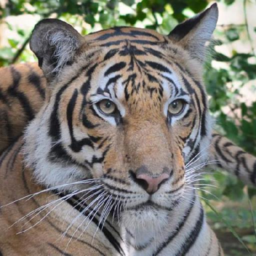} &
\includegraphics[width=0.32\linewidth]{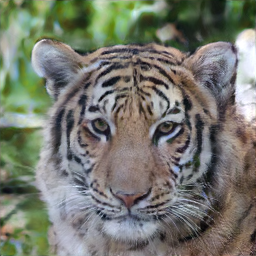} 
\end{tabular}
\caption{Pose and Appearance}
\vspace{0.5em}
\end{subfigure}
\begin{subfigure}{0.49\textwidth}
\begin{tabular}{c@{\hskip1pt}c@{\hskip1pt}c}
Identity & Gender+Hair & \textbf{Translation} \\
\includegraphics[width=0.32\linewidth]{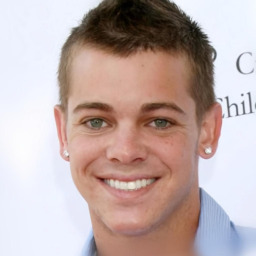} &
\includegraphics[width=0.32\linewidth]{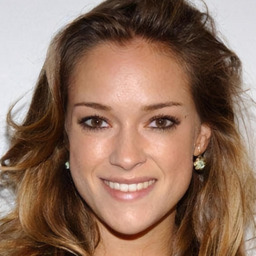} &
\includegraphics[width=0.32\linewidth]{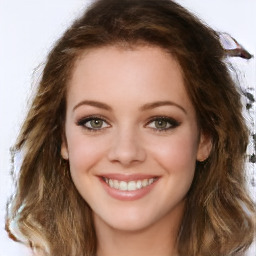}
\end{tabular}
\caption{Labeled Attribute (Gender) With Localized Correlation (Hair)}
\end{subfigure}
\begin{subfigure}{0.49\textwidth}
\begin{tabular}{c@{\hskip1pt}c@{\hskip1pt}c}
Shape & Texture & \textbf{Translation} \\
\includegraphics[width=0.32\linewidth]{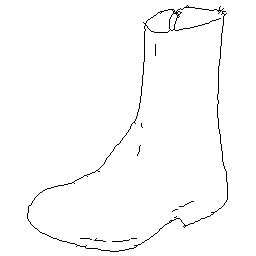} &
\includegraphics[width=0.32\linewidth]{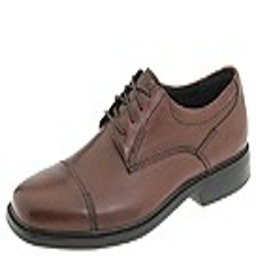} &
\includegraphics[width=0.32\linewidth]{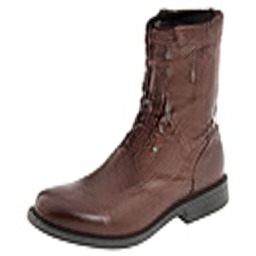}
\end{tabular}
\caption{Shape and Texture}
\end{subfigure}
\end{center}
\caption{A summary of the different attribute forms covered by our disentanglement framework.}
\label{fig:teaser}
\vspace{-0.75em}
\end{figure*}

\section{Introduction}
\begin{figure*}
\begin{center}

\begin{tabular}{c@{\hskip0pt}c@{\hskip3pt}c@{\hskip0pt}c@{\hskip0pt}c}
Pose & Appearance & FUNIT \cite{liu2019few} & StarGAN-v2 \cite{choi2019stargan} & \textbf{Ours} \\
\includegraphics[width=0.18\linewidth]{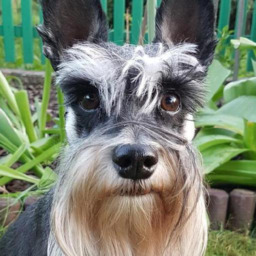} &
\includegraphics[width=0.18\linewidth]{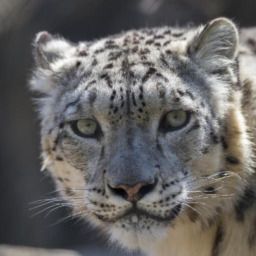} &
\includegraphics[width=0.18\linewidth]{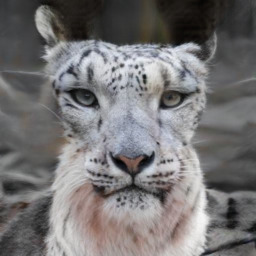} & \includegraphics[width=0.18\linewidth]{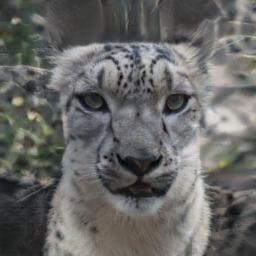} &
\includegraphics[width=0.18\linewidth]{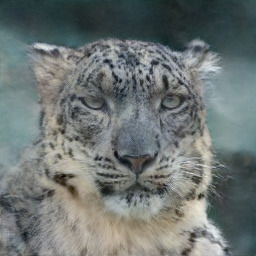} \\
\includegraphics[width=0.18\linewidth]{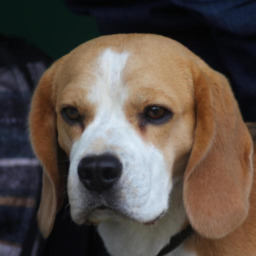} &
\includegraphics[width=0.18\linewidth]{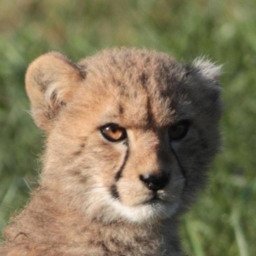} &
\includegraphics[width=0.18\linewidth]{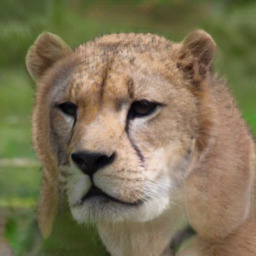} & \includegraphics[width=0.18\linewidth]{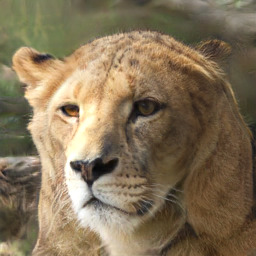} &
\includegraphics[width=0.18\linewidth]{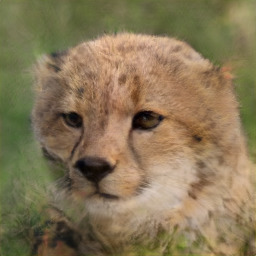} \\
\end{tabular}
\end{center}
\caption{Entanglement in image translation models on AFHQ. Domain label (cat, dog or wild) and its correlated attributes (e.g. breed) are guided by the appearance image, while the uncorrelated attributes (e.g. head pose) of the source image should be preserved. Current approaches and their architectural biases tightly preserve the original structure and generate unreliable facial shapes which are unique to the source domain. We disentangle the pose and capture the target breed faithfully.}
\label{fig:afhq}
\vspace{-0.5em}
\end{figure*}

Learning disentangled representations for different factors of variation in a set of observations is a fundamental problem in machine learning. Such representations can facilitate generalization to downstream generative and discriminative tasks such as novel image synthesis \cite{zhu2018visual} and person re-identification \cite{eom2019learning}, as well as improving interpretability \cite{hsu2017unsupervised}, reasoning \cite{van2019disentangled} and fairness \cite{creager2019flexibly}. A popular task which benefits from disentanglement is image translation, in which the goal is to translate a given input image from a source domain (e.g. cats) to an analogous image in a target domain (e.g. dogs). Although this task is generally poorly specified, it is often satisfied under the assumption that images in different domains share common attributes (e.g. head pose) which can remain unchanged during translation. In this paper, we divide the set of all attributes that define a target image precisely into two subsets; (i) \textit{labeled} attributes: the attributes that are supervised at training time e.g. whether the image belongs to the ``cats'' or ``dogs'' domain; (ii) \textit{unlabeled} attributes: all the remaining attributes that we do not have supervision for e.g. breed of the animal, head pose, background etc. While several methods (e.g. LORD \cite{gabbay2019demystifying}, StarGAN \cite{choi2018stargan}, Fader Networks \cite{lample2017fader}) have been proposed for disentangling labeled and unlabeled attributes, we explain and show why they can not deal with cases where the labeled and unlabeled attributes are correlated. For example, when translating cats to dogs, the unlabeled breed (which specifies fur texture or facial shape) is highly correlated with the domain label (cat or dog), and can not be translated between species. This dependency motivates the specification of more fine-grained attributes that we wish the translated image to have. Several methods as MUNIT \cite{huang2018multimodal}, FUNIT \cite{liu2019few} and StarGAN-v2 \cite{choi2019stargan} attempt to learn a domain-dependent style representation which ideally should account for the correlated attributes. Unfortunately, we show that despite their visually pleasing results, the translated images still retain many domain-specific attributes of the source image. As demonstrated in Fig.~\ref{fig:afhq}, when translating dogs to wild animals, current methods transfer facial shapes which are unique to dogs and should not be transferred precisely to wild animals, while our model transfers the semantic head pose more reliably.

In this work, we analyze the different settings for disentanglement of labeled and unlabeled attributes. In the case where they are perfectly uncorrelated, we improve over LORD \cite{gabbay2019demystifying} and scale to higher perceptual quality by a novel synthesis stage. In cases where the labeled and unlabeled attributes are correlated, we rely on the existence of spatial transformations which can retain the correlated attributes while exhibiting different or no uncorrelated attributes. We suggest simple forms of transformations for learning \textit{pose-independent} or \textit{localized} correlated attributes, by which we achieve better disentanglement both quantitatively and qualitatively than state-of-the-art methods (e.g. FUNIT \cite{liu2019few} and StarGAN-v2 \cite{choi2019stargan}). Our approach suggests that adversarial optimization, which is typically used for domain translation, is not necessary for disentanglement, and its main utility lies in generating perceptually pleasing images. Fig.~\ref{fig:teaser} summarizes the settings covered by our framework.

Our contributions are as follows: (i) Introducing a non-adversarial disentanglement method that carefully extends to cases where the attributes are correlated. (ii) Scaling disentanglement methods to high perceptual quality with a final synthesis stage while learning disentangled representations. (iii) State-of-the-art results in multiple image-translation settings within a unified framework.

\section{Related Work}
\textbf{Image Translation} Translating the content of images across different domains has attracted much attention. In the unsupervised setting, CycleGAN \cite{zhu2017unpaired} introduces a cycle consistency loss to encourage the translated images to preserve the domain-invariant attributes (e.g. pose) of the source image. MUNIT \cite{huang2018multimodal} recognizes that a given content image could be translated to many different styles (e.g. colors and textures) in a target domain and extends UNIT \cite{huang2017arbitrary} to learn multi-modal mappings by learning style representations. DRIT \cite{DRIT} tackles the same setting using an adversarial constraint at the representation level. MSGAN \cite{msgan} adds a regularization term to prevent mode collapse. StarGAN-v2 \cite{choi2019stargan} and DMIT \cite{yu2019dmit} extend previous frameworks to translation across more than two domains. FUNIT \cite{liu2019few} further allows translation to novel domains. COCO-FUNIT \cite{saito2020coco} aims to preserve the structure of the image by conditioning the style of a reference image on the actual content image. 

\textbf{Class-Supervised Disentanglement} In this parallel line of work, the goal is to anchor the semantics of each class into a single representation while modeling all the remaining class-independent attributes by a residual representation. Several methods encourage disentanglement by adversarial constraints \cite{denton2017unsupervised, szabo2017challenges, mathieu2016disentangling} while others rely on cycle consistency \cite{harsh2018disentangling} or group accumulation \cite{bouchacourt2018multi}. LORD \cite{gabbay2019demystifying} takes a non-adversarial approach and trains a generative model while directly optimizing over the latent codes. Methods in this area assume that the class label and the unlabeled attributes are perfectly uncorrelated. We show that in cases where this assumption does not hold, these methods fail to disentangle the attributes and perform poorly. Moreover, they work on synthetic and low-resolution datasets and cannot compete in recent image translation benchmarks. 

\textbf{Disentanglement of Shape and Texture}
Several methods aim to disentangle the shape of an object from its texture. Lorenz \etal \cite{lorenz2019unsupervised} exploit equivariance and invariance constraints between synthetically transformed images to learn shape and appearance representations. Park \etal \cite{park2020swapping} learn the texture by co-occurrence patch statistics across different parts of the image. In this setting, no supervision is used, limiting current methods to low-level manipulations i.e. no significant semantic changes are performed.

\section{Disentanglement in Image Translation}
\label{sec:analysis}

\begin{figure*}[t]
\begin{center}
\includegraphics[width=0.99\linewidth]{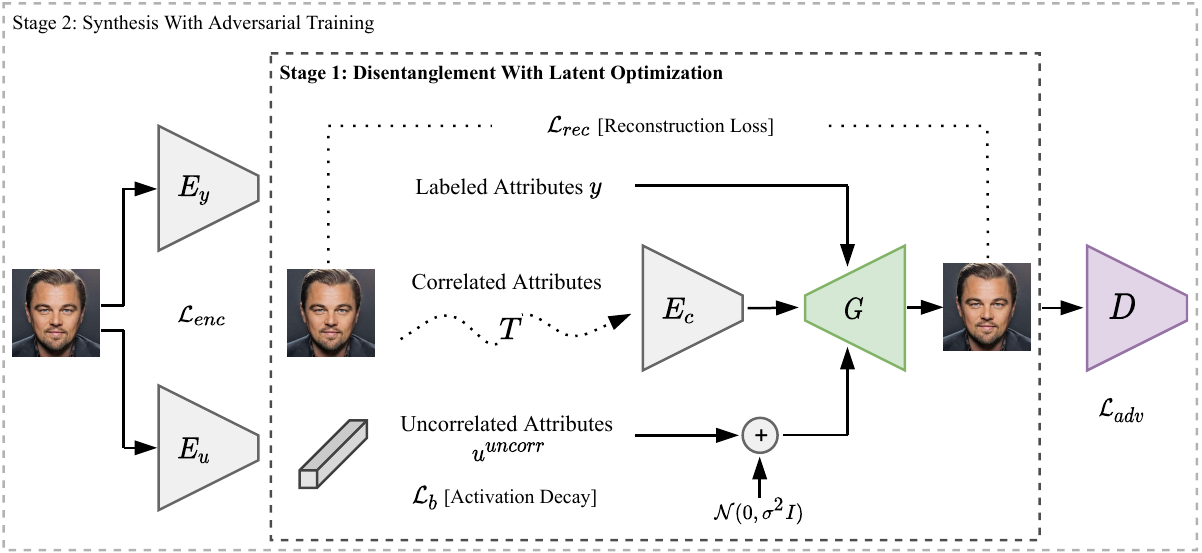}
\end{center}
\caption{A sketch of our method. In the disentanglement stage, $T$ outputs a transformed version of the image, retaining only attributes that correlate with $y$. $u^{uncorr}$ is regularized and optimized to recover the minimal residual information required for reconstructing the input image, resulting in the remaining uncorrelated attributes, as they are not represented by $y$ or the output of $T$. In the synthesis stage, we tune all modules in an amortized fashion using the learned embeddings as targets of the new encoders, $E_y$ and $E_u$. During this stage, an adversarial discriminator is trained to increase the visual fidelity.}
\label{fig:sketch}
\vspace{-0.75em}
\end{figure*}

Assume a set of images $x_1,x_2,...,x_n \in \mathcal{X}$ in which every image $x_i$ is specified precisely by a set of attributes. Some of the attributes are labeled by $y_i$, while the rest are unlabeled and denoted as $u_i$. 
%The unlabeled attributes include attributes which are correlated with $y_i$, denoted as $u_i^{corr}$, and uncorrelated $u_i^{uncorr}$.
~\\\\
\textbf{Case 1:}
If the labeled and unlabeled attributes are \textit{approximately uncorrelated}, we seek representations $u^{uncorr}$ for the unlabeled attributes. An unknown function $G$ maps $y$ and $u^{uncorr}$ to $x$:

\begin{equation}
\label{eq:formation_uncorrelated}
    x_i = G(y_i, u_i^{uncorr})
\end{equation}
% ~\\
\textbf{Case 2:}
If the labeled and unlabeled attributes are correlated, $y$ and $u^{uncorr}$ do not uniquely specify $x$. For example, in animal face images, $y$ defines the species (e.g. cat or dog) and $u^{uncorr}$ specifies uncorrelated attributes such as head pose. The attributes which are correlated with species (e.g. breed) are denoted as $u^{corr}$, and should be learned as well. $G$ now maps $y$, $u^{corr}$ and $u^{uncorr}$ to $x$:

\begin{equation}
\label{eq:formation_correlated}
    x_i = G(y_i, u_i^{corr}, u_i^{uncorr})
\end{equation}

The goal of image translation is to replace the labeled attributes $y_i$ along with their correlated attributes $u_i^{corr}$ of source image $x_i$ by those of target image $x_j$.
% \begin{equation}
% \label{eq:transfer}
%     x_{ij} = G(y_j, u_j^{corr}, u_i^{uncorr})
% \end{equation}
Let us briefly review the main ideas proposed in previous work and discuss their drawbacks.

\paragraph{Learning the uncorrelated attributes $u^{uncorr}$} Adversarial methods (e.g. StarGAN-v2, FUNIT, MUNIT) train a domain confusion discriminator on the translated images to retain only the uncorrelated attributes. Unfortunately, we empirically show in Tab.~\ref{tab:afhq} and Fig.~\ref{fig:afhq} that these methods do not learn disentangled representations and the translated images do not retain only the uncorrelated attributes. However, some of the correlated attributes leak into the representations and the translated images. We hypothesize that this failure lies in the challenging adversarial optimization.

\paragraph{Learning the correlated attributes $u^{corr}$}
Current methods rely on locality-preserving architectures (e.g. AdaIN \cite{choi2019stargan,liu2019few,huang2018multimodal}) that bias the uncorrelated attributes to represent the spatial structure of the image and the correlated attributes to control its appearance. Unfortunately, architectural biases of this sort inevitably restrict the correlated attributes from altering spatial features when needed. As shown in Fig.~\ref{fig:afhq}, the uncorrelated pose is indeed spatial, but some of the correlated attributes (as the facial shape of different breeds) are of spatial nature as well.

\section{Proposed Framework: OverLORD}
\label{sec:method}

Motivated by the above analysis, we present OverLORD, which is composed of: \textit{disentanglement} and \textit{synthesis}.

\subsection{Learning a Disentanglement Model}
In order to learn disentangled representations for the labeled and unlabeled attributes, we draw inspiration from LORD \cite{gabbay2019demystifying} and propose a non-adversarial method based on latent optimization. Differently from LORD, we do not strictly assume that the labeled and unlabeled attributes are uncorrelated. We \textit{relax} this assumption and assume the existence of correlated attributes, as described next.

\paragraph{Unlabeled correlated attributes} We design a simple yet effective method for learning some of the correlated attributes independently from the uncorrelated ones. We first form a function $T$ that outputs an image $x^{corr} = T(x)$, which retains the correlated attributes but exhibits different uncorrelated attributes. The exact realization of $T$ depends on the nature of the uncorrelated and correlated attributes. In this paper, we propose two different forms of $T$:
\\\\
\indent (i) \textit{Random spatial transformation}: If the uncorrelated attributes should capture the \textit{pose} of an object, we set $T$ as a sequence of random flips, rotations or crops, making $x^{corr}$ retain attributes that are \textit{pose-independent}:

\begin{equation}
  \label{eq:transformation_random}
  T(x) = (f_{crop} \circ f_{rotate} \circ f_{flip})(x)
\end{equation}

(ii) \textit{Masking}: If the correlated attributes are \textit{localized}, we set $T$ to mask-out the uncorrelated attributes, retaining only correlated regions included in the segmentation mask $m$:

\begin{equation}
  \label{eq:transformation_mask}
  T(x;m) = x \odot m
\end{equation}

E.g. when translating males to females, $T$ might mask the hair region which is highly correlated with the gender (i.e. masking out all the other regions). Masks can be obtained using external supervision or unsupervised methods \cite{ji2019invariant, van2021unsupervisedsseg} and are used only for training. The correlated attributes are modeled by: $u^{corr} = E_c(x^{corr})=E_c(T(x))$.

\paragraph{Unlabeled uncorrelated attributes} As $T$ accounts for the dependency of the labeled and unlabeled attributes, we can assume that $u^{uncorr}$ is \textit{approximately uncorrelated} with $y$ and $u^{corr}$. To obtain an independent representation, we aim $u^{uncorr}$ to recover the \textit{minimal residual} information, not represented by $y$ and $u^{uncorr}$, that is required to reconstruct the image. We therefore parameterize a noisy channel consisting of a vector $u'_i$ and an additive Gaussian noise $z \sim \mathcal{N}(0, I)$, $u_i^{uncorr} = u'_i + z$, similarly to the non-adversarial \textit{bottleneck} proposed in \cite{gabbay2019demystifying}. We therefore define the bottleneck loss as: $\mathcal{L}_{b} = \sum_i \|u'_i\|^2$.
~\\\\
\textit{Reconstruction:} An image $x_i$ should be fully specified by the representations $y_i,u_i^{corr},u_i^{uncorr}$:
\begin{equation}
  \label{eq:loss_rec}
  \mathcal{L}_{rec} = \sum_i \ell \big(G(y_i,u_i^{corr},u_i^{uncorr}), x_i \big)
\end{equation}

Our complete objective can be summarized as follows:
\begin{equation}
   \label{eq:objective_disentanglement}
  \min_{u'_i,E_c,G} \mathcal{L}_{disent} = \mathcal{L}_{rec} + \lambda_{b} \mathcal{L}_{b}
\end{equation}

For reconstruction we use $\ell$ as the VGG-based perceptual loss.
Note that we optimize over $u'_i$ directly, as they are \textit{not} parameterized by a feed-forward encoder. Inspired by \cite{gabbay2019demystifying}, we leverage latent optimization as it improves disentanglement significantly over encoder-based methods. To get an intuition, we should consider their initialization: each $u'_i$ in latent optimization is initialized i.i.d, and therefore is totally uncorrelated with the labeled attributes. However, a feed-forward encoder starts with near perfect correlation (the labeled attributes can be predicted even from the output of a randomly initialized encoder). We further elaborate on latent optimization and its inductive bias in Appendix~\ref{app:latent_optimization}.

\subsection{Generalization and Perceptual Quality}
After the disentanglement stage, we possess representations of the uncorrelated attributes $u_i^{uncorr}$ for every image in the training set. In order to generalize to unseen images and novel labeled attributes (e.g. new face identities), we train feed-forward encoders $E_y: \mathcal{X} \xrightarrow{} \mathcal{Y}$ and $E_u: \mathcal{X} \xrightarrow{} \mathcal{U}$ to infer the labeled and uncorrelated attributes ($E_c$ is already trained in the previous stage), respectively:
\begin{equation}
    \label{eq:loss_enc}
    \mathcal{L}_{enc} = \sum_i \|E_y(x_i) - y_i\|^2 + \|E_u(x_i) - u_i^{uncorr}\|^2
\end{equation}

The reconstruction term is changed accordingly:
\begin{equation}
    \label{eq:loss_gen}
    \mathcal{L}_{gen} = \sum_i \ell\Big(G\big(E_y(x_i), E_c(x_i), E_u(x_i)\big), x_i\Big)
\end{equation}

These two objectives ensure that the training set can be reconstructed in an amortized fashion ($\mathcal{L}_{gen}$), without violating the disentanglement criterion established in the previous stage ($\mathcal{L}_{enc}$). Note that the targets for $\mathcal{L}_{enc}$ are the ones learned by our own model in the previous stage.

Although these constraints are sufficient for disentanglement and generalization, the generated images exhibit relatively low perceptual quality, as can be seen in Fig.~\ref{fig:ablation_summary}. While we argue that achieving disentanglement by adversarial methods is notoriously difficult, as observed in our experiments and in \cite{gabbay2019demystifying}, we find that tuning the model with an adversarial term after disentangled representations have been already learned significantly improves the perceptual fidelity. 
Therefore, we jointly train an unconditional discriminator $D$ and employ an adversarial loss to distinguish between real and reconstructed images:

\begin{equation}
\label{eq:adv}
    \mathcal{L}_{adv} = \sum_i \log D(x_i) + \log \big(1 - D(\bar{x_i})\big)
\end{equation}

The entire objective of the synthesis stage is as follows:
\begin{equation}
    \label{eq:objective_synthesis}
    \min_{E_y, E_u, G} ~ \max_{D} ~~ \mathcal{L}_{gen} + \lambda_{enc} \mathcal{L}_{enc} + \lambda_{adv} \mathcal{L}_{adv}
\end{equation}

A sketch of our entire framework is shown in Fig.~\ref{fig:sketch}.

\section{Experiments}
\label{sec:experiments}
Our framework is evaluated in the two cases considered along this paper: when the labeled and unlabeled attributes are uncorrelated, and when they are correlated. %Note that we do not compare to unsupervised disentanglement methods such as betaVAE \cite{higgins2016beta} as they can not compete with methods that are supervised with labeled attributes.

\subsection{Evaluation Protocol}
We measure disentanglement by training two auxiliary classifiers to predict the original labeled attributes from: (i) the learned representations of the uncorrelated attributes. (ii) the translated images in which the labeled attributes have been changed. In both metrics, lower accuracy indicates better disentanglement. In cases where annotations for evalution exist (e.g. CelebA) we measure task-specific metrics as identity similarity (Id), expression (Exp) and head pose (Pose) errors given a target image. Where no annotations exist (e.g. AFHQ, CelebA-HQ) we measure how the translated images fit the target domain (FID) and their diversity (LPIPS). More implementation and evaluation details are provided in Appendix~\ref{app:implementation_details}.

\subsection{Uncorrelated Attributes}
\begin{figure*}[t]
\begin{center}
\begin{tabular}{@{\hskip0pt}c@{\hskip2pt}c@{\hskip1pt}c@{\hskip0pt}c@{\hskip0pt}c@{\hskip2pt}c@{\hskip1pt}c@{\hskip0pt}c@{\hskip0pt}c}
Identity & Attributes 1 & FUNIT \cite{liu2019few} & LORD \cite{gabbay2019demystifying} & \textbf{Ours} & Attributes 2 & FUNIT \cite{liu2019few} & LORD \cite{gabbay2019demystifying} & \textbf{Ours} \\

\includegraphics[width=0.105\linewidth]{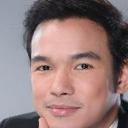} & \includegraphics[width=0.105\linewidth]{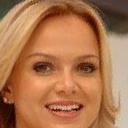} & \includegraphics[width=0.105\linewidth]{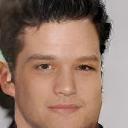} &
\includegraphics[width=0.105\linewidth]{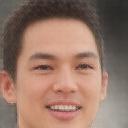} &
\includegraphics[width=0.105\linewidth]{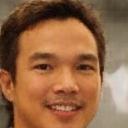} &
\includegraphics[width=0.105\linewidth]{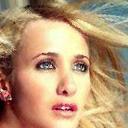} &
\includegraphics[width=0.105\linewidth]{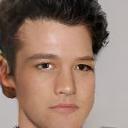} &
\includegraphics[width=0.105\linewidth]{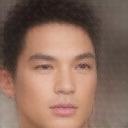} &
\includegraphics[width=0.105\linewidth]{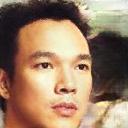} \\

\includegraphics[width=0.105\linewidth]{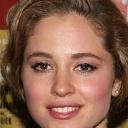} & \includegraphics[width=0.105\linewidth]{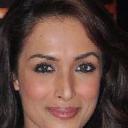} & \includegraphics[width=0.105\linewidth]{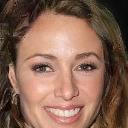} &
\includegraphics[width=0.105\linewidth]{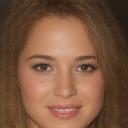} &
\includegraphics[width=0.105\linewidth]{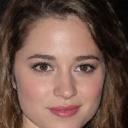} &
\includegraphics[width=0.105\linewidth]{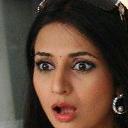} &
\includegraphics[width=0.105\linewidth]{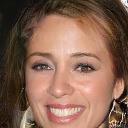} &
\includegraphics[width=0.105\linewidth]{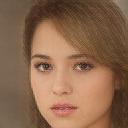} &
\includegraphics[width=0.105\linewidth]{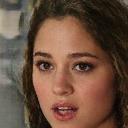} \\

\end{tabular}
\end{center}
\caption{Disentanglement of 10K facial identities from unlabeled attributes (e.g. head pose, expression, illumination). FUNIT preserves pose but fails to model expression and exact identity. LORD captures the identity but generates low-fidelity images. We preserve head pose and facial expression while transferring the identity and generating appealing images.}
\label{fig:celeba}
\end{figure*}

\begin{figure*}[t]
\begin{center}
\begin{tabular}{@{\hskip0pt}c@{\hskip3pt}c@{\hskip2pt}c@{\hskip0pt}c@{\hskip0pt}c@{\hskip0pt}c@{\hskip0pt}c@{\hskip0pt}c@{\hskip0pt}c@{\hskip0pt}c}

& Original & [0-9] & [10-19] & [20-29] & [30-39] & [40-49] & [50-59] & [60-69] & [70-79] \\

\begin{turn}{90} Lifespan \cite{orel2020lifespan} \end{turn} &
\includegraphics[width=0.105\linewidth]{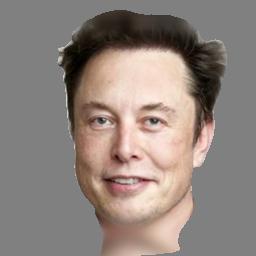} &
\includegraphics[width=0.105\linewidth]{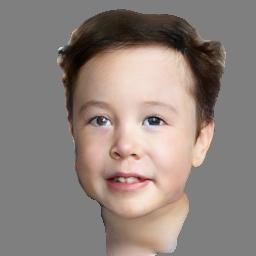} &
\includegraphics[width=0.105\linewidth]{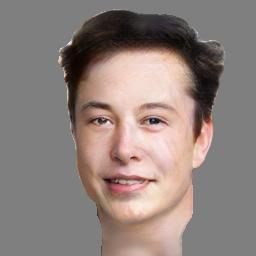} &
\includegraphics[width=0.105\linewidth]{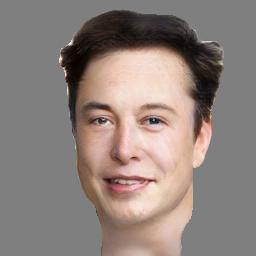} &
\includegraphics[width=0.105\linewidth]{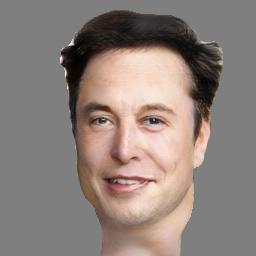} &
\includegraphics[width=0.105\linewidth]{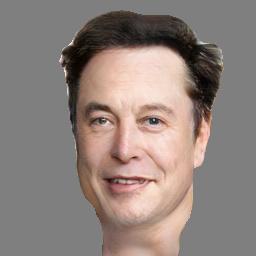} &
\includegraphics[width=0.105\linewidth]{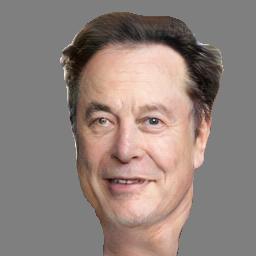} &
\includegraphics[width=0.105\linewidth]{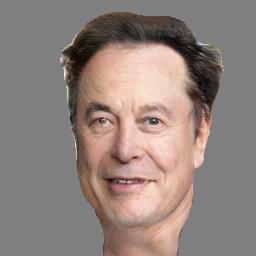} &
~ \\

\begin{turn}{90} ~~~SAM \cite{alaluf2021sam} \end{turn} &
\includegraphics[width=0.105\linewidth]{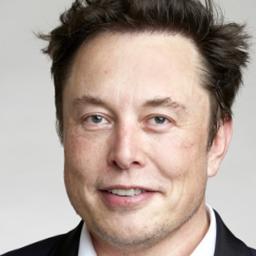} &
\includegraphics[width=0.105\linewidth]{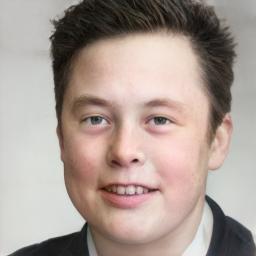} &
\includegraphics[width=0.105\linewidth]{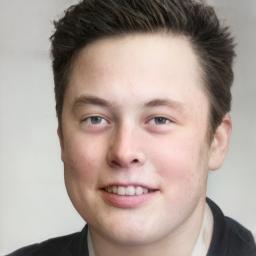} &
\includegraphics[width=0.105\linewidth]{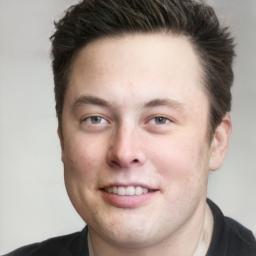} &
\includegraphics[width=0.105\linewidth]{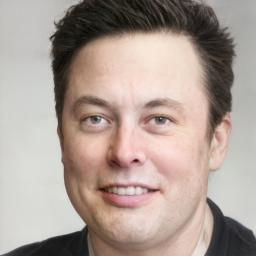} &
\includegraphics[width=0.105\linewidth]{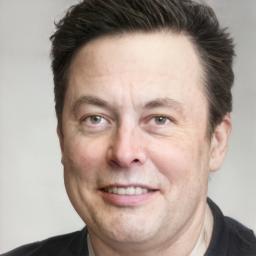} &
\includegraphics[width=0.105\linewidth]{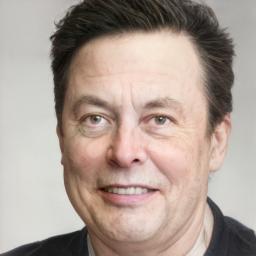} &
\includegraphics[width=0.105\linewidth]{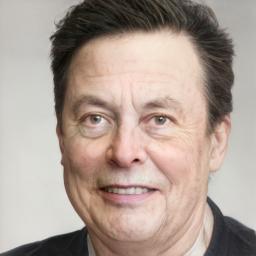} &
\includegraphics[width=0.105\linewidth]{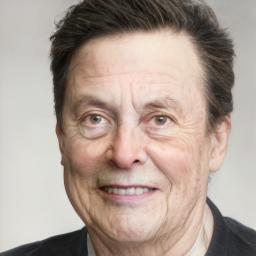} \\

\begin{turn}{90} ~~~~~~\textbf{Ours} \end{turn} &
\includegraphics[width=0.105\linewidth]{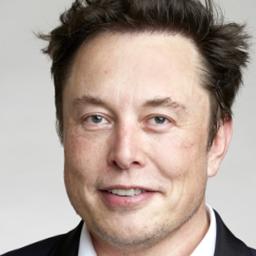} &
\includegraphics[width=0.105\linewidth]{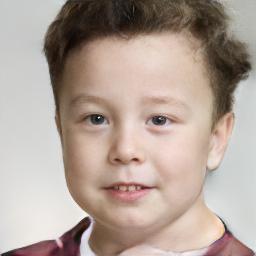} &
\includegraphics[width=0.105\linewidth]{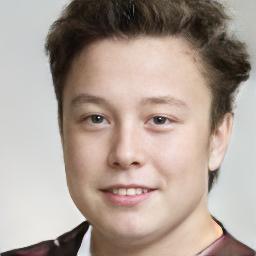} &
\includegraphics[width=0.105\linewidth]{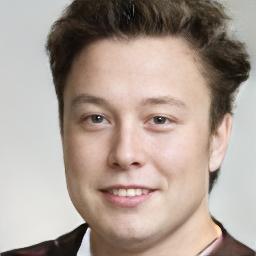} &
\includegraphics[width=0.105\linewidth]{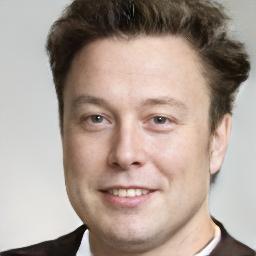} &
\includegraphics[width=0.105\linewidth]{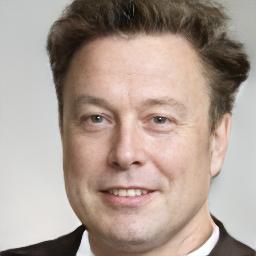} &
\includegraphics[width=0.105\linewidth]{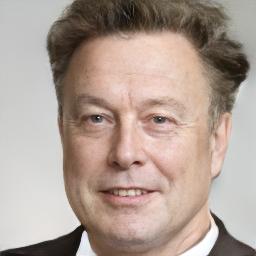} &
\includegraphics[width=0.105\linewidth]{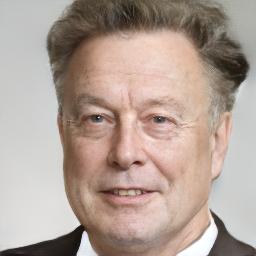} &
\includegraphics[width=0.105\linewidth]{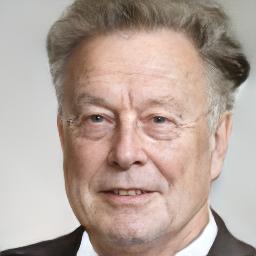} \\

\begin{turn}{90} Lifespan \cite{orel2020lifespan} \end{turn} &
\includegraphics[width=0.105\linewidth]{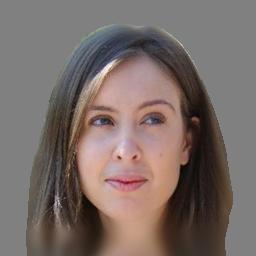} &
\includegraphics[width=0.105\linewidth]{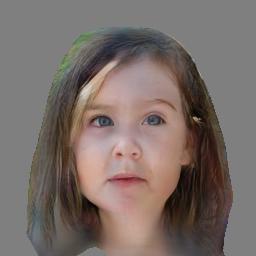} &
\includegraphics[width=0.105\linewidth]{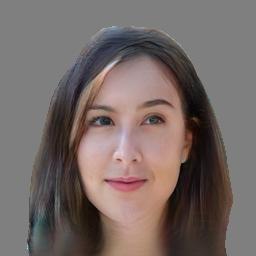} &
\includegraphics[width=0.105\linewidth]{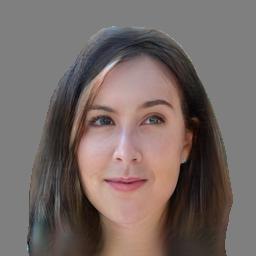} &
\includegraphics[width=0.105\linewidth]{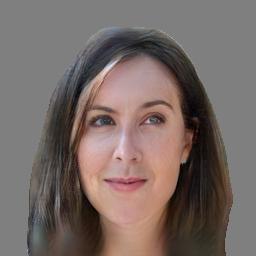} &
\includegraphics[width=0.105\linewidth]{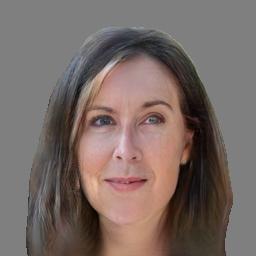} &
\includegraphics[width=0.105\linewidth]{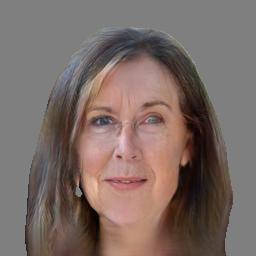} &
\includegraphics[width=0.105\linewidth]{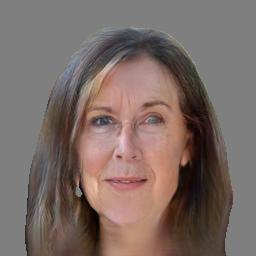} &
~ \\

\begin{turn}{90} ~~~SAM \cite{alaluf2021sam} \end{turn} &
\includegraphics[width=0.105\linewidth]{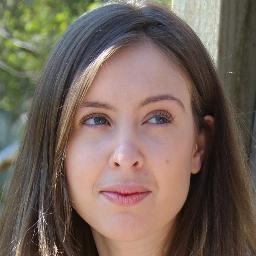} &
\includegraphics[width=0.105\linewidth]{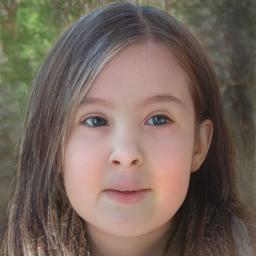} &
\includegraphics[width=0.105\linewidth]{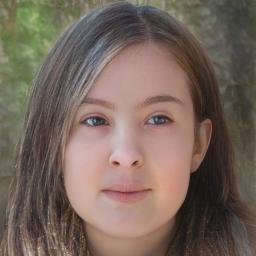} &
\includegraphics[width=0.105\linewidth]{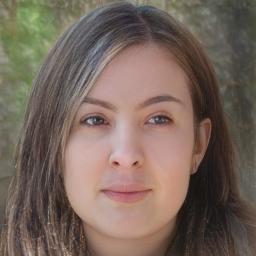} &
\includegraphics[width=0.105\linewidth]{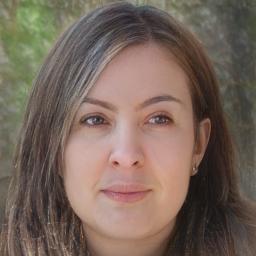} &
\includegraphics[width=0.105\linewidth]{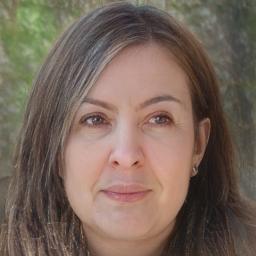} &
\includegraphics[width=0.105\linewidth]{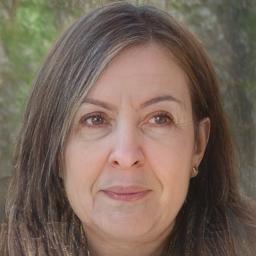} &
\includegraphics[width=0.105\linewidth]{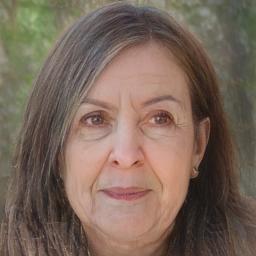} &
\includegraphics[width=0.105\linewidth]{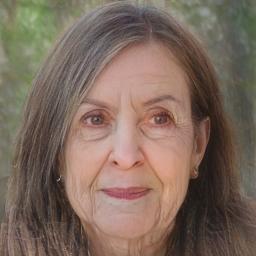} \\

\begin{turn}{90} ~~~~~~\textbf{Ours} \end{turn} &
\includegraphics[width=0.105\linewidth]{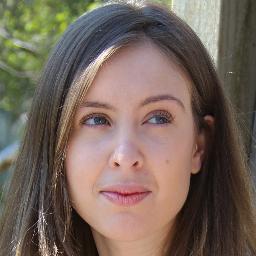} &
\includegraphics[width=0.105\linewidth]{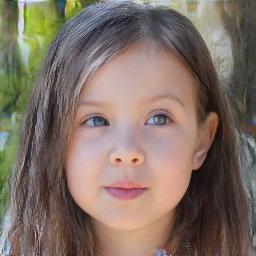} &
\includegraphics[width=0.105\linewidth]{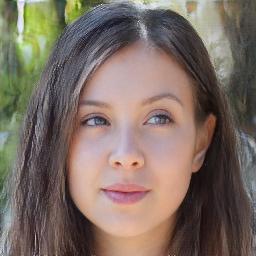} &
\includegraphics[width=0.105\linewidth]{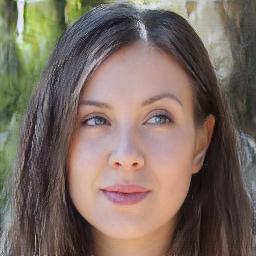} &
\includegraphics[width=0.105\linewidth]{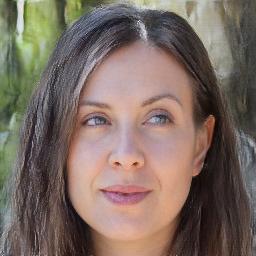} &
\includegraphics[width=0.105\linewidth]{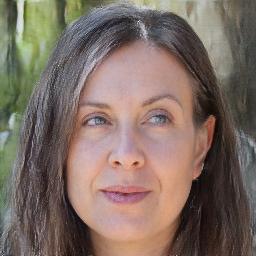} &
\includegraphics[width=0.105\linewidth]{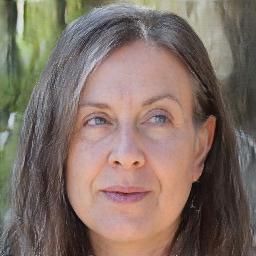} &
\includegraphics[width=0.105\linewidth]{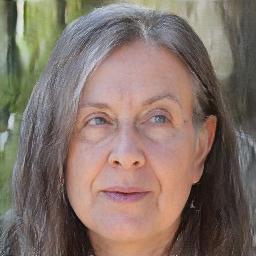} &
\includegraphics[width=0.105\linewidth]{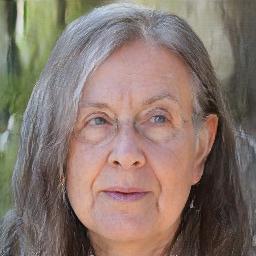} \\

\end{tabular}
\end{center}
\caption{Disentanglement of age and unlabeled attributes. Lifespan introduces artifacts and does not change hair color, while StyleGAN-based SAM struggles to preserve identity. Note that we \textbf{do not use any supervised identity loss}.}
\label{fig:aging}
\vspace{-1em}
\end{figure*}

\paragraph{Attribute Manipulation} In this setting there is a single labeled attribute which is assumed to be \textit{approximately uncorrelated} with all the unlabeled attributes. We perform an experiment on CelebA \cite{liu2015faceattributes} and define the facial identity label as the only labeled attribute. Tab.~\ref{tab:celeba} and Fig.~\ref{fig:celeba} show results of disentangling face identity from the unlabeled attributes including head pose, expression and illumination. We compare to LORD and FUNIT as they can handle fine-grained labels (i.e. 10K face identities) and generalize to unseen identities. It can be seen that our method preserves the head pose, expression and illumination well while translating the identity more faithfully than the baselines. Another task which has attracted much attention recently is facial age editing \cite{alaluf2021sam, orel2020lifespan}. We define the age (categorized to 8 ranges: 0-9, ..., 70-79) as the only labeled attribute in FFHQ and compare to state-of-the-art aging methods: lifespan \cite{orel2020lifespan} and SAM \cite{alaluf2021sam}. We demonstrate in Fig.~\ref{fig:aging} that while the baselines rely on a pretrained face identity loss \cite{deng2019arcface} we are able to preserve the identity better. 
Finally, we explore the task of translating Males to Females. We define the gender in CelebA-HQ \cite{karras2018progressive} as the labeled attribute and compare to FaderNetworks \cite{lample2017fader} and mGANprior \cite{gu2020mganprior} which operates in the latent space of StyleGAN. As shown in Tab.~\ref{tab:celebahq_gender} and Fig.~\ref{fig:celebahq_gender}, our model achieves near perfect score in fooling a target classifier and generates visually pleasing results. We later show that the assumption that the gender is uncorrelated with the other attributes can be relaxed by masking regions (e.g. hair) that are highly correlated with the gender to improve translation control and reliability.
~\\\\
\textbf{Shape-Texture} In this task the goal is to disentangle the shape of an object from its texture. We define the shape as the labeled attribute (represented by an edge map) and demonstrate texture transfer between different shoes on Edges2Shoes \cite{yu2014edges2shoes} in Fig.~\ref{fig:teaser} and Appendix~\ref{app:additional_results}. Note that in order to guide the labeled attribute by an image (and not by a categorical label), we train $E_y$ with the rest of the models in the disentanglement stage instead of the synthesis stage.

\subsection{Correlated Attributes}
\begin{figure*}[t]
\begin{center}
\begin{tabular}{@{\hskip0pt}c@{\hskip3pt}c@{\hskip2pt}c@{\hskip0pt}c@{\hskip0pt}c@{\hskip6pt}c@{\hskip2pt}c@{\hskip0pt}c@{\hskip0pt}c@{\hskip0pt}c@{\hskip0pt}c}

& Input & Fader & mGANprior & \textbf{Ours [uncorr]} & Reference & StarGAN-v2 & \textbf{Ours [corr]} \\
\multirow{2}{*}[5ex]{\rotatebox[origin=c]{90}{Male-to-Female}} &
\includegraphics[width=0.13\linewidth]{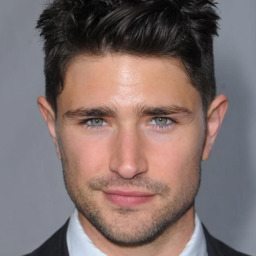} &
\includegraphics[width=0.13\linewidth]{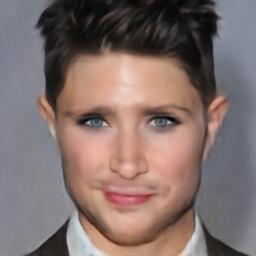} &
\includegraphics[width=0.13\linewidth]{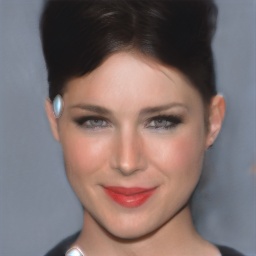} &
\includegraphics[width=0.13\linewidth]{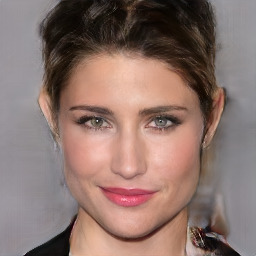} &
\includegraphics[width=0.13\linewidth]{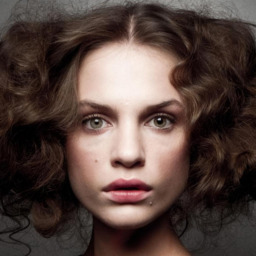} &
\includegraphics[width=0.13\linewidth]{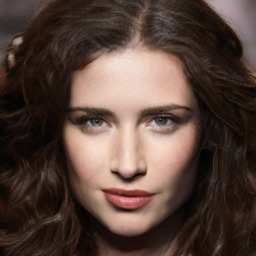} &
\includegraphics[width=0.13\linewidth]{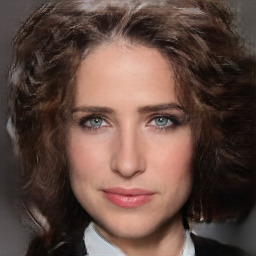} \\

&
\includegraphics[width=0.13\linewidth]{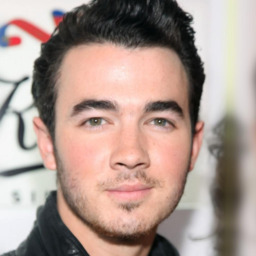} &
\includegraphics[width=0.13\linewidth]{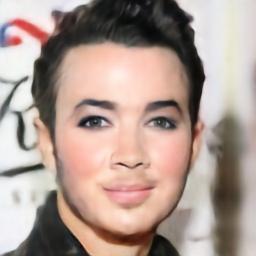} &
\includegraphics[width=0.13\linewidth]{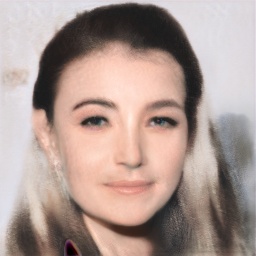} &
\includegraphics[width=0.13\linewidth]{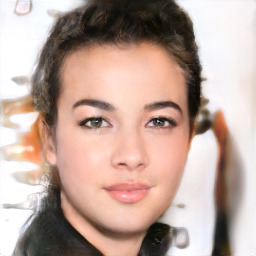} &
\includegraphics[width=0.13\linewidth]{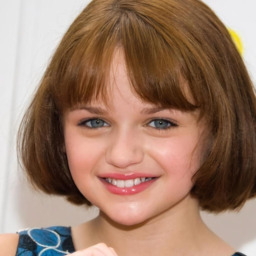} &
\includegraphics[width=0.13\linewidth]{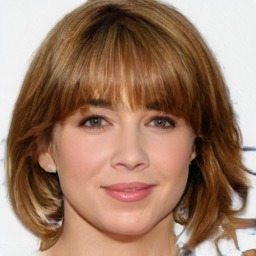} &
\includegraphics[width=0.13\linewidth]{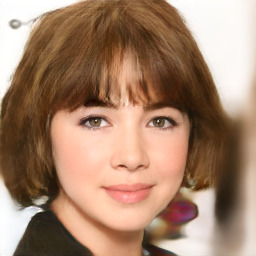} \\

\multirow{2}{*}[5ex]{\rotatebox[origin=c]{90}{Female-to-Male}} &
\includegraphics[width=0.13\linewidth]{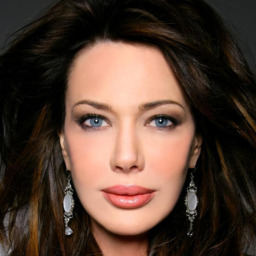} &
\includegraphics[width=0.13\linewidth]{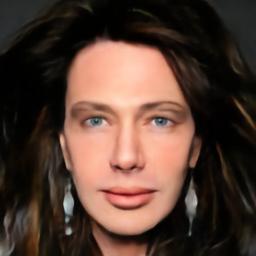} &
\includegraphics[width=0.13\linewidth]{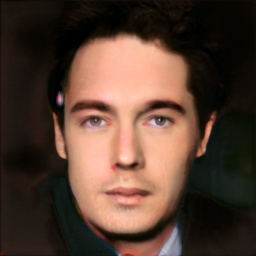} &
\includegraphics[width=0.13\linewidth]{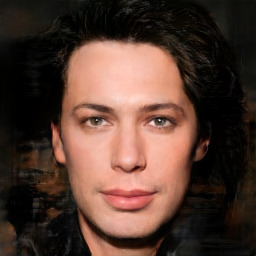} &
\includegraphics[width=0.13\linewidth]{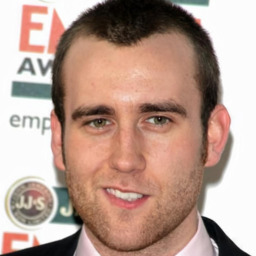} &
\includegraphics[width=0.13\linewidth]{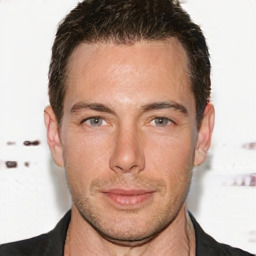} &
\includegraphics[width=0.13\linewidth]{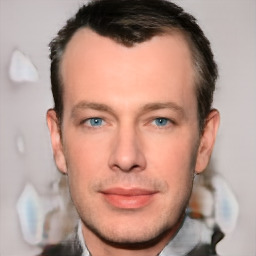} \\

& \includegraphics[width=0.13\linewidth]{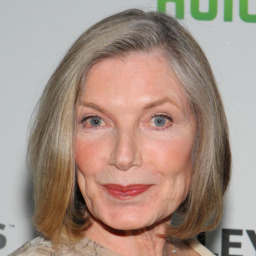} &
\includegraphics[width=0.13\linewidth]{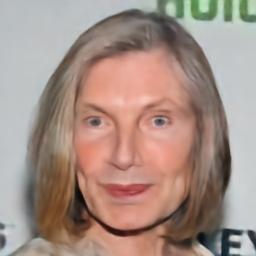} &
\includegraphics[width=0.13\linewidth]{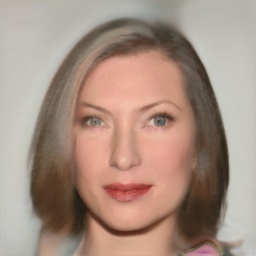} &
\includegraphics[width=0.13\linewidth]{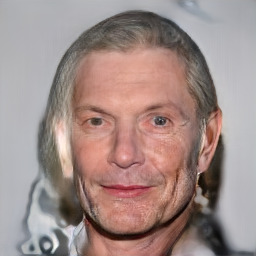} &
\includegraphics[width=0.13\linewidth]{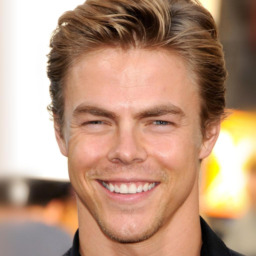} &
\includegraphics[width=0.13\linewidth]{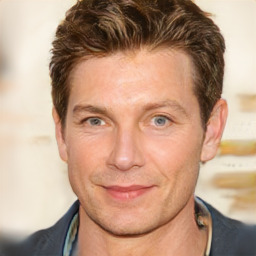} &
\includegraphics[width=0.13\linewidth]{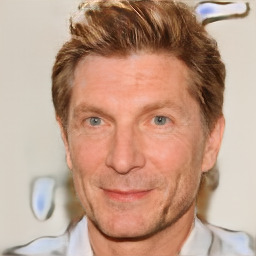} \\

\begin{turn}{90} ~~~~~~~~Failure \end{turn} & 
\includegraphics[width=0.13\linewidth]{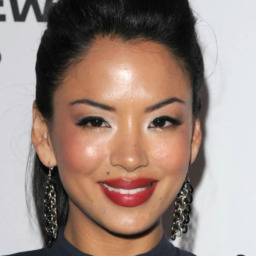} &
\includegraphics[width=0.13\linewidth]{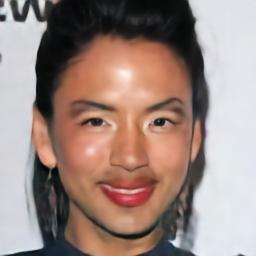} &
\includegraphics[width=0.13\linewidth]{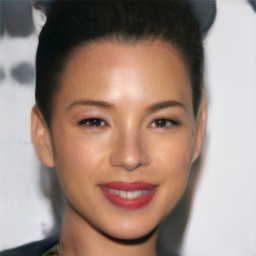} &
\includegraphics[width=0.13\linewidth]{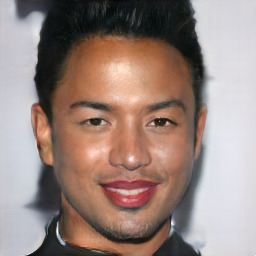} &
\includegraphics[width=0.13\linewidth]{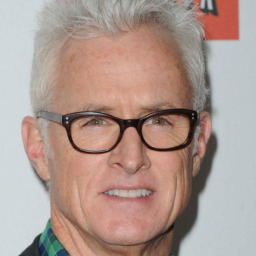} &
\includegraphics[width=0.13\linewidth]{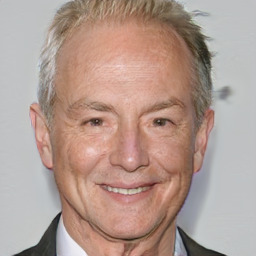} &
\includegraphics[width=0.13\linewidth]{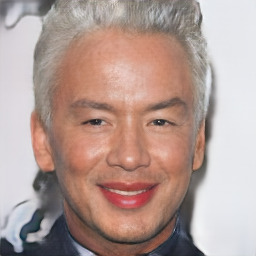} 

\end{tabular}
\end{center}
\caption{Male-to-Female translation in two settings: (i) When the attributes are assumed to be \textit{uncorrelated}, our method makes greater changes than Fader while preserving the identity better than mGANprior. (ii) As translating hairstyle between genders is poorly specified, we model it as the \textit{correlated} attribute and utilize a reference image specifying its target. Our method preserves the uncorrelated attributes including identity, age and illumination better than StarGAN-v2, while transferring hairstyle more faithfully. \textit{Failure case}: as the lipstick is not modeled as a correlated attribute, it transfers over unreliably.}
\label{fig:celebahq_gender}
\vspace{-1em}
\end{figure*}

\paragraph{Pose-Appearance} When the uncorrelated attributes should encode the pose of an object, we set $T$ to be a series of random transformation of horizontal flips, rotations and crops to let $x^{corr}$ retain all the \textit{pose-independent} attributes. We perform an experiment on AFHQ and define the domain label (cat, dog, wildlife) as the labeled attribute. Tab.~\ref{tab:afhq} and Fig.~\ref{fig:afhq} show that our method outperforms all baselines achieving near perfect disentanglement with better visual quality (FID) and higher translation diversity (LPIPS). Note that while StarGAN-v2 and FUNIT support multiple domains, MUNIT, DRIT and MSGAN are trained multiple times for every possible pair of domains. Moreover, as standard LORD does not differentiate between correlated and uncorrelated attributes i.e. can not utilize a reference image, we make an effort to extend it to the correlated case by clustering the images into 512 sub-categories before training (denoted as LORD clusters).
~\\\\
\textbf{Localized correlation} When the correlated attributes are \textit{localized}, $T$ masks their corresponding regions within the image. We repeat the Male-to-Female experiment with masks provided in CelebAMask-HQ \cite{CelebAMask-HQ}. In this experiment, $x^{corr}$ contains only the hair region. As shown in Fig.~\ref{fig:celebahq_gender}, our method translates the gender with the general target hair style while preserving the uncorrelated attributes including identity, age and illumination better than StarGAN-v2. More results are in Appendix.~\ref{app:additional_results}. 

\begin{table}
\centering
\caption{Disentanglement performance on CelebA. Lower classification accuracy of identity from the learned representations of the uncorrelated attributes ($y \xleftarrow{} u^{uncorr}$) and higher error of landmarks regression from identity representations ($y \xrightarrow{} lnd$) indicate better disentanglement. Id = FaceNet cosine similarity, Exp = Expression by RMSE on facial landmarks, Pose = average of yaw, pitch and roll angle errors. *Indicates detail loss.}
\label{tab:celeba}
\begin{tabular}{l|cc|ccc}
    \toprule
    & \multicolumn{2}{c}{Representation} & \multicolumn{3}{|c}{Image} \\
    \cmidrule(lr){2-6}
    & \scriptsize $y \xleftarrow{} u^{uncorr}$ & \scriptsize $y \xrightarrow{} lnd$ & Id & Exp & Pose \\
    \midrule
    FUNIT & 0.16 & 2.6 & 0.24 & 3.8 & 4.7 \\
    LORD & $\mathbf{10^{-4}*}$ & \textbf{3.6} & 0.48 & 3.2 & 3.5 \\
    Ours & $\mathbf{10^{-3}}$ & \textbf{3.6} & \textbf{0.63} & \textbf{2.7} & \textbf{2.5} \\
    \midrule
    Optimal & $10^{-3}$ & - & 1 & 0 & 0 \\
	\bottomrule
\end{tabular}
% \vspace{-0.5em}
\end{table}

\begin{table}
\centering
\caption{Disentanglement performance on AFHQ. Classification accuracy of domain label from the learned representations of the uncorrelated attributes ($y \xleftarrow{} u^{uncorr}$), classification accuracy of source domain label from the translated image ($y_i \xleftarrow{} x_{ij}$), translation fidelity (FID) and translation diversity (LPIPS).}
\label{tab:afhq}
\begin{tabular}{l|c|ccc}
    \toprule
    & \multicolumn{1}{c}{Rep.} & \multicolumn{3}{|c}{Image} \\
    \cmidrule(lr){2-5}
    & \scriptsize $y \xleftarrow{} u^{uncorr}$ & \scriptsize $y_i \xleftarrow{} x_{ij}$ & FID & LPIPS \\
    \midrule
    MUNIT \cite{huang2018multimodal} & 1.0 & 1.0 & 223.9 & 0.20 \\
    DRIT \cite{DRIT} & 1.0 & 1.0 & 114.8 & 0.16 \\
    MSGAN \cite{msgan} & 1.0 & 1.0 & 69.8 & 0.38 \\
    LORD \cite{gabbay2019demystifying} & 0.74 & 0.47 & 97.1 & 0 \\
    LORD clusters & 0.53 & 0.43 & 37.1 & 0.36 \\
    StarGAN-v2 & 0.89 & 0.75 & 19.8 & 0.43 \\
    FUNIT \cite{liu2019few} & 0.94 & 0.85 & 18.8 & 0.44 \\
    \midrule
    % Ours w/o class (unsupervised) & 0.65 & 0.40 & 19.2 & 0.51 \\
    Ours w/o $x^{corr}$ & 0.80 & 0.79 & 55.9 & 0 \\
    Ours w/o adv. & \textbf{0.33} & \textbf{0.38} & 29.1 & 0.45 \\
    Ours & \textbf{0.33} & 0.42 & \textbf{16.5} & \textbf{0.51} \\
    \midrule
    Optimal & 0.33 & 0.33 & 12.9 & - \\
	\bottomrule
\end{tabular}
% \vspace{-0.75em}
\end{table}

\begin{figure}
\begin{center}
\begin{tabular}{@{\hskip0pt}c@{\hskip0pt}c@{\hskip1pt}c@{\hskip1pt}c@{\hskip0pt}c@{\hskip0pt}}
Pose & w/o $x^{corr}$ & Appearance & w/o adv. & \textbf{Ours} \\
\includegraphics[width=0.19\linewidth]{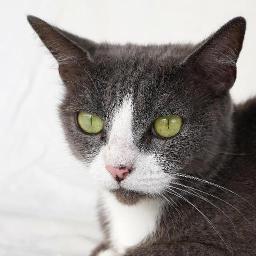} &
\includegraphics[width=0.19\linewidth]{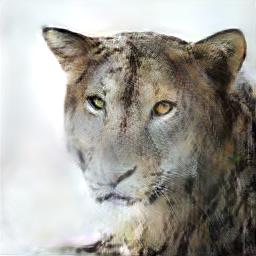} &
\includegraphics[width=0.19\linewidth]{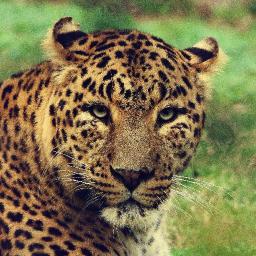} &
\includegraphics[width=0.19\linewidth]{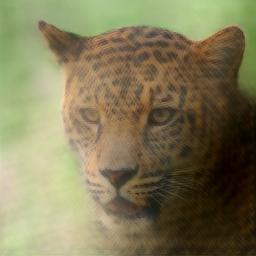} &
\includegraphics[width=0.19\linewidth]{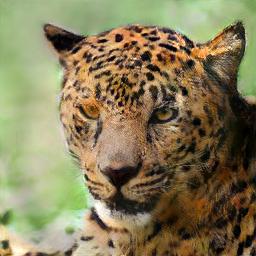}
\end{tabular}
\end{center}
\caption{Ablations: w/o $x^{corr}$: Ignoring the correlated attributes leads to unreliable translations. w/o adv.: the attributes are disentangled but images exhibit low quality. Adversarial synthesis greatly improves perceptual quality.}
\label{fig:ablation_summary}
\vspace{-0.5em}
\end{figure}

\subsection{Ablation Study}
\paragraph{Ignoring correlation with the labeled attribute} We train our model on AFHQ without $x^{corr}$, assuming all the unlabeled attributes are uncorrelated with the labeled attribute. However, as breed is highly correlated with species, the animal faces are translated in an unreliable and entangled manner, as shown in Fig.~\ref{fig:ablation_summary} and in Appendix.~\ref{app:ablation}. Tab.~\ref{tab:afhq} includes results from the ablation analysis which suggest that this strategy does not reach disentanglement neither at the representation nor at the image level. %i.e. the original species can be predicted from the codes of the uncorrelated attributes.

\paragraph{Adversarial loss for perceptual quality} We train our method without the adversarial loss in the synthesis stage. Tab.~\ref{tab:afhq} suggests that disentanglement is achieved by our non-adversarial framework, while the additional adversarial loss contributes to increasing the output fidelity (FID). Qualitative evidence for the claims in the ablation analysis are presented in Fig.~\ref{fig:ablation_summary} and in Appendix~\ref{app:ablation}.
% ~\\\\
\paragraph{Limitations} There should be noted two limitations of the proposed framework; (i) We design $T$ for learning \textit{pose-independent} or \textit{localized} correlated attributes, which covers common image translation settings. Nonetheless, other settings may require different forms of $T$. (ii) As our framework relies on latent optimization and avoids locality-biased architectures for improving disentanglement, it is not well optimized for cases in which the object is not the major part of the image e.g. where the background contains other objects or large variation. We hypothesize that this can be better solved with unsupervised segmentation techniques. %We leave this direction to future work.

\begin{table}
\centering
\caption{Male-to-Female results on CelebA-HQ. Accuracy of fooling a gender classifier and translation fidelity (FID).}
\label{tab:celebahq_gender}
\begin{tabular}{l|cc|cc}
    \toprule
    & \multicolumn{2}{c}{Target Classification} & \multicolumn{2}{|c}{FID} \\
    \cmidrule(lr){2-5}
    & F2M & M2F & F2M & M2F \\
    \midrule
    Fader \cite{lample2017fader} & 0.82 & 0.80 & 119.7 & 81.7 \\
    mGANprior \cite{gu2020mganprior} & 0.59 & 0.76 & 78.2 & 45.3 \\
    Ours [uncorr] & \textbf{0.98} & \textbf{0.97} & \textbf{54.0} & \textbf{42.9} \\
    \midrule
    StarGAN-v2 \cite{choi2019stargan} & \textbf{0.98} & \textbf{0.99} & \textbf{27.9} & 20.1 \\
    Ours [corr] & \textbf{0.98} & \textbf{0.99} & \textbf{28.1} & \textbf{16.4} \\
    \midrule
    Optimal (Real) & 0.99 & 0.99 & 15.6 & 14.0 \\
	\bottomrule
\end{tabular}
\vspace{-1.5em}
\end{table}

\section{Conclusion}
We presented OverLORD, a framework for representation disentanglement in the case where supervision is provided for some attributes while not for others. Our model extends latent-optimization-based methods to cases where the attributes are correlated, and scales-up for high-fidelity image synthesis. We further showed how adversarial optimization can be decoupled from representation disentanglement and be applied only to increase the perceptual quality of the generated images. With our unified framework, we achieve state-of-the-art performance compared to both general and task-specific methods on various image translation tasks with different forms of labeled attributes.

{\small
\bibliographystyle{ieee_fullname}
\bibliography{egbib}
}

\newpage
\clearpage

 %%%%%%%%%%%%%%%%%%%%%%%%%%%%%%%%%%%%% appendix %%%%%%%%%%%%%%%%%%%%%%%%%%%%%%%%%%%%%
\appendix
\section{Appendix}

\subsection{Implementation details}
\label{app:implementation_details}
\paragraph{Architectures} We borrow the generator and discriminator architectures from StyleGAN2 with two modifications; (i) The latent code is adapted to our formulation and forms a concatenation of three separate latent codes: $y$, $u^{corr}$, $u^{uncorr}$. (ii) We do not apply any noise injection during training. A brief summary of the architectures is presented in Tab.~\ref{tab:generator} and \ref{tab:discriminator} for completeness. The architecture of the feed-forward encoders trained in the synthesis stage is influenced by StarGAN-v2 \cite{choi2019stargan} and presented in Tab.~\ref{tab:encoder}. Note that we do not use any domain-specific layers and our model is able to generalize to unseen domains (e.g. labeled attribute) at inference time (e.g. new face identities in CelebA).

\paragraph{Optimization} In the disentanglement stage, we optimize over a single $u^{uncorr}$ embedding per image (dim = 64), a single $y$ embedding per known attribute (dim = 512) and the parameters of $E_c$ (dim = 64) and $G$. We set the learning rate of the latent codes to 0.01, the learning rate of the generator to 0.001 and the learning rate of the encoder to 0.0001. The penalty of the uncorrelated bottleneck is set to $\lambda_{b} = 0.001$. We train the disentanglement stage for 200 epochs. For each mini-batch, we update the parameters of the models and the latent codes with a single gradient step each. In the synthesis stage, we add two encoders to infer the latent codes learned in the disentanglement stage directly from an image. We also optimize a discriminator in an end-to-end manner to increase the perceptual fidelity of the images. This stage is trained for 100 epochs and the learning rate for all the parameters is set to 0.0001.

\paragraph{Baseline models} For the evaluation of competing methods, we use the following official publicly available pretrained models: Lifespan \cite{orel2020lifespan} and SAM \cite{alaluf2021sam} (for age editing on FFHQ), StarGAN-v2 (for AFHQ and CelebA-HQ) and StyleGAN (for mGANprior on CelebA-HQ). We train the rest of the baselines using the official repositories of their authors and make an effort to select the best configurations available for the target resolution (for example, FUNIT trained by us for AFHQ achieves similar results to the public StarGAN-v2 which was known as the SOTA on this benchmark).

\paragraph{Evaluation Protocol}
We assess the disentanglement at two levels: the learned representations and the generated images. At the representation level, we follow the protocol in LORD \cite{gabbay2019demystifying} and train a two-layer multi-layer perceptron to classify the labeled attributes from the learned uncorrelated codes (lower accuracy indicates better disentanglement). In CelebA, where annotations of part of the uncorrelated attributes are available (for evaluation only) such as 68-facial landmark locations, we train a linear regression model to locate the landmarks given the learned identity codes (higher error indicates better disentanglement). At the image level, we follow StarGAN-v2 and translate all images in the test set to each of the other domains multiple times, borrowing correlated attribute codes from random reference images in the target domain. We then train a classifier to classify the domain of the source image from the translated image. A lower accuracy indicates better disentanglement as the source domain does not leak into the translated image. We also compute FID \cite{heusel2017fid} in a conditional manner to measure the discrepancy between the distribution of images in each target domain and the corresponding translations generated by the models. A lower FID score indicates that the translations are more reliable and better fit to the target domain. FID between real train and test images of the same domain forms the optimal score for this metric. In order to assess the diversity of translation results, we measure the perceptual pairwise distances using LPIPS \cite{zhang2018unreasonable} between all translations of the same input image. Higher average distances indicate greater diversity in image translation. In cases where external annotation methods are available (for evaluation only), such as face recognition \cite{cao2018vggface2} and head pose \cite{headpose} and landmark detection for CelebA, we further measure the similarity of the identity of the generated face and the reference, as well as expression (represented by landmarks) and head pose errors. To validate that the head pose and expression are distributed evenly across identities, we use landmark annotations together with pose-related attributes from CelebA (Open Mouth and Smiling) and train a classifier to infer the identity. The accuracy of this classifier (0.001) forms the optimal result for the representation metric.
%While we do not have ground truth for translation in CelebA as well, we present a reconstruction error (LPIPS similarity) by inferring the identity representation from one image and reconstructing a second of the same person by extracting the content representation from an image of a different identity which has the closest pose (nearest neighbour in the 68 facial-landmarks space).
For translating males to females on CelebA-HQ, we measure the accuracy of fooling a target classifier trained on real images, as well as FID to evaluate how the images fit the target domain.

\subsection{Datasets}
\textbf{FFHQ \cite{karras2019style}} $70,000$ high-quality images containing considerable variation in terms of age, ethnicity and image background. We use the images at $256 \times 256$ resolution. FFHQ-Aging \cite{orel2020lifespan} provides age labels for these images.

\textbf{AFHQ \cite{choi2019stargan}} $15,000$ high quality images categorized into three domains: cat, dog and wildlife. We follow the protocol used in StarGAN-v2 and use the images at $256 \times 256$ resolution, holding out 500 images from each domain for testing.

\textbf{CelebA \cite{liu2015faceattributes}}
202,599 images of 10,177 celebrities. We designate the person identity as class. We crop the images to $128 \times 128$ and use 9,177 classes for training and 1,000 for testing.

\textbf{CelebA-HQ \cite{karras2018progressive}} 30,000 high quality images from CelebA. We set the gender as class. We resize the images to $256 \times 256$ and leave 1,000 images from each class for testing. The masks provided in CelebAMask-HQ \cite{CelebAMask-HQ} are used to disentangle the correlated hairstyle.

\textbf{Edges2Shoes \cite{yu2014edges2shoes}} A collection of 50,000 shoe images and their edge maps.

\paragraph{Training resources} Training each of the models presented in this paper takes approximately 3 days for 256x256 resolution on a single NVIDIA RTX-2080 TI.

\subsection{Additional results}
\label{app:additional_results}

\begin{figure*}[t]
\begin{center}
\begin{tabular}{@{\hskip0pt}c@{\hskip3pt}c@{\hskip2pt}c@{\hskip0pt}c@{\hskip0pt}c@{\hskip0pt}c@{\hskip0pt}c@{\hskip0pt}c@{\hskip0pt}c@{\hskip0pt}c}

& Original & [0-9] & [10-19] & [20-29] & [30-39] & [40-49] & [50-59] & [60-69] & [70-79] \\

\begin{turn}{90} Lifespan \cite{orel2020lifespan} \end{turn} &
\includegraphics[width=0.105\linewidth]{figures/aging/e/lifespan/0-0.jpg} &
\includegraphics[width=0.105\linewidth]{figures/aging/e/lifespan/0-1.jpg} &
\includegraphics[width=0.105\linewidth]{figures/aging/e/lifespan/0-2.jpg} &
\includegraphics[width=0.105\linewidth]{figures/aging/e/lifespan/0-3.jpg} &
\includegraphics[width=0.105\linewidth]{figures/aging/e/lifespan/0-4.jpg} &
\includegraphics[width=0.105\linewidth]{figures/aging/e/lifespan/0-5.jpg} &
\includegraphics[width=0.105\linewidth]{figures/aging/e/lifespan/0-6.jpg} &
\includegraphics[width=0.105\linewidth]{figures/aging/e/lifespan/0-7.jpg} &
~ \\

\begin{turn}{90} ~~~SAM \cite{alaluf2021sam} \end{turn} &
\includegraphics[width=0.105\linewidth]{figures/aging/e/sam/0-0.jpg} &
\includegraphics[width=0.105\linewidth]{figures/aging/e/sam/0-1.jpg} &
\includegraphics[width=0.105\linewidth]{figures/aging/e/sam/0-2.jpg} &
\includegraphics[width=0.105\linewidth]{figures/aging/e/sam/0-3.jpg} &
\includegraphics[width=0.105\linewidth]{figures/aging/e/sam/0-4.jpg} &
\includegraphics[width=0.105\linewidth]{figures/aging/e/sam/0-5.jpg} &
\includegraphics[width=0.105\linewidth]{figures/aging/e/sam/0-6.jpg} &
\includegraphics[width=0.105\linewidth]{figures/aging/e/sam/0-7.jpg} &
\includegraphics[width=0.105\linewidth]{figures/aging/e/sam/0-8.jpg} \\

\begin{turn}{90} ~~~~~~\textbf{Ours} \end{turn} &
\includegraphics[width=0.105\linewidth]{figures/aging/e/ours/0-0.jpg} &
\includegraphics[width=0.105\linewidth]{figures/aging/e/ours/0-1.jpg} &
\includegraphics[width=0.105\linewidth]{figures/aging/e/ours/0-2.jpg} &
\includegraphics[width=0.105\linewidth]{figures/aging/e/ours/0-3.jpg} &
\includegraphics[width=0.105\linewidth]{figures/aging/e/ours/0-4.jpg} &
\includegraphics[width=0.105\linewidth]{figures/aging/e/ours/0-5.jpg} &
\includegraphics[width=0.105\linewidth]{figures/aging/e/ours/0-6.jpg} &
\includegraphics[width=0.105\linewidth]{figures/aging/e/ours/0-7.jpg} &
\includegraphics[width=0.105\linewidth]{figures/aging/e/ours/0-8.jpg} \\

\begin{turn}{90} Lifespan \cite{orel2020lifespan} \end{turn} &
\includegraphics[width=0.105\linewidth]{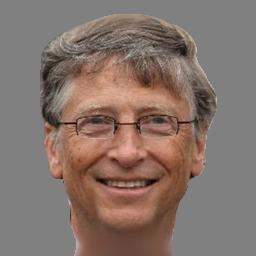} &
\includegraphics[width=0.105\linewidth]{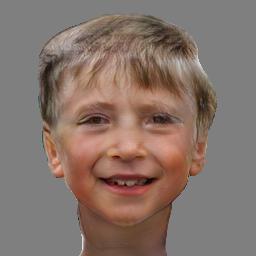} &
\includegraphics[width=0.105\linewidth]{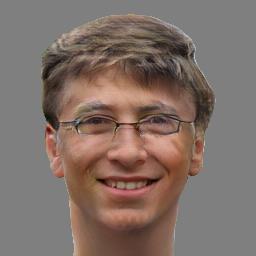} &
\includegraphics[width=0.105\linewidth]{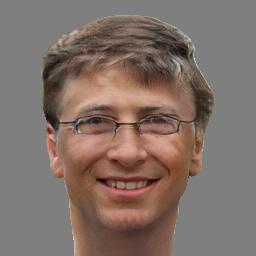} &
\includegraphics[width=0.105\linewidth]{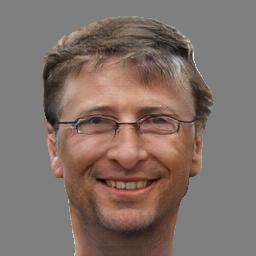} &
\includegraphics[width=0.105\linewidth]{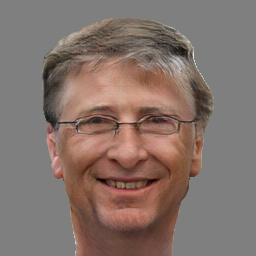} &
\includegraphics[width=0.105\linewidth]{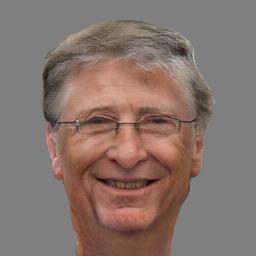} &
\includegraphics[width=0.105\linewidth]{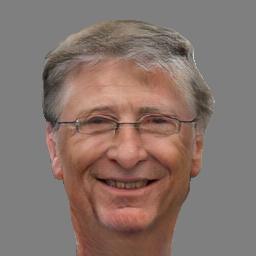} &
~ \\

\begin{turn}{90} ~~~SAM \cite{alaluf2021sam} \end{turn} &
\includegraphics[width=0.105\linewidth]{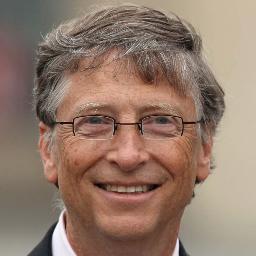} &
\includegraphics[width=0.105\linewidth]{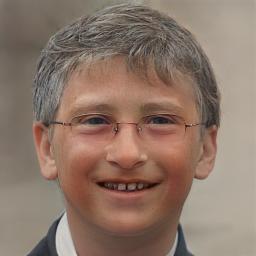} &
\includegraphics[width=0.105\linewidth]{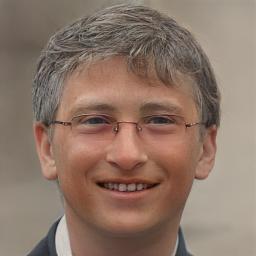} &
\includegraphics[width=0.105\linewidth]{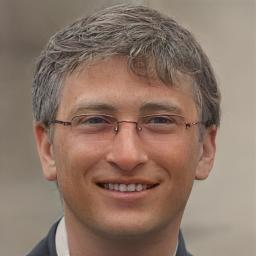} &
\includegraphics[width=0.105\linewidth]{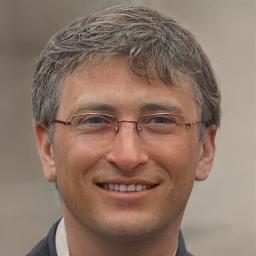} &
\includegraphics[width=0.105\linewidth]{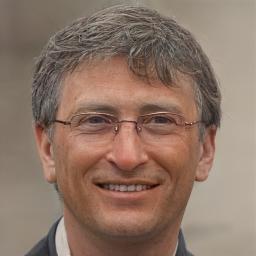} &
\includegraphics[width=0.105\linewidth]{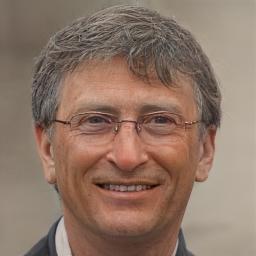} &
\includegraphics[width=0.105\linewidth]{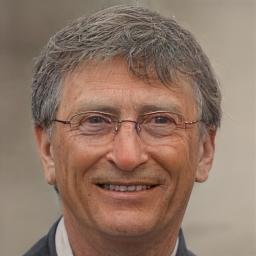} &
\includegraphics[width=0.105\linewidth]{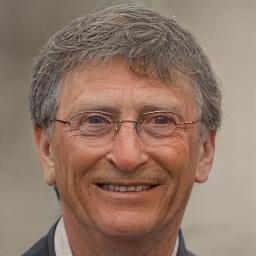} \\

\begin{turn}{90} ~~~~~~\textbf{Ours} \end{turn} &
\includegraphics[width=0.105\linewidth]{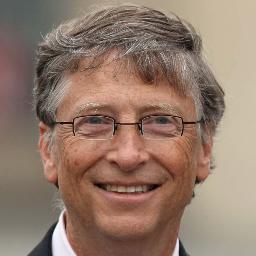} &
\includegraphics[width=0.105\linewidth]{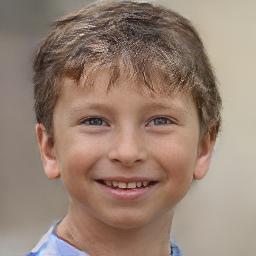} &
\includegraphics[width=0.105\linewidth]{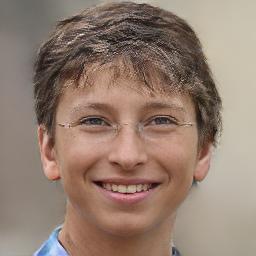} &
\includegraphics[width=0.105\linewidth]{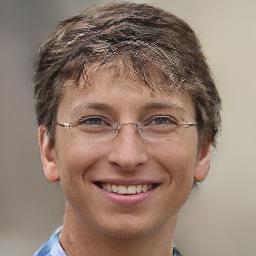} &
\includegraphics[width=0.105\linewidth]{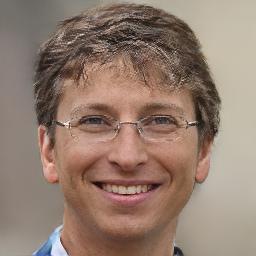} &
\includegraphics[width=0.105\linewidth]{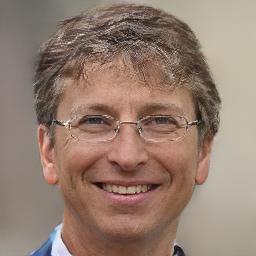} &
\includegraphics[width=0.105\linewidth]{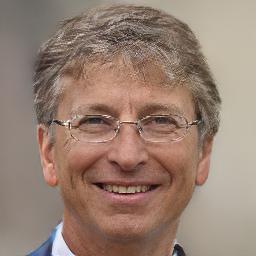} &
\includegraphics[width=0.105\linewidth]{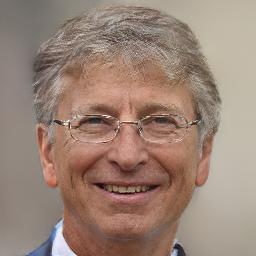} &
\includegraphics[width=0.105\linewidth]{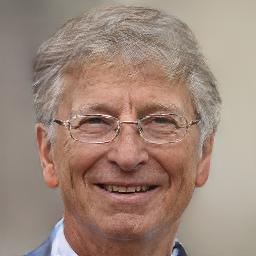} \\

\begin{turn}{90} Lifespan \cite{orel2020lifespan} \end{turn} &
\includegraphics[width=0.105\linewidth]{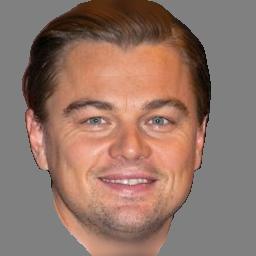} &
\includegraphics[width=0.105\linewidth]{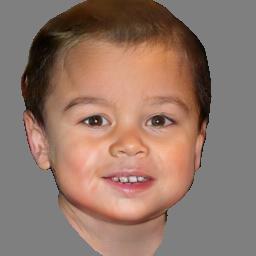} &
\includegraphics[width=0.105\linewidth]{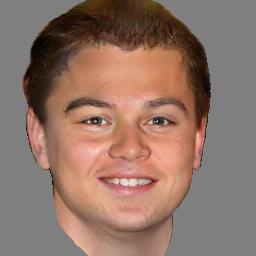} &
\includegraphics[width=0.105\linewidth]{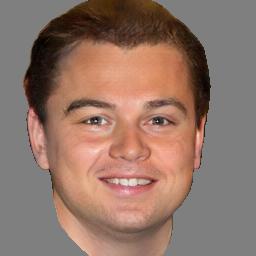} &
\includegraphics[width=0.105\linewidth]{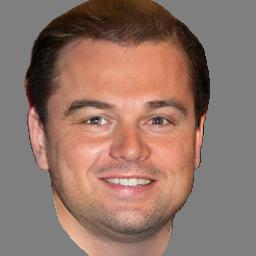} &
\includegraphics[width=0.105\linewidth]{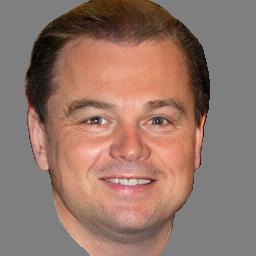} &
\includegraphics[width=0.105\linewidth]{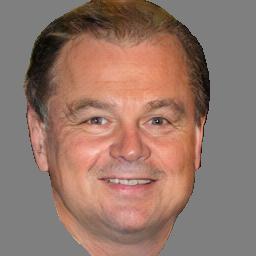} &
\includegraphics[width=0.105\linewidth]{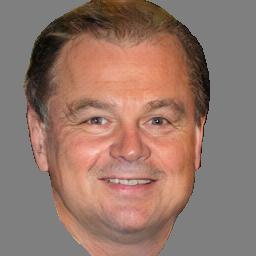} &
~ \\

\begin{turn}{90} ~~~SAM \cite{alaluf2021sam} \end{turn} &
\includegraphics[width=0.105\linewidth]{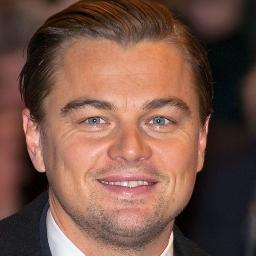} &
\includegraphics[width=0.105\linewidth]{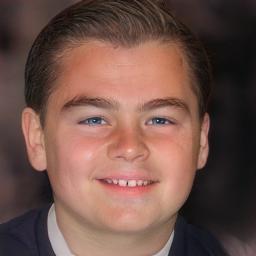} &
\includegraphics[width=0.105\linewidth]{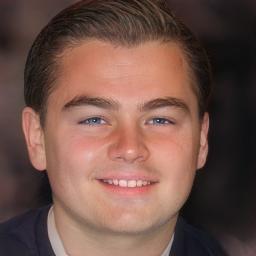} &
\includegraphics[width=0.105\linewidth]{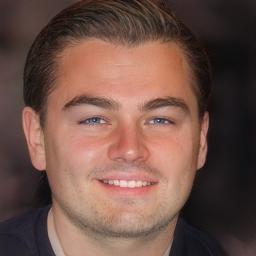} &
\includegraphics[width=0.105\linewidth]{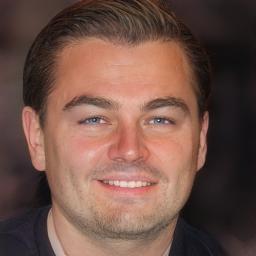} &
\includegraphics[width=0.105\linewidth]{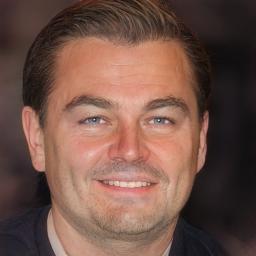} &
\includegraphics[width=0.105\linewidth]{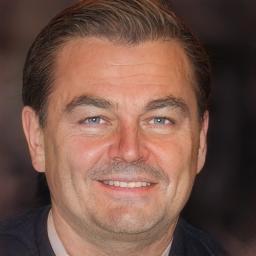} &
\includegraphics[width=0.105\linewidth]{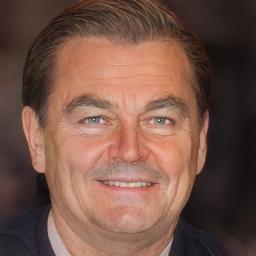} &
\includegraphics[width=0.105\linewidth]{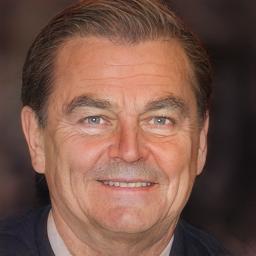} \\

\begin{turn}{90} ~~~~~~\textbf{Ours} \end{turn} &
\includegraphics[width=0.105\linewidth]{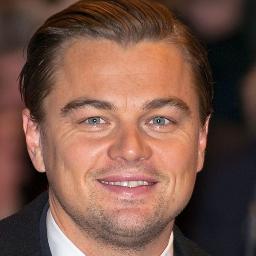} &
\includegraphics[width=0.105\linewidth]{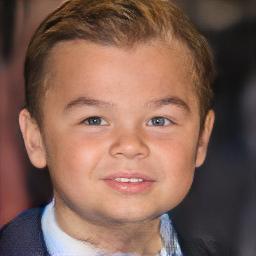} &
\includegraphics[width=0.105\linewidth]{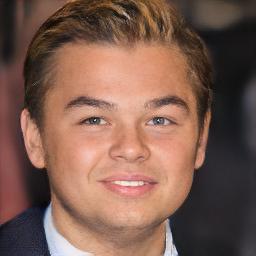} &
\includegraphics[width=0.105\linewidth]{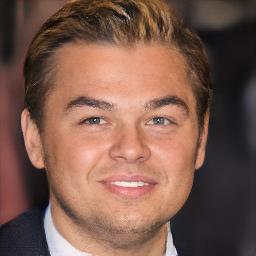} &
\includegraphics[width=0.105\linewidth]{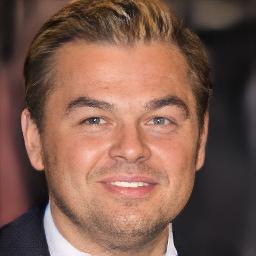} &
\includegraphics[width=0.105\linewidth]{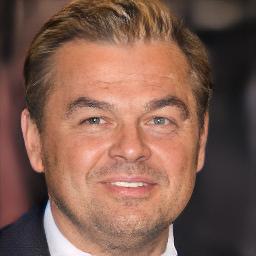} &
\includegraphics[width=0.105\linewidth]{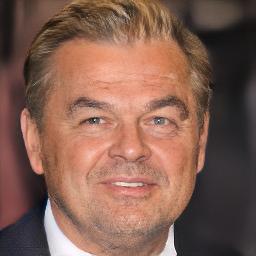} &
\includegraphics[width=0.105\linewidth]{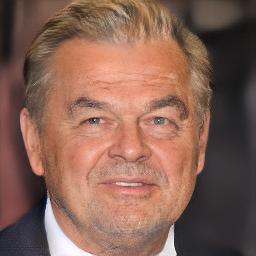} &
\includegraphics[width=0.105\linewidth]{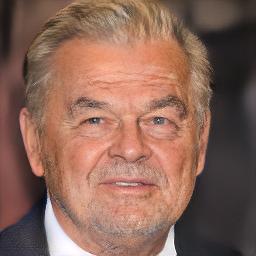} \\

\end{tabular}
\end{center}
\caption{An extended comparison of disentangling age and unlabeled attributes (e.g. identity). Our general method performs better on age-editing than the two task-specific baselines which rely on a supervised identity loss.}
\label{fig:aging_more_qualitative_a}
\end{figure*}

\begin{figure*}[t]
\begin{center}
\begin{tabular}{@{\hskip0pt}c@{\hskip3pt}c@{\hskip2pt}c@{\hskip0pt}c@{\hskip0pt}c@{\hskip0pt}c@{\hskip0pt}c@{\hskip0pt}c@{\hskip0pt}c@{\hskip0pt}c}

& Original & [0-9] & [10-19] & [20-29] & [30-39] & [40-49] & [50-59] & [60-69] & [70-79] \\

\begin{turn}{90} Lifespan \cite{orel2020lifespan} \end{turn} &
\includegraphics[width=0.105\linewidth]{figures/aging/h/lifespan/0-0.jpg} &
\includegraphics[width=0.105\linewidth]{figures/aging/h/lifespan/0-1.jpg} &
\includegraphics[width=0.105\linewidth]{figures/aging/h/lifespan/0-2.jpg} &
\includegraphics[width=0.105\linewidth]{figures/aging/h/lifespan/0-3.jpg} &
\includegraphics[width=0.105\linewidth]{figures/aging/h/lifespan/0-4.jpg} &
\includegraphics[width=0.105\linewidth]{figures/aging/h/lifespan/0-5.jpg} &
\includegraphics[width=0.105\linewidth]{figures/aging/h/lifespan/0-6.jpg} &
\includegraphics[width=0.105\linewidth]{figures/aging/h/lifespan/0-7.jpg} &
~ \\

\begin{turn}{90} ~~~SAM \cite{alaluf2021sam} \end{turn} &
\includegraphics[width=0.105\linewidth]{figures/aging/h/sam/0-0.jpg} &
\includegraphics[width=0.105\linewidth]{figures/aging/h/sam/0-1.jpg} &
\includegraphics[width=0.105\linewidth]{figures/aging/h/sam/0-2.jpg} &
\includegraphics[width=0.105\linewidth]{figures/aging/h/sam/0-3.jpg} &
\includegraphics[width=0.105\linewidth]{figures/aging/h/sam/0-4.jpg} &
\includegraphics[width=0.105\linewidth]{figures/aging/h/sam/0-5.jpg} &
\includegraphics[width=0.105\linewidth]{figures/aging/h/sam/0-6.jpg} &
\includegraphics[width=0.105\linewidth]{figures/aging/h/sam/0-7.jpg} &
\includegraphics[width=0.105\linewidth]{figures/aging/h/sam/0-8.jpg} \\

\begin{turn}{90} ~~~~~~\textbf{Ours} \end{turn} &
\includegraphics[width=0.105\linewidth]{figures/aging/h/ours/0-0.jpg} &
\includegraphics[width=0.105\linewidth]{figures/aging/h/ours/0-1.jpg} &
\includegraphics[width=0.105\linewidth]{figures/aging/h/ours/0-2.jpg} &
\includegraphics[width=0.105\linewidth]{figures/aging/h/ours/0-3.jpg} &
\includegraphics[width=0.105\linewidth]{figures/aging/h/ours/0-4.jpg} &
\includegraphics[width=0.105\linewidth]{figures/aging/h/ours/0-5.jpg} &
\includegraphics[width=0.105\linewidth]{figures/aging/h/ours/0-6.jpg} &
\includegraphics[width=0.105\linewidth]{figures/aging/h/ours/0-7.jpg} &
\includegraphics[width=0.105\linewidth]{figures/aging/h/ours/0-8.jpg} \\

\begin{turn}{90} Lifespan \cite{orel2020lifespan} \end{turn} &
\includegraphics[width=0.105\linewidth]{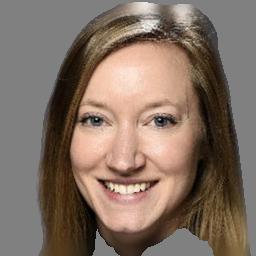} &
\includegraphics[width=0.105\linewidth]{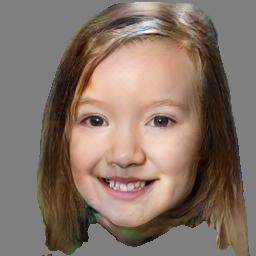} &
\includegraphics[width=0.105\linewidth]{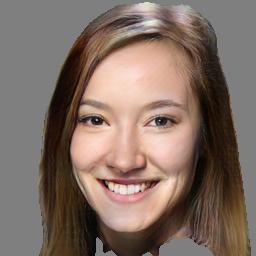} &
\includegraphics[width=0.105\linewidth]{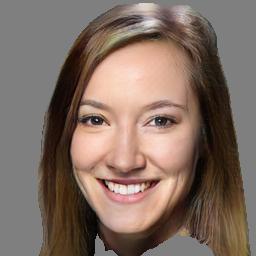} &
\includegraphics[width=0.105\linewidth]{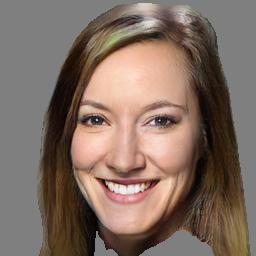} &
\includegraphics[width=0.105\linewidth]{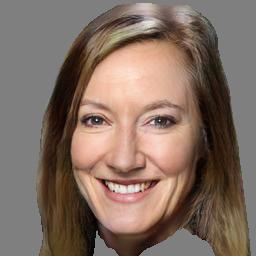} &
\includegraphics[width=0.105\linewidth]{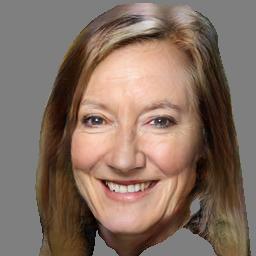} &
\includegraphics[width=0.105\linewidth]{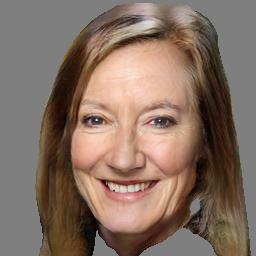} &
~ \\

\begin{turn}{90} ~~~SAM \cite{alaluf2021sam} \end{turn} &
\includegraphics[width=0.105\linewidth]{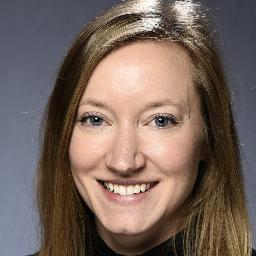} &
\includegraphics[width=0.105\linewidth]{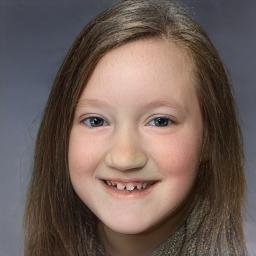} &
\includegraphics[width=0.105\linewidth]{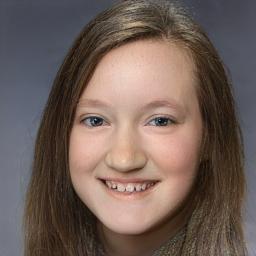} &
\includegraphics[width=0.105\linewidth]{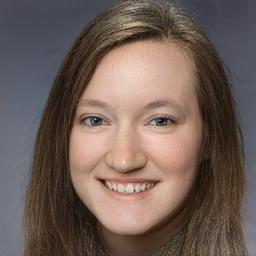} &
\includegraphics[width=0.105\linewidth]{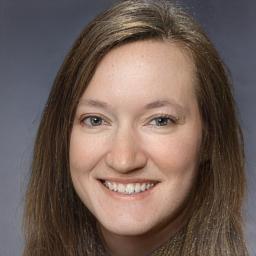} &
\includegraphics[width=0.105\linewidth]{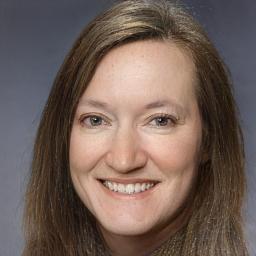} &
\includegraphics[width=0.105\linewidth]{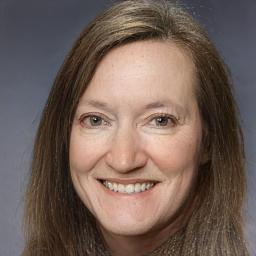} &
\includegraphics[width=0.105\linewidth]{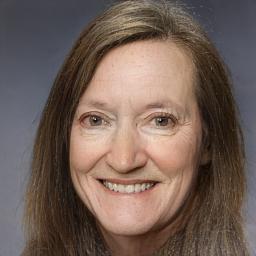} &
\includegraphics[width=0.105\linewidth]{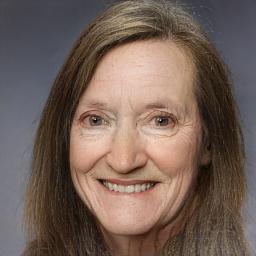} \\

\begin{turn}{90} ~~~~~~\textbf{Ours} \end{turn} &
\includegraphics[width=0.105\linewidth]{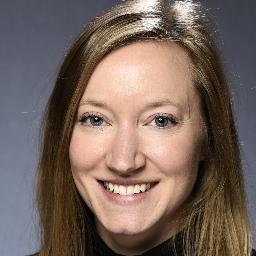} &
\includegraphics[width=0.105\linewidth]{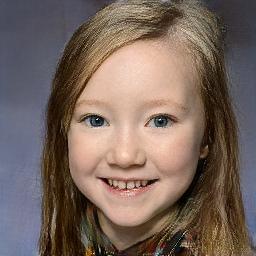} &
\includegraphics[width=0.105\linewidth]{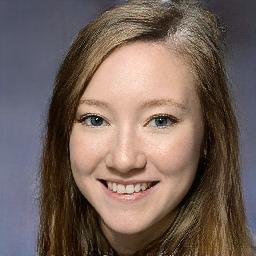} &
\includegraphics[width=0.105\linewidth]{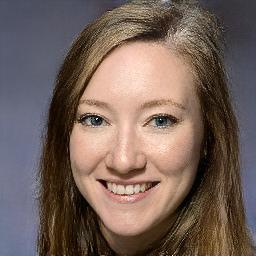} &
\includegraphics[width=0.105\linewidth]{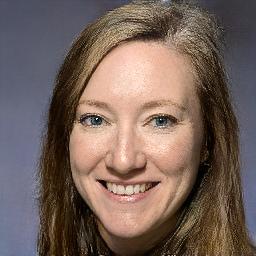} &
\includegraphics[width=0.105\linewidth]{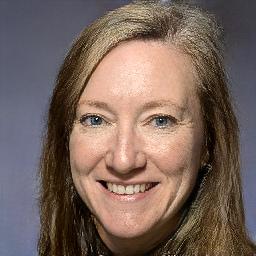} &
\includegraphics[width=0.105\linewidth]{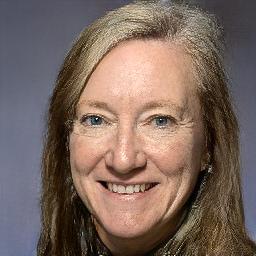} &
\includegraphics[width=0.105\linewidth]{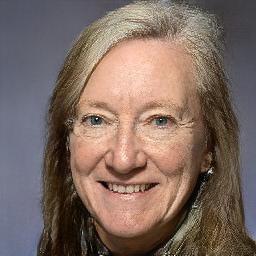} &
\includegraphics[width=0.105\linewidth]{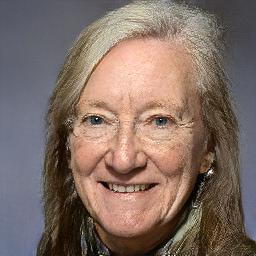} \\

\begin{turn}{90} Lifespan \cite{orel2020lifespan} \end{turn} &
\includegraphics[width=0.105\linewidth]{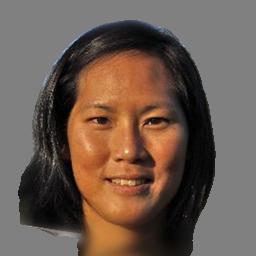} &
\includegraphics[width=0.105\linewidth]{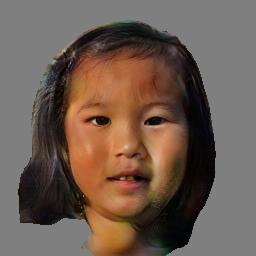} &
\includegraphics[width=0.105\linewidth]{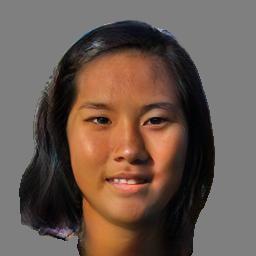} &
\includegraphics[width=0.105\linewidth]{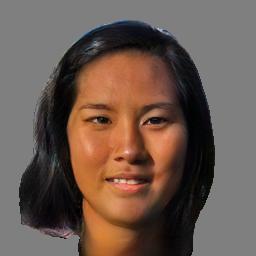} &
\includegraphics[width=0.105\linewidth]{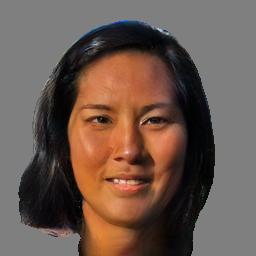} &
\includegraphics[width=0.105\linewidth]{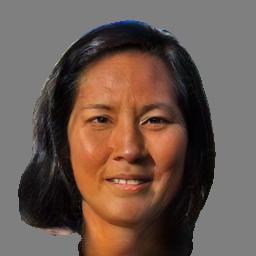} &
\includegraphics[width=0.105\linewidth]{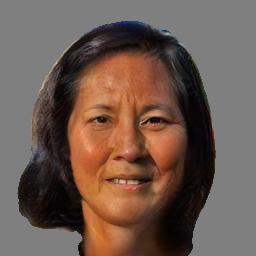} &
\includegraphics[width=0.105\linewidth]{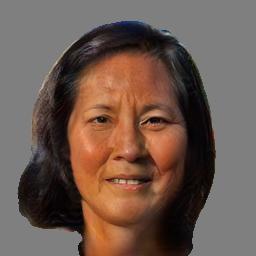} &
~ \\

\begin{turn}{90} ~~~SAM \cite{alaluf2021sam} \end{turn} &
\includegraphics[width=0.105\linewidth]{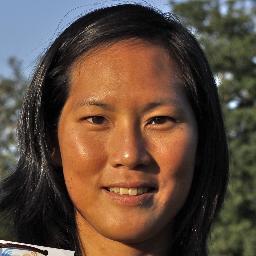} &
\includegraphics[width=0.105\linewidth]{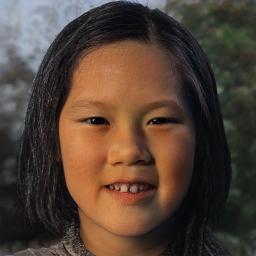} &
\includegraphics[width=0.105\linewidth]{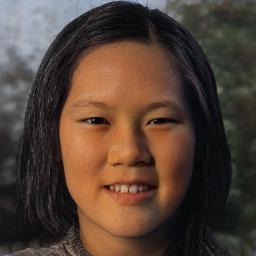} &
\includegraphics[width=0.105\linewidth]{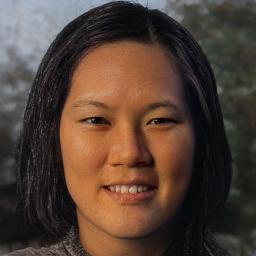} &
\includegraphics[width=0.105\linewidth]{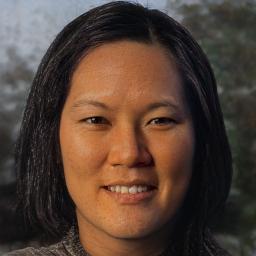} &
\includegraphics[width=0.105\linewidth]{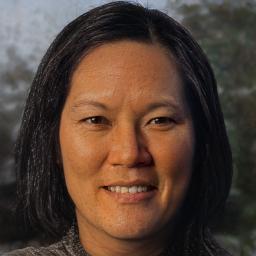} &
\includegraphics[width=0.105\linewidth]{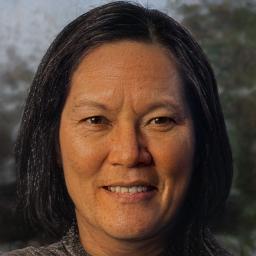} &
\includegraphics[width=0.105\linewidth]{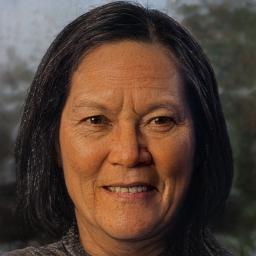} &
\includegraphics[width=0.105\linewidth]{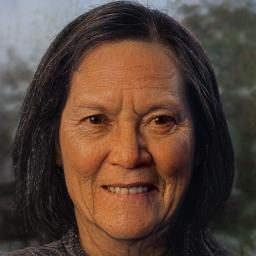} \\

\begin{turn}{90} ~~~~~~\textbf{Ours} \end{turn} &
\includegraphics[width=0.105\linewidth]{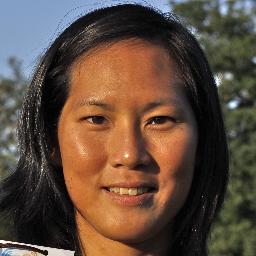} &
\includegraphics[width=0.105\linewidth]{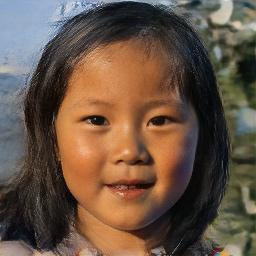} &
\includegraphics[width=0.105\linewidth]{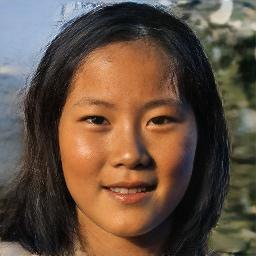} &
\includegraphics[width=0.105\linewidth]{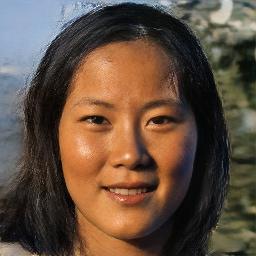} &
\includegraphics[width=0.105\linewidth]{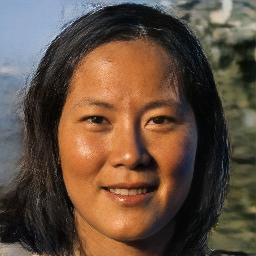} &
\includegraphics[width=0.105\linewidth]{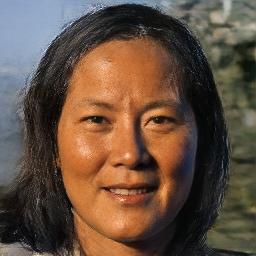} &
\includegraphics[width=0.105\linewidth]{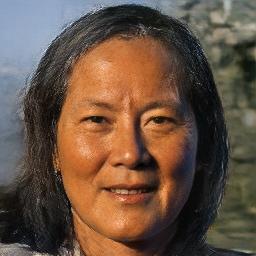} &
\includegraphics[width=0.105\linewidth]{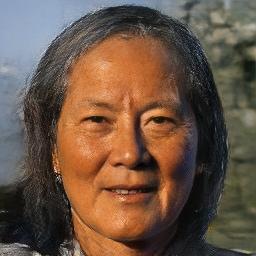} &
\includegraphics[width=0.105\linewidth]{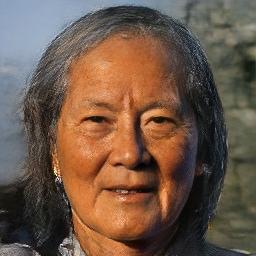} \\

\end{tabular}
\end{center}
\caption{More qualitative results of facial age editing. Our model makes more significant changes (e.g. hair color) while preserving the identity better than the baselines.}
\label{fig:aging_more_qualitative_b}
\end{figure*}

\begin{figure*}[t]
\begin{center}
\begin{tabular}{@{\hskip0pt}c@{\hskip2pt}c@{\hskip0pt}c@{\hskip0pt}c@{\hskip0pt}c@{\hskip0pt}c@{\hskip0pt}c@{\hskip0pt}c@{\hskip0pt}c}

Original & [0-9] & [10-19] & [20-29] & [30-39] & [40-49] & [50-59] & [60-69] & [70-79] \\

\includegraphics[width=0.11\linewidth]{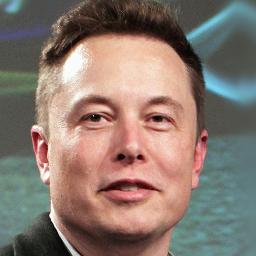} &
\includegraphics[width=0.11\linewidth]{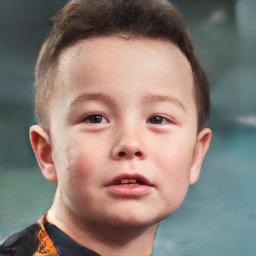} &
\includegraphics[width=0.11\linewidth]{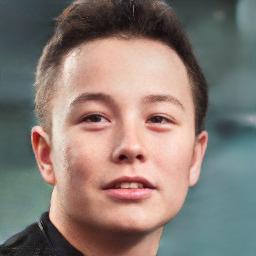} &
\includegraphics[width=0.11\linewidth]{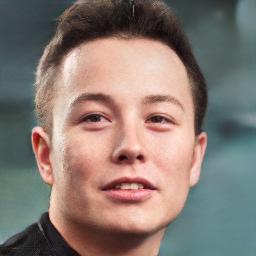} &
\includegraphics[width=0.11\linewidth]{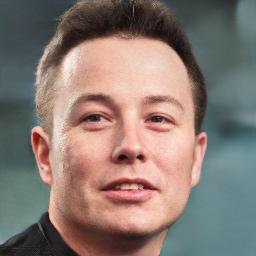} &
\includegraphics[width=0.11\linewidth]{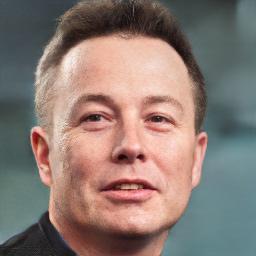} &
\includegraphics[width=0.11\linewidth]{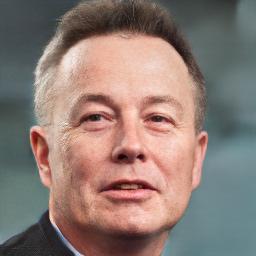} &
\includegraphics[width=0.11\linewidth]{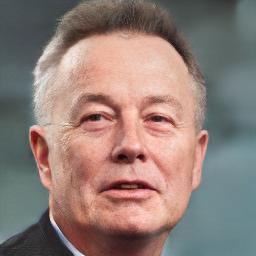} &
\includegraphics[width=0.11\linewidth]{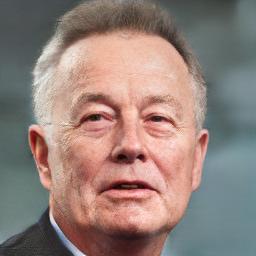} \\

\includegraphics[width=0.11\linewidth]{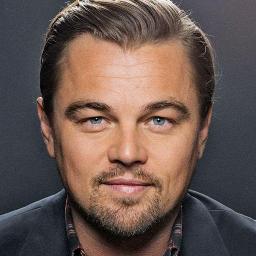} &
\includegraphics[width=0.11\linewidth]{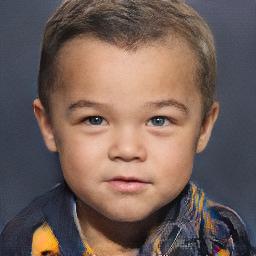} &
\includegraphics[width=0.11\linewidth]{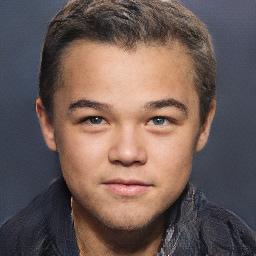} &
\includegraphics[width=0.11\linewidth]{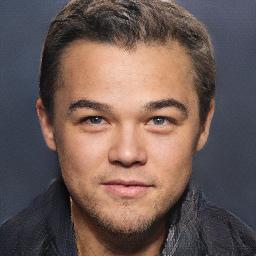} &
\includegraphics[width=0.11\linewidth]{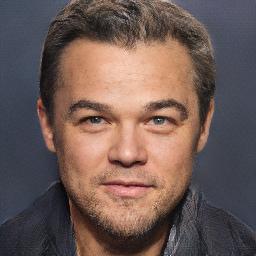} &
\includegraphics[width=0.11\linewidth]{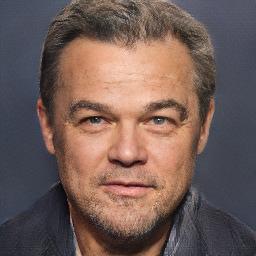} &
\includegraphics[width=0.11\linewidth]{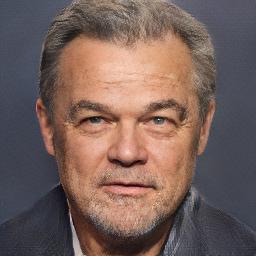} &
\includegraphics[width=0.11\linewidth]{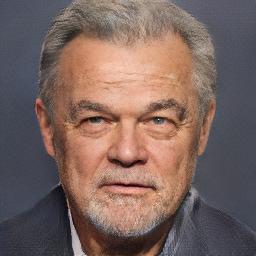} &
\includegraphics[width=0.11\linewidth]{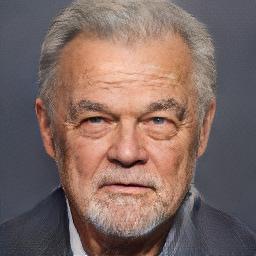} \\

\includegraphics[width=0.11\linewidth]{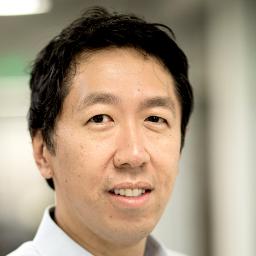} &
\includegraphics[width=0.11\linewidth]{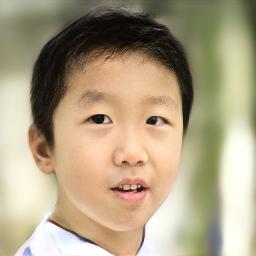} &
\includegraphics[width=0.11\linewidth]{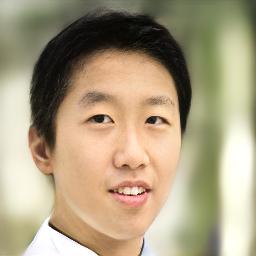} &
\includegraphics[width=0.11\linewidth]{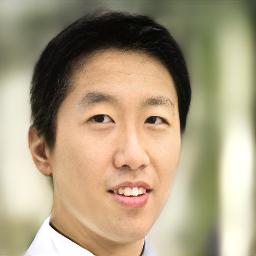} &
\includegraphics[width=0.11\linewidth]{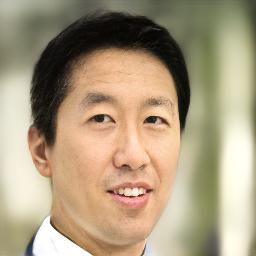} &
\includegraphics[width=0.11\linewidth]{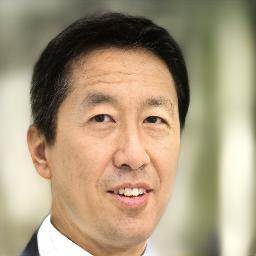} &
\includegraphics[width=0.11\linewidth]{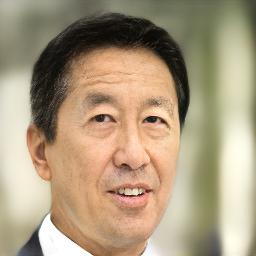} &
\includegraphics[width=0.11\linewidth]{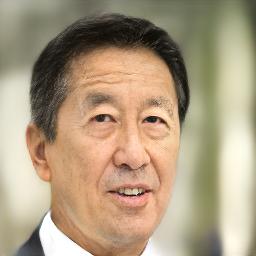} &
\includegraphics[width=0.11\linewidth]{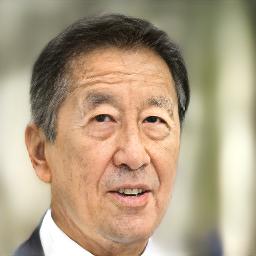} \\

\includegraphics[width=0.11\linewidth]{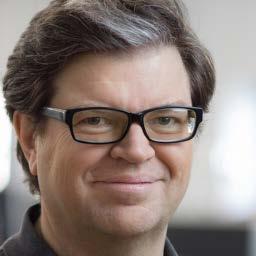} &
\includegraphics[width=0.11\linewidth]{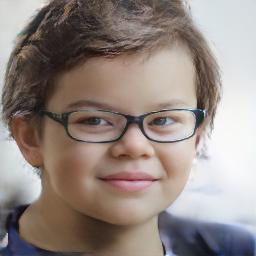} &
\includegraphics[width=0.11\linewidth]{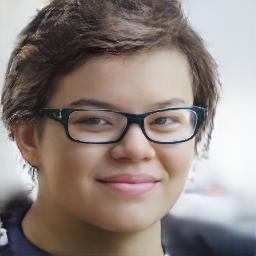} &
\includegraphics[width=0.11\linewidth]{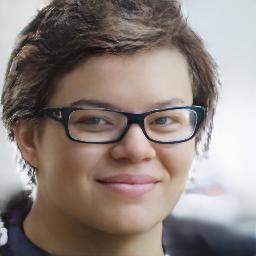} &
\includegraphics[width=0.11\linewidth]{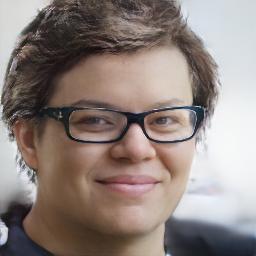} &
\includegraphics[width=0.11\linewidth]{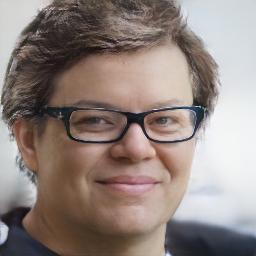} &
\includegraphics[width=0.11\linewidth]{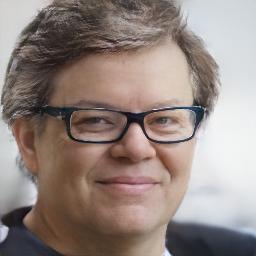} &
\includegraphics[width=0.11\linewidth]{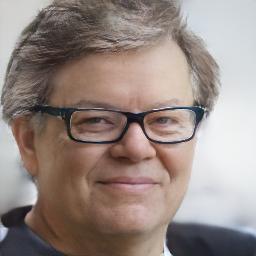} &
\includegraphics[width=0.11\linewidth]{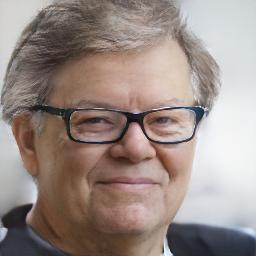} \\

\includegraphics[width=0.11\linewidth]{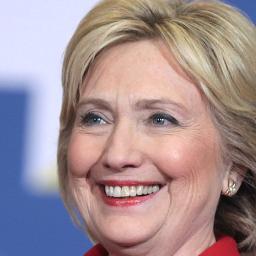} &
\includegraphics[width=0.11\linewidth]{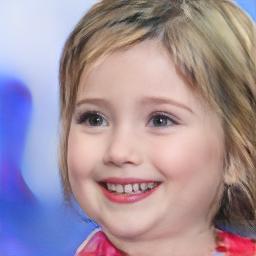} &
\includegraphics[width=0.11\linewidth]{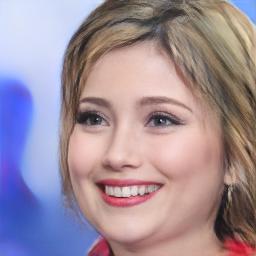} &
\includegraphics[width=0.11\linewidth]{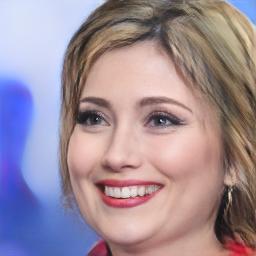} &
\includegraphics[width=0.11\linewidth]{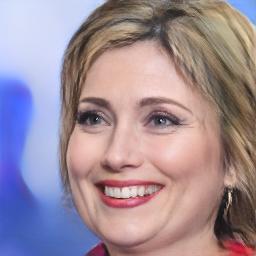} &
\includegraphics[width=0.11\linewidth]{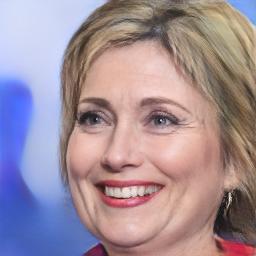} &
\includegraphics[width=0.11\linewidth]{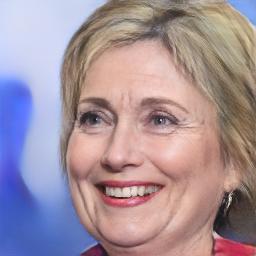} &
\includegraphics[width=0.11\linewidth]{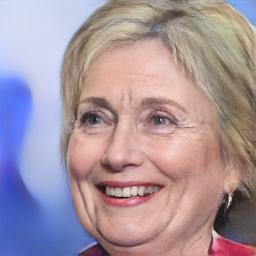} &
\includegraphics[width=0.11\linewidth]{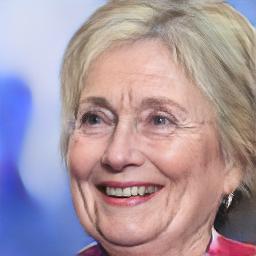} \\

\includegraphics[width=0.11\linewidth]{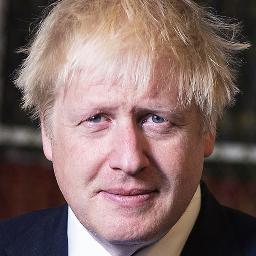} &
\includegraphics[width=0.11\linewidth]{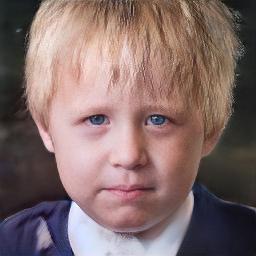} &
\includegraphics[width=0.11\linewidth]{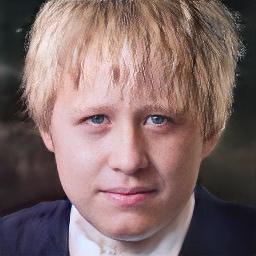} &
\includegraphics[width=0.11\linewidth]{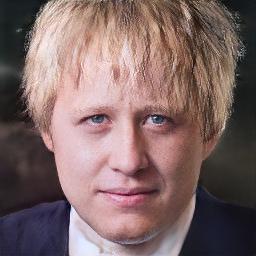} &
\includegraphics[width=0.11\linewidth]{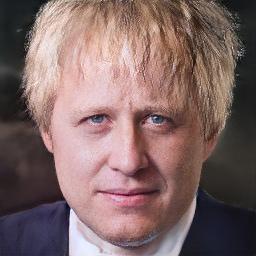} &
\includegraphics[width=0.11\linewidth]{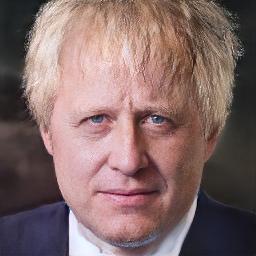} &
\includegraphics[width=0.11\linewidth]{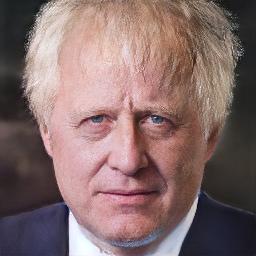} &
\includegraphics[width=0.11\linewidth]{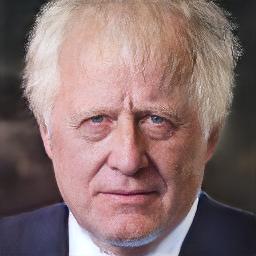} &
\includegraphics[width=0.11\linewidth]{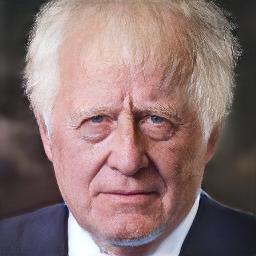} \\

\includegraphics[width=0.11\linewidth]{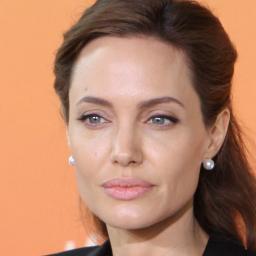} &
\includegraphics[width=0.11\linewidth]{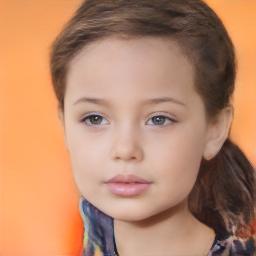} &
\includegraphics[width=0.11\linewidth]{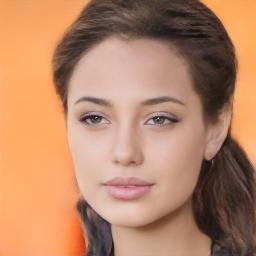} &
\includegraphics[width=0.11\linewidth]{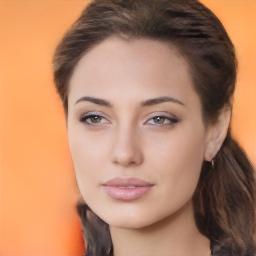} &
\includegraphics[width=0.11\linewidth]{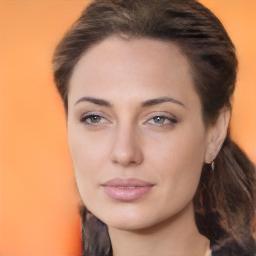} &
\includegraphics[width=0.11\linewidth]{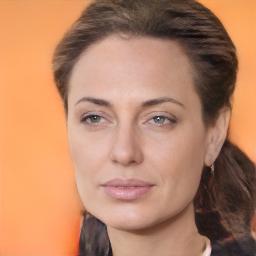} &
\includegraphics[width=0.11\linewidth]{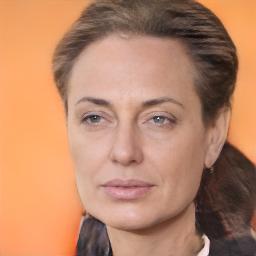} &
\includegraphics[width=0.11\linewidth]{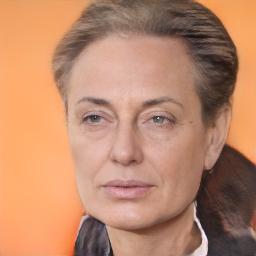} &
\includegraphics[width=0.11\linewidth]{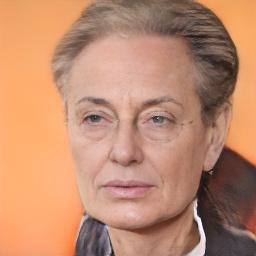} \\

\includegraphics[width=0.11\linewidth]{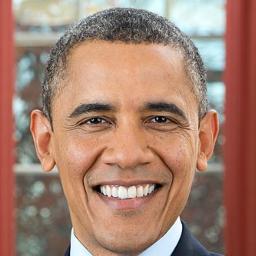} &
\includegraphics[width=0.11\linewidth]{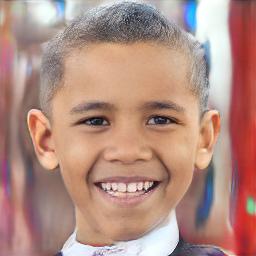} &
\includegraphics[width=0.11\linewidth]{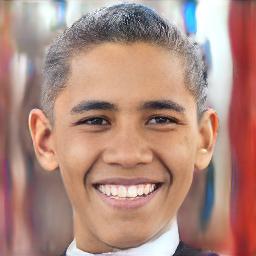} &
\includegraphics[width=0.11\linewidth]{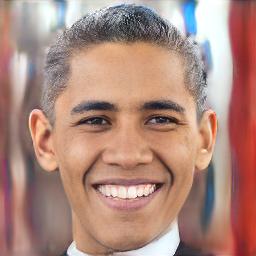} &
\includegraphics[width=0.11\linewidth]{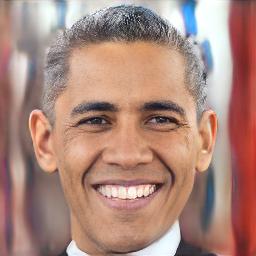} &
\includegraphics[width=0.11\linewidth]{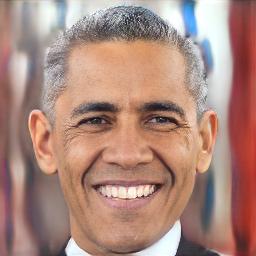} &
\includegraphics[width=0.11\linewidth]{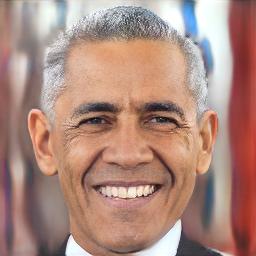} &
\includegraphics[width=0.11\linewidth]{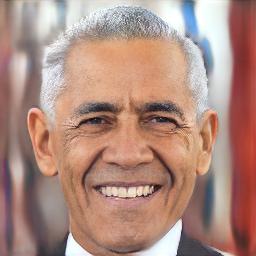} &
\includegraphics[width=0.11\linewidth]{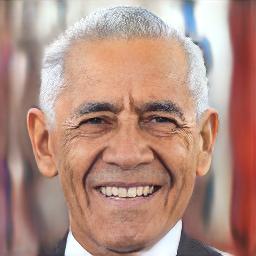} \\

\includegraphics[width=0.11\linewidth]{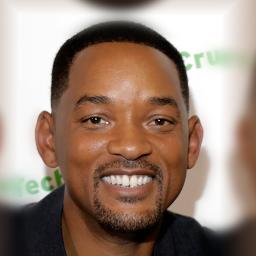} &
\includegraphics[width=0.11\linewidth]{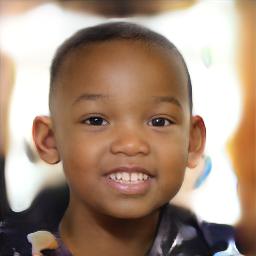} &
\includegraphics[width=0.11\linewidth]{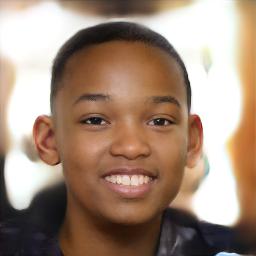} &
\includegraphics[width=0.11\linewidth]{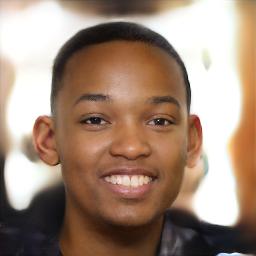} &
\includegraphics[width=0.11\linewidth]{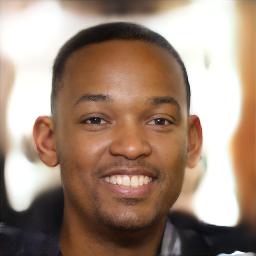} &
\includegraphics[width=0.11\linewidth]{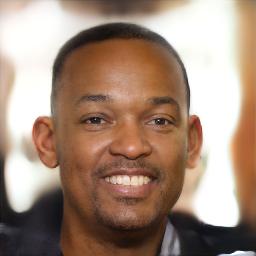} &
\includegraphics[width=0.11\linewidth]{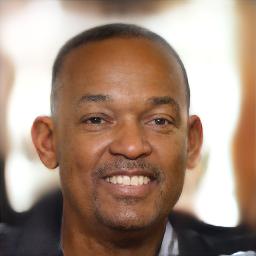} &
\includegraphics[width=0.11\linewidth]{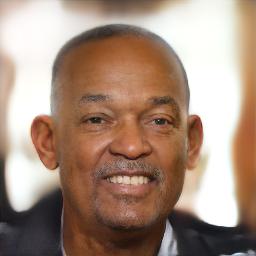} &
\includegraphics[width=0.11\linewidth]{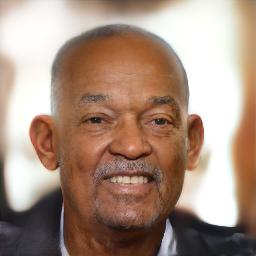} \\

\includegraphics[width=0.11\linewidth]{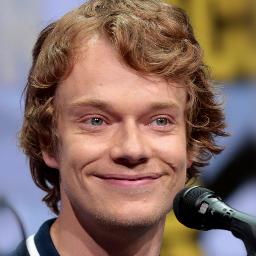} &
\includegraphics[width=0.11\linewidth]{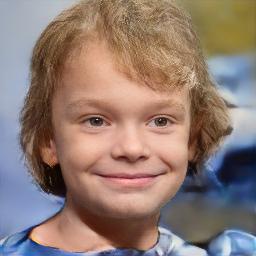} &
\includegraphics[width=0.11\linewidth]{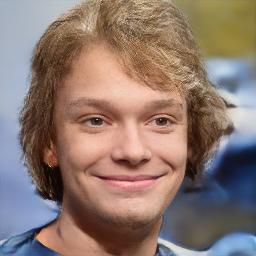} &
\includegraphics[width=0.11\linewidth]{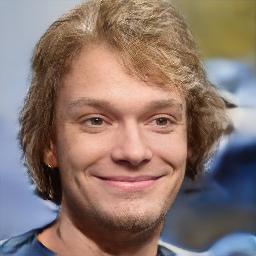} &
\includegraphics[width=0.11\linewidth]{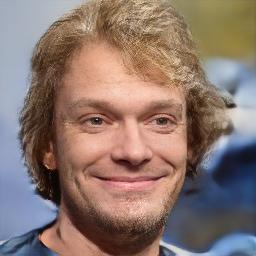} &
\includegraphics[width=0.11\linewidth]{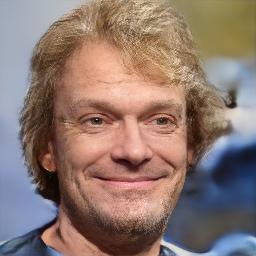} &
\includegraphics[width=0.11\linewidth]{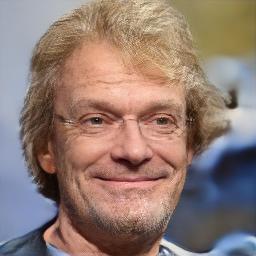} &
\includegraphics[width=0.11\linewidth]{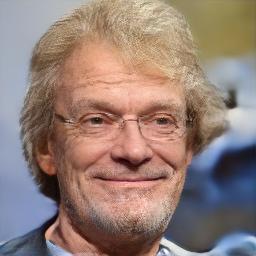} &
\includegraphics[width=0.11\linewidth]{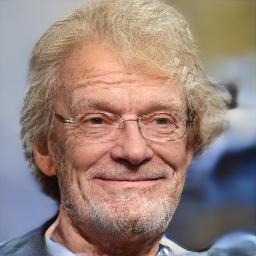} \\

\end{tabular}
\end{center}
\caption{More qualitative results of facial age editing with our model.}
\label{fig:aging_more_qualitative_c}
\end{figure*}

\begin{figure*}[t]
\begin{center}

\begin{tabular}{c@{\hskip2pt}c@{\hskip0pt}c@{\hskip0pt}c@{\hskip0pt}c@{\hskip2pt}c}

~~~~ Appearance \begin{turn}{90} ~~~~~~~~~ Pose \end{turn} & \includegraphics[width=0.15\linewidth]{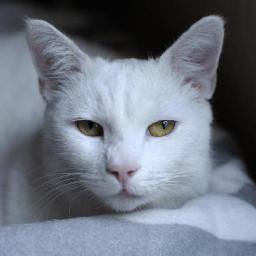} & \includegraphics[width=0.15\linewidth]{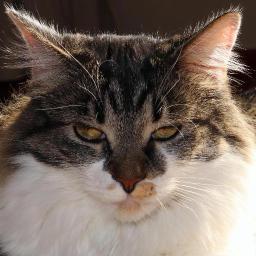} &
\includegraphics[width=0.15\linewidth]{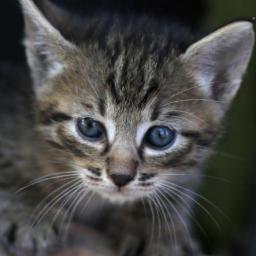} & \includegraphics[width=0.15\linewidth]{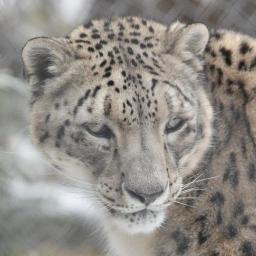} & \\

\includegraphics[width=0.15\linewidth]{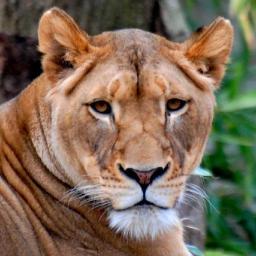} &
\includegraphics[width=0.15\linewidth]{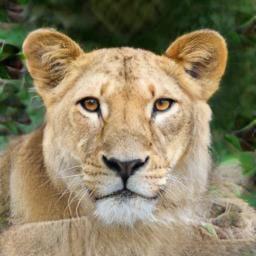} & \includegraphics[width=0.15\linewidth]{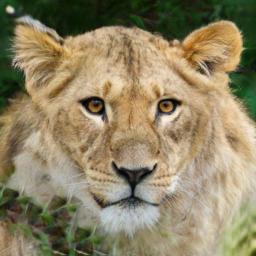} & \includegraphics[width=0.15\linewidth]{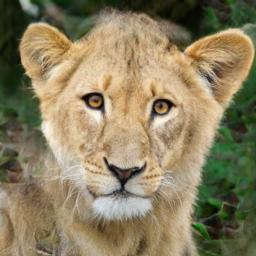} &
\includegraphics[width=0.15\linewidth]{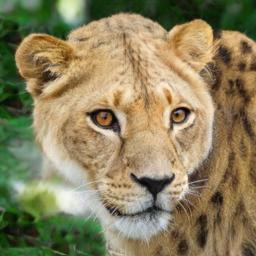} & ~ \rotatebox[origin=r]{270}{FUNIT~~~~~~} \\

& \includegraphics[width=0.15\linewidth]{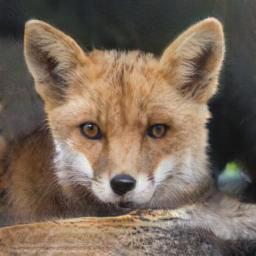} & \includegraphics[width=0.15\linewidth]{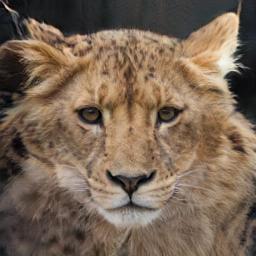} & \includegraphics[width=0.15\linewidth]{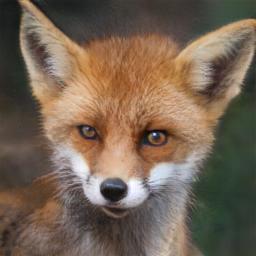} &
\includegraphics[width=0.15\linewidth]{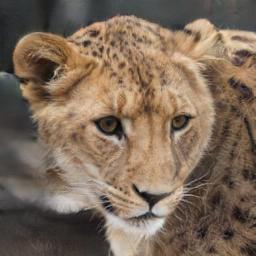} & ~ \rotatebox[origin=r]{270}{StarGAN-v2~~~} \\

& \includegraphics[width=0.15\linewidth]{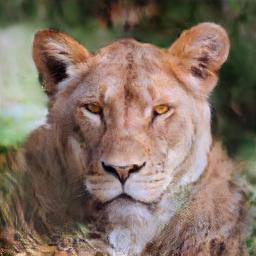} & \includegraphics[width=0.15\linewidth]{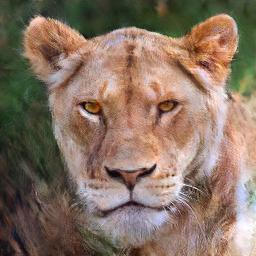} & \includegraphics[width=0.15\linewidth]{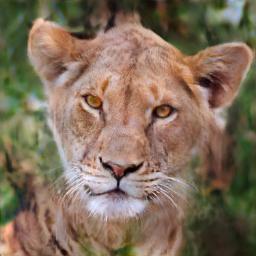} &
\includegraphics[width=0.15\linewidth]{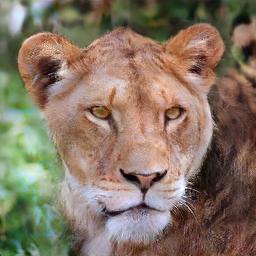} & ~ \rotatebox[origin=r]{270}{\textbf{Ours}~~~~~~~~~}  \\

\includegraphics[width=0.15\linewidth]{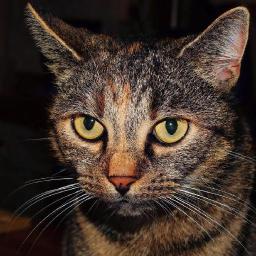} &
\includegraphics[width=0.15\linewidth]{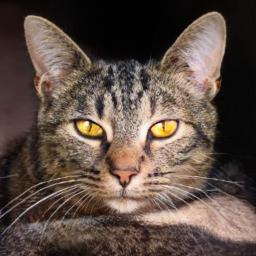} & \includegraphics[width=0.15\linewidth]{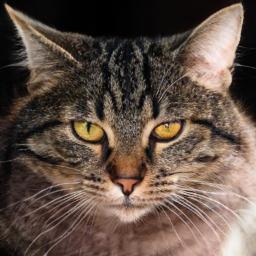} & \includegraphics[width=0.15\linewidth]{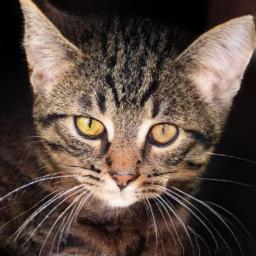} &
\includegraphics[width=0.15\linewidth]{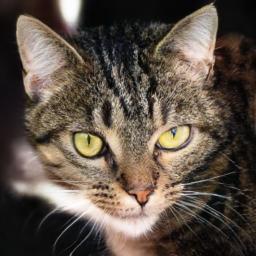} & ~ 
\rotatebox[origin=r]{270}{FUNIT~~~~~~} \\

& \includegraphics[width=0.15\linewidth]{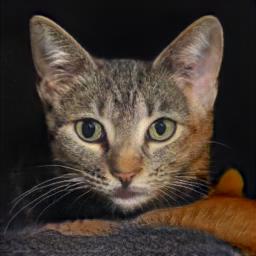} & \includegraphics[width=0.15\linewidth]{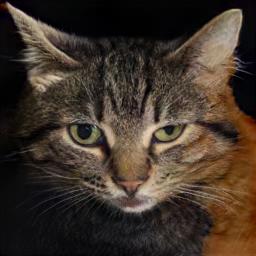} & \includegraphics[width=0.15\linewidth]{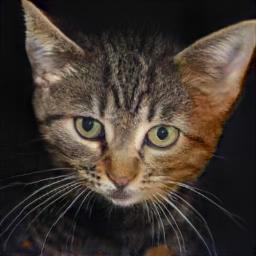} &
\includegraphics[width=0.15\linewidth]{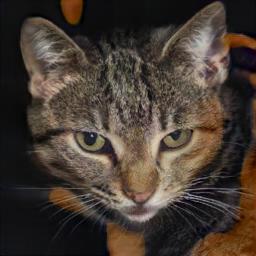} & ~ \rotatebox[origin=r]{270}{StarGAN-v2~~~} \\

& \includegraphics[width=0.15\linewidth]{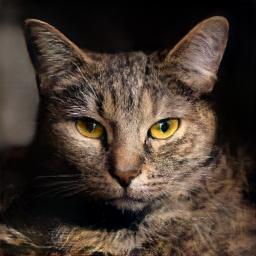} & \includegraphics[width=0.15\linewidth]{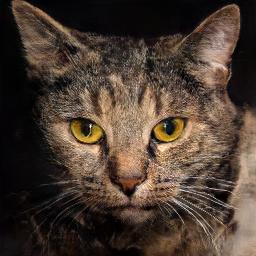} & \includegraphics[width=0.15\linewidth]{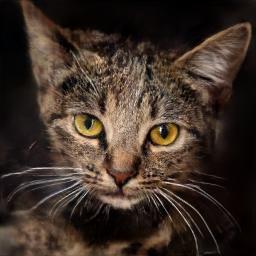} &
\includegraphics[width=0.15\linewidth]{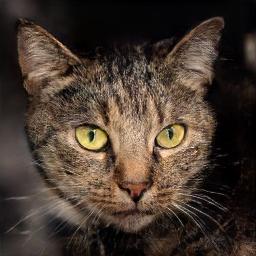} & ~ \rotatebox[origin=r]{270}{\textbf{Ours}~~~~~~~~~}  \\
\end{tabular}

\end{center}
\caption{More qualitative results on AFHQ.}
\label{fig:afhq_more_qualitative_a}
\end{figure*}

\begin{figure*}[t]
\begin{center}

\begin{tabular}{c@{\hskip2pt}c@{\hskip0pt}c@{\hskip0pt}c@{\hskip0pt}c@{\hskip2pt}c}

~~~~ Appearance \begin{turn}{90} ~~~~~~~~~ Pose \end{turn} & \includegraphics[width=0.15\linewidth]{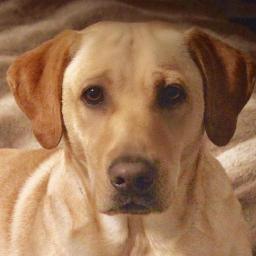} & \includegraphics[width=0.15\linewidth]{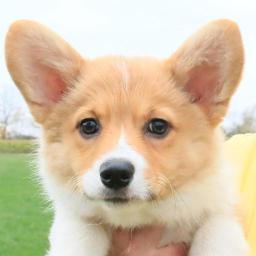} &
\includegraphics[width=0.15\linewidth]{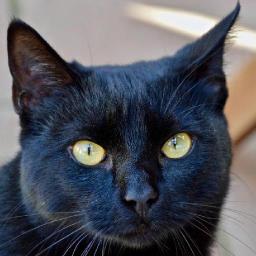} & \includegraphics[width=0.15\linewidth]{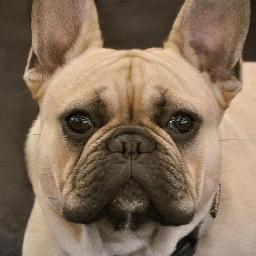} & \\

\includegraphics[width=0.15\linewidth]{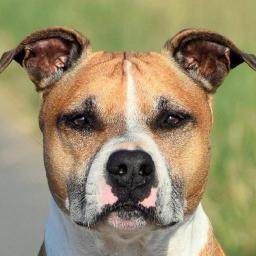} &
\includegraphics[width=0.15\linewidth]{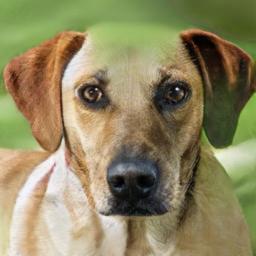} & \includegraphics[width=0.15\linewidth]{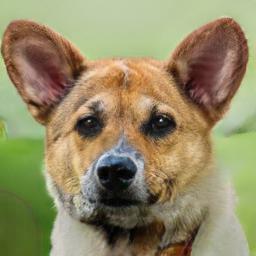} & \includegraphics[width=0.15\linewidth]{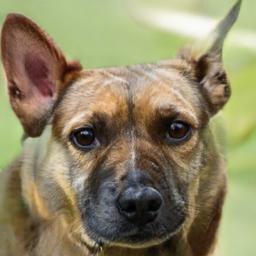} &
\includegraphics[width=0.15\linewidth]{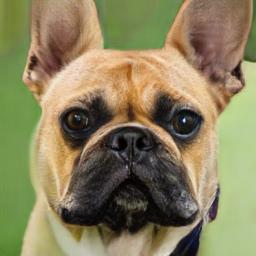} & \rotatebox[origin=r]{270}{FUNIT~~~~~~} \\

& \includegraphics[width=0.15\linewidth]{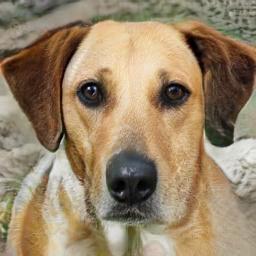} & \includegraphics[width=0.15\linewidth]{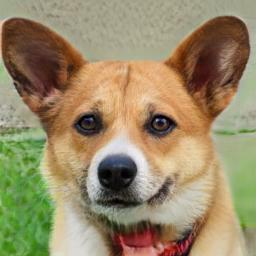} & \includegraphics[width=0.15\linewidth]{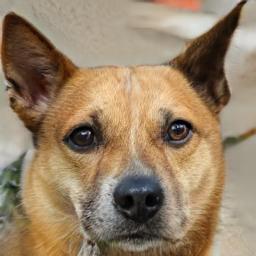} &
\includegraphics[width=0.15\linewidth]{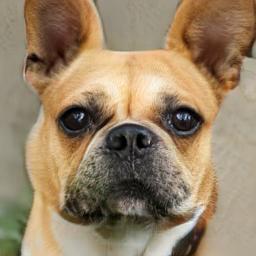} & \rotatebox[origin=r]{270}{StarGAN-v2~~~} \\

& \includegraphics[width=0.15\linewidth]{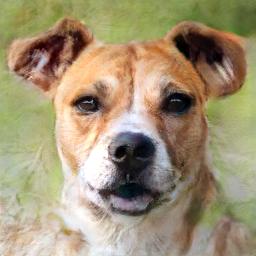} & \includegraphics[width=0.15\linewidth]{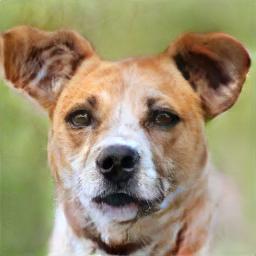} & \includegraphics[width=0.15\linewidth]{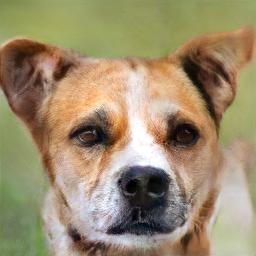} &
\includegraphics[width=0.15\linewidth]{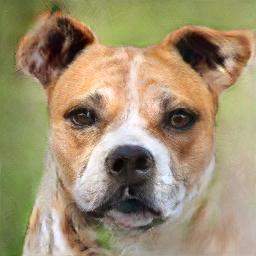} & \rotatebox[origin=r]{270}{\textbf{Ours}~~~~~~~~~}  \\

\includegraphics[width=0.15\linewidth]{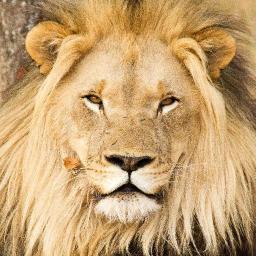} & 
\includegraphics[width=0.15\linewidth]{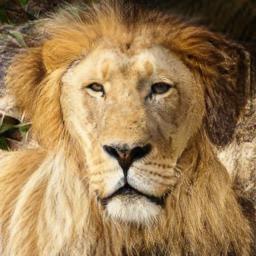} & \includegraphics[width=0.15\linewidth]{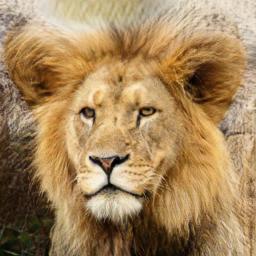} & \includegraphics[width=0.15\linewidth]{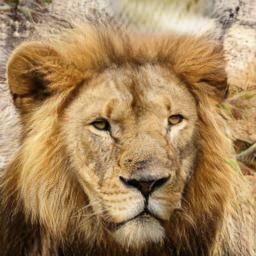} &
\includegraphics[width=0.15\linewidth]{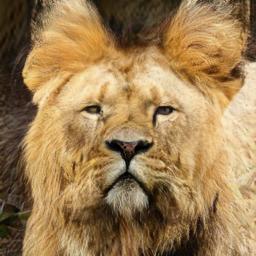} & ~ \rotatebox[origin=r]{270}{FUNIT~~~~~~} \\

& \includegraphics[width=0.15\linewidth]{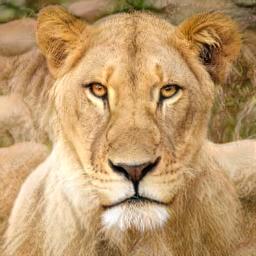} & \includegraphics[width=0.15\linewidth]{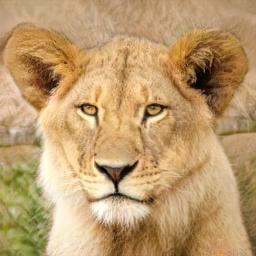} & \includegraphics[width=0.15\linewidth]{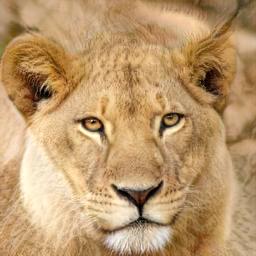} &
\includegraphics[width=0.15\linewidth]{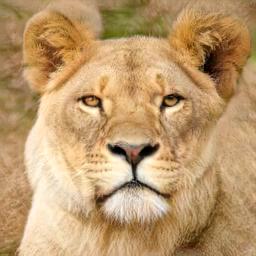} & ~ \rotatebox[origin=r]{270}{StarGAN-v2~~~} \\

& \includegraphics[width=0.15\linewidth]{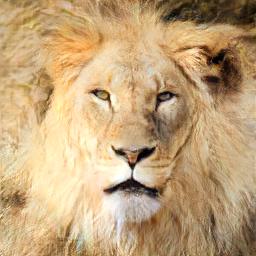} & \includegraphics[width=0.15\linewidth]{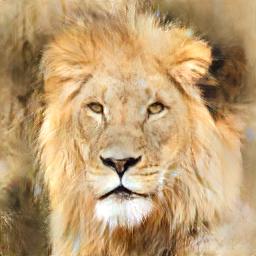} & \includegraphics[width=0.15\linewidth]{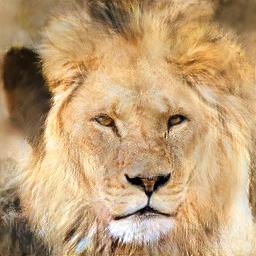} &
\includegraphics[width=0.15\linewidth]{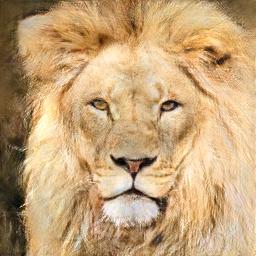} & ~ \rotatebox[origin=r]{270}{\textbf{Ours}~~~~~~~~~}  \\

\end{tabular}

\end{center}
\caption{More qualitative results on AFHQ.}
\label{fig:afhq_more_qualitative_b}
\end{figure*}

\begin{figure*}[t]
\begin{center}
\begin{tabular}{@{\hskip0pt}c@{\hskip2pt}c@{\hskip3pt}c@{\hskip0pt}c@{\hskip0pt}c@{\hskip3pt}c@{\hskip3pt}c@{\hskip0pt}c@{\hskip0pt}c@{\hskip0pt}c@{\hskip0pt}c}

& Input & Fader \cite{lample2017fader} & mGANprior \cite{gu2020mganprior} & \textbf{Ours [uncorr]} & Reference & StarGAN-v2 \cite{choi2019stargan} & \textbf{Ours [corr]} \\
\multirow{4}{*}[-10ex]{\rotatebox[origin=c]{90}{Male-to-Female}} &

\includegraphics[width=0.14\linewidth]{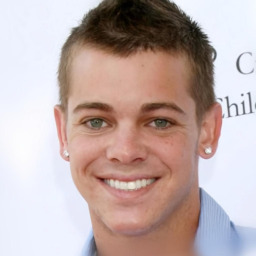} &
\includegraphics[width=0.14\linewidth]{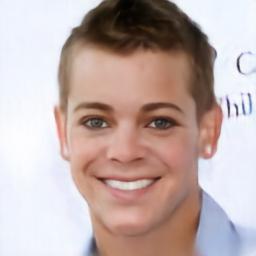} &
\includegraphics[width=0.14\linewidth]{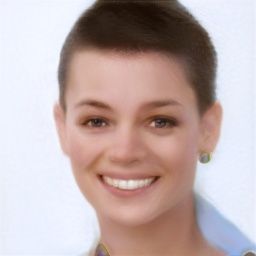} &
\includegraphics[width=0.14\linewidth]{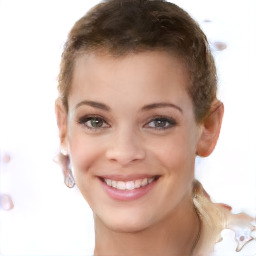} &
\includegraphics[width=0.14\linewidth]{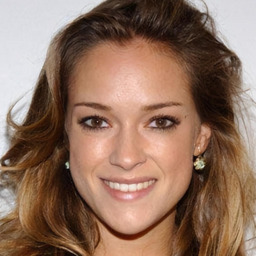} &
\includegraphics[width=0.14\linewidth]{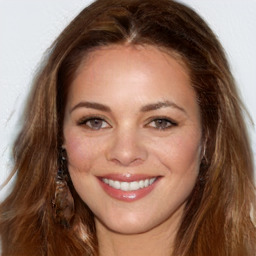} &
\includegraphics[width=0.14\linewidth]{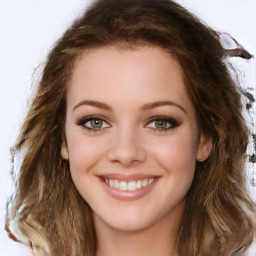} \\

&
\includegraphics[width=0.14\linewidth]{figures/celeba-hq/male-to-female/input/0803.jpg} &
\includegraphics[width=0.14\linewidth]{figures/celeba-hq/male-to-female/fader/004056.jpg} &
\includegraphics[width=0.14\linewidth]{figures/celeba-hq/male-to-female/mganprior/004056.jpg} &
\includegraphics[width=0.14\linewidth]{figures/celeba-hq/male-to-female/ours-uncorr/0803.jpg} &
\includegraphics[width=0.14\linewidth]{figures/celeba-hq/male-to-female/hairstyle/0803_06.jpg} &
\includegraphics[width=0.14\linewidth]{figures/celeba-hq/male-to-female/stargan/0803_06.jpg} &
\includegraphics[width=0.14\linewidth]{figures/celeba-hq/male-to-female/ours-corr/0803_06.jpg} \\

&
\includegraphics[width=0.14\linewidth]{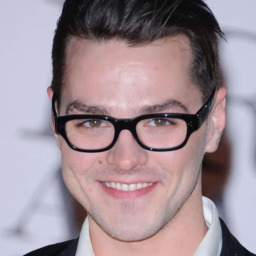} &
\includegraphics[width=0.14\linewidth]{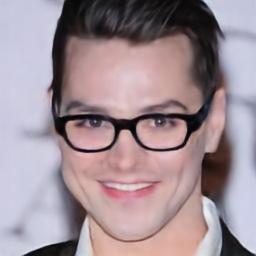} &
\includegraphics[width=0.14\linewidth]{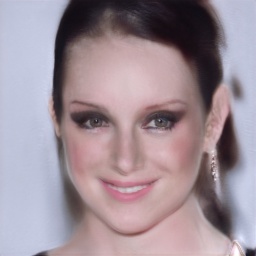} &
\includegraphics[width=0.14\linewidth]{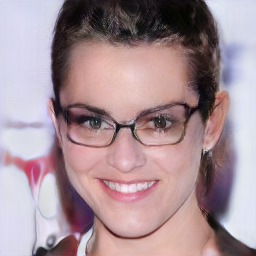} &
\includegraphics[width=0.14\linewidth]{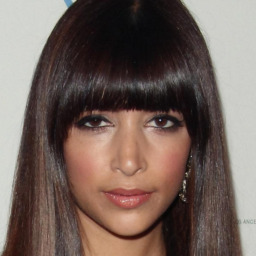} &
\includegraphics[width=0.14\linewidth]{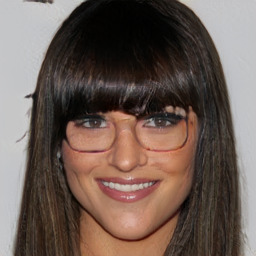} &
\includegraphics[width=0.14\linewidth]{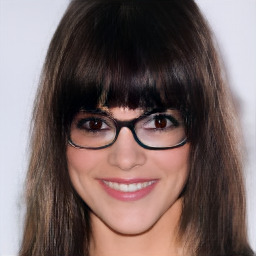} \\

&
\includegraphics[width=0.14\linewidth]{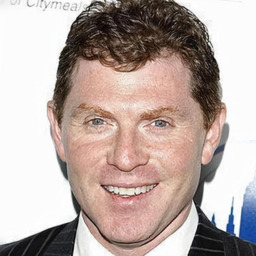} &
\includegraphics[width=0.14\linewidth]{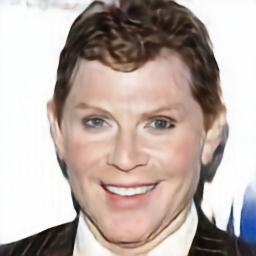} &
\includegraphics[width=0.14\linewidth]{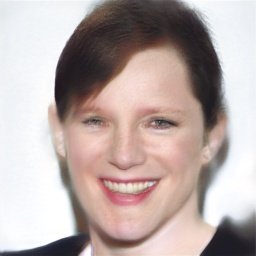} &
\includegraphics[width=0.14\linewidth]{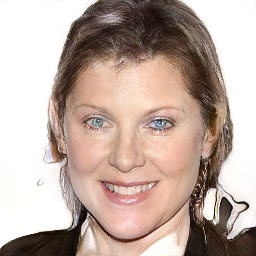} &
\includegraphics[width=0.14\linewidth]{figures/celeba-hq/male-to-female/hairstyle/0149_02.jpg} &
\includegraphics[width=0.14\linewidth]{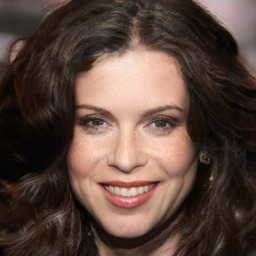} &
\includegraphics[width=0.14\linewidth]{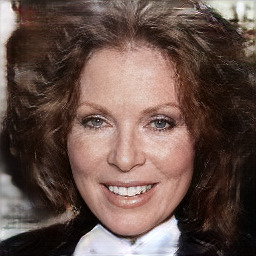} \\

\multirow{4}{*}[-10ex]{\rotatebox[origin=c]{90}{Female-to-Male}} &
\includegraphics[width=0.14\linewidth]{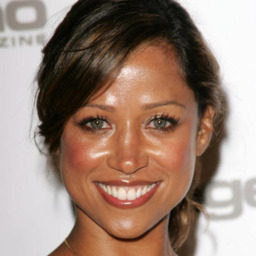} &
\includegraphics[width=0.14\linewidth]{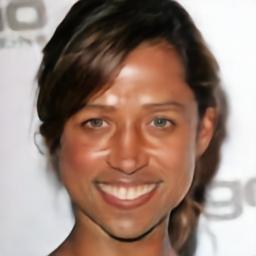} &
\includegraphics[width=0.14\linewidth]{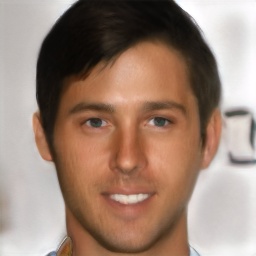} &
\includegraphics[width=0.14\linewidth]{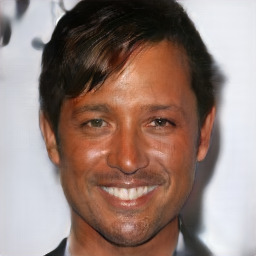} &
\includegraphics[width=0.14\linewidth]{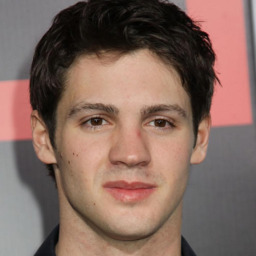} &
\includegraphics[width=0.14\linewidth]{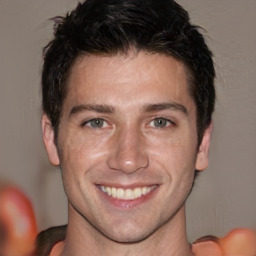} &
\includegraphics[width=0.14\linewidth]{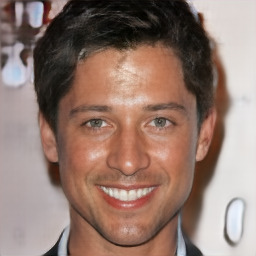} \\

& 
\includegraphics[width=0.14\linewidth]{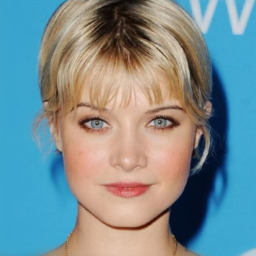} &
\includegraphics[width=0.14\linewidth]{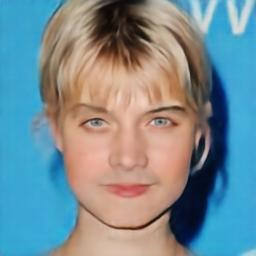} &
\includegraphics[width=0.14\linewidth]{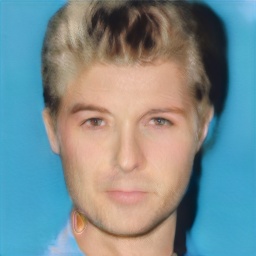} &
\includegraphics[width=0.14\linewidth]{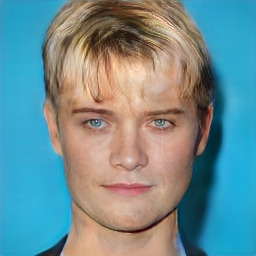} &
\includegraphics[width=0.14\linewidth]{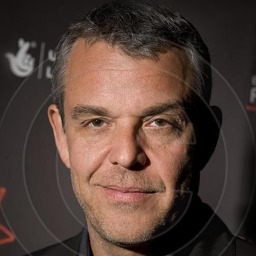} &
\includegraphics[width=0.14\linewidth]{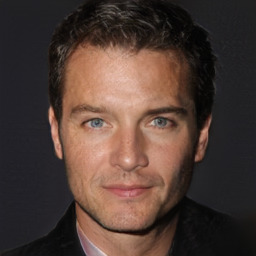} &
\includegraphics[width=0.14\linewidth]{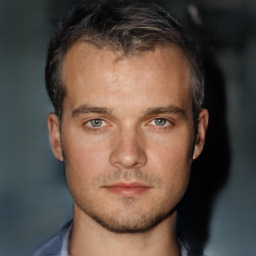} \\

& 
\includegraphics[width=0.14\linewidth]{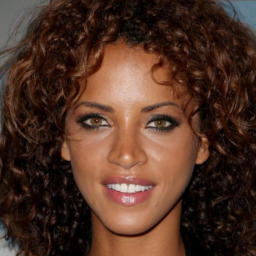} &
\includegraphics[width=0.14\linewidth]{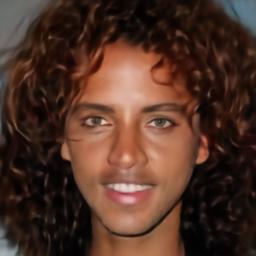} &
\includegraphics[width=0.14\linewidth]{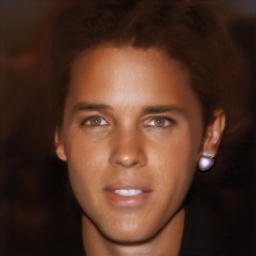} &
\includegraphics[width=0.14\linewidth]{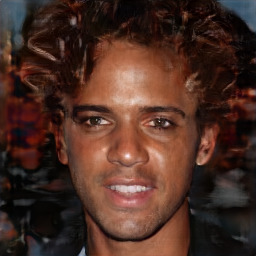} &
\includegraphics[width=0.14\linewidth]{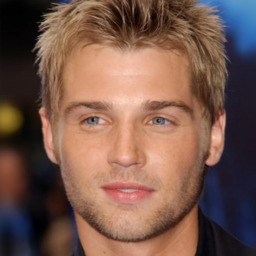} &
\includegraphics[width=0.14\linewidth]{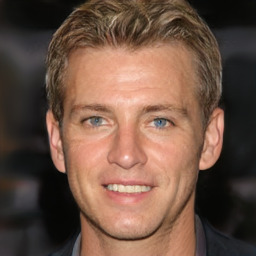} &
\includegraphics[width=0.14\linewidth]{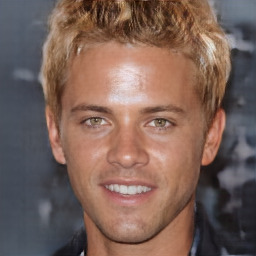} \\

& 
\includegraphics[width=0.14\linewidth]{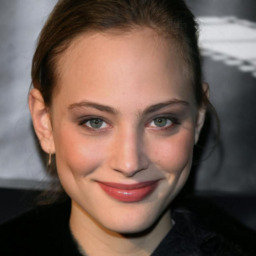} &
\includegraphics[width=0.14\linewidth]{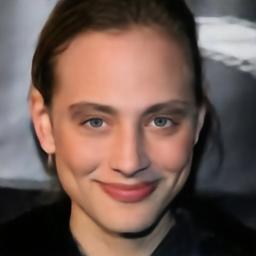} &
\includegraphics[width=0.14\linewidth]{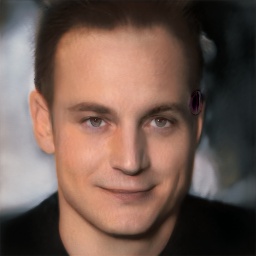} &
\includegraphics[width=0.14\linewidth]{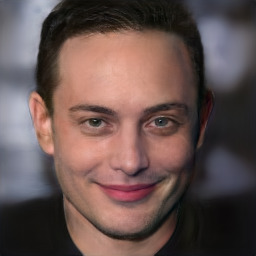} &
\includegraphics[width=0.14\linewidth]{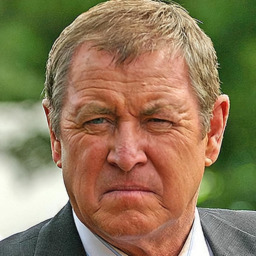} &
\includegraphics[width=0.14\linewidth]{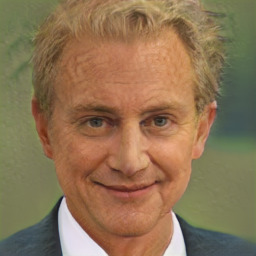} &
\includegraphics[width=0.14\linewidth]{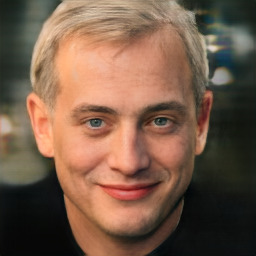} \\

\end{tabular}
\end{center}
\caption{More qualitative results of Male-to-Female translation in two settings: (i) When the attributes are assumed to be \textit{uncorrelated}. (ii) When we model the hair style as the \textit{correlated} attribute and utilize a reference image specifying its target. Our method preserves the uncorrelated attributes including identity, age and illumination better than StarGAN-v2.}
\label{fig:celebahq_gender_more_qualitative_a}
\end{figure*}

\begin{figure*}
\begin{center}

\begin{subfigure}{0.4\textwidth}
\centering
\includegraphics[width=1\textwidth]{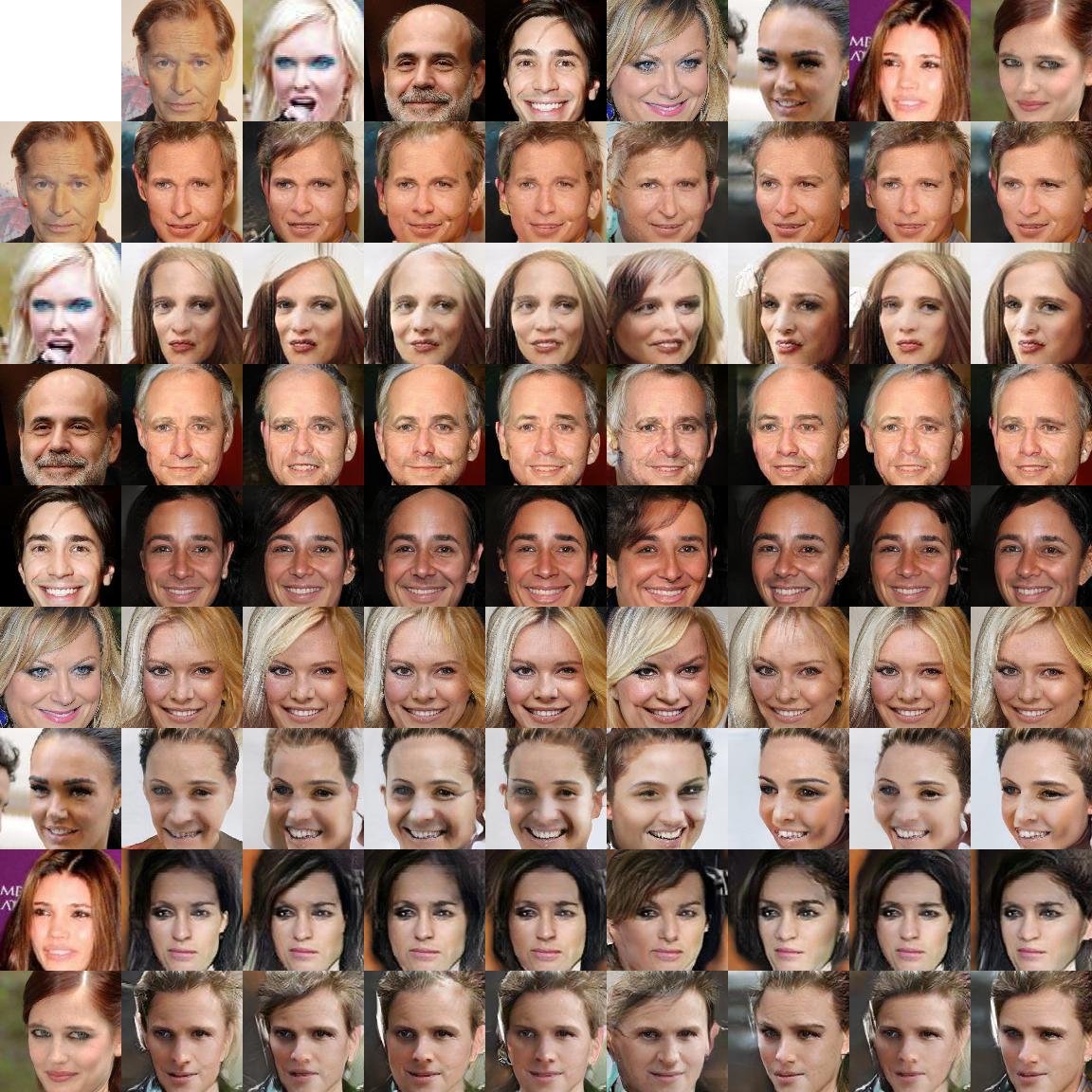} 
\caption{FUNIT}
\vspace{1em}
\end{subfigure}
\quad
\begin{subfigure}{0.4\textwidth}
\centering
\includegraphics[width=1\textwidth]{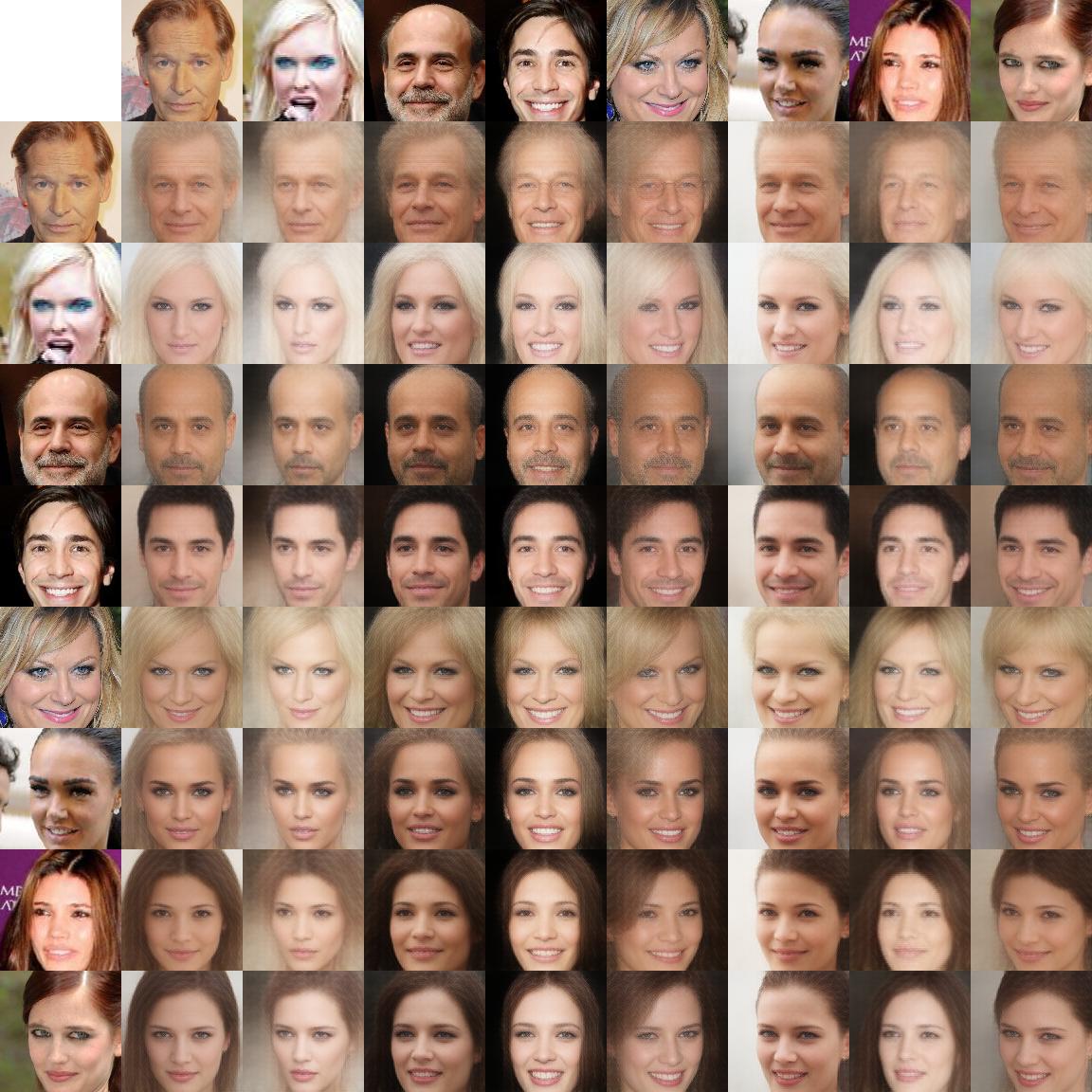}
\caption{LORD}
\vspace{1em}
\end{subfigure}
\begin{subfigure}{0.7\textwidth}
\centering
\includegraphics[width=1\textwidth]{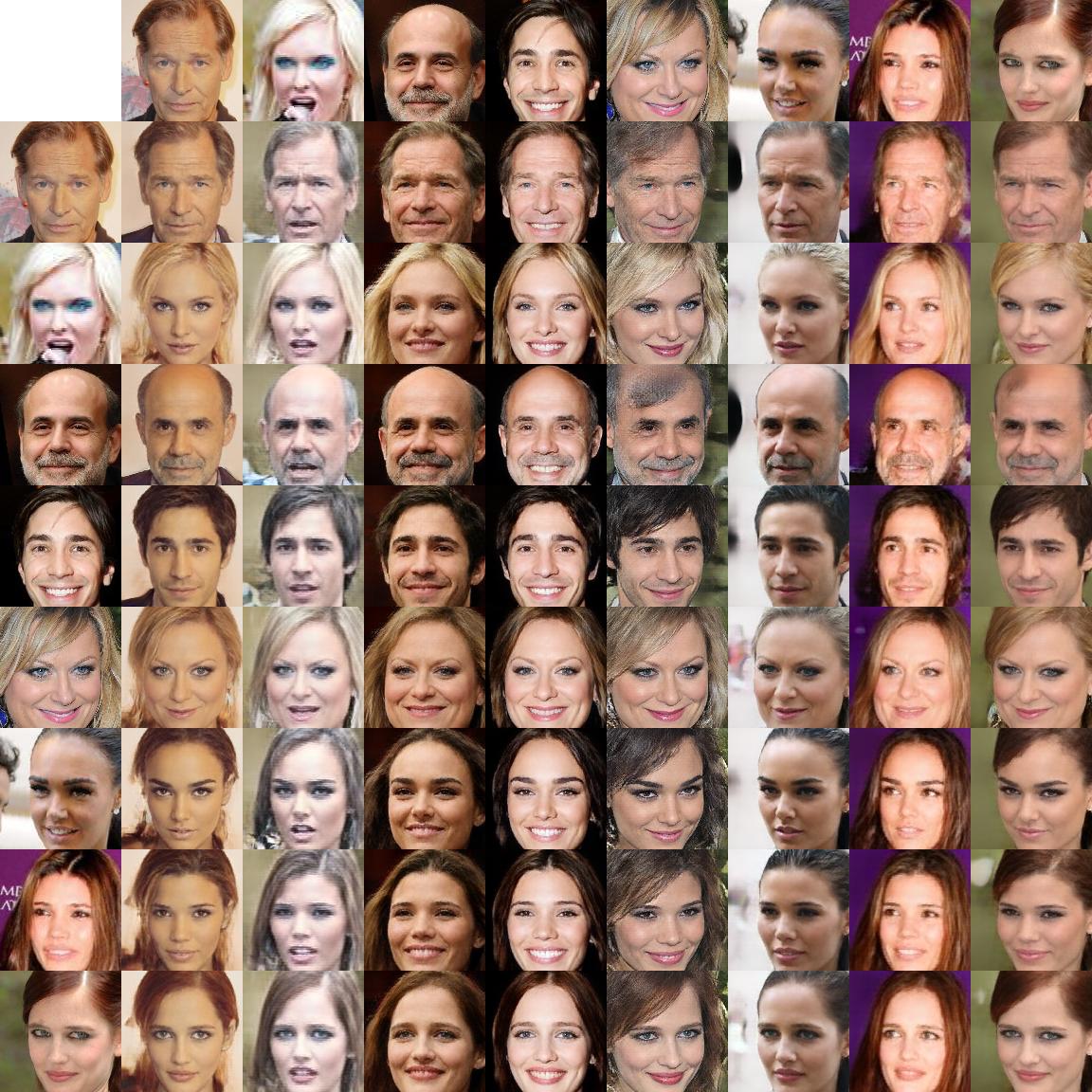} 
\caption{Ours}
\end{subfigure}
\quad

\end{center}
\caption{More qualitative results on CelebA in the task of translating facial identities (left column) across different unlabeled head poses, expressions and illumination conditions (top row).} 
\label{fig:celeba_more_qualitative_a}
\end{figure*}

\begin{figure*}[t]
\begin{center}

\begin{tabular}{c@{\hskip2pt}c@{\hskip0pt}c@{\hskip0pt}c@{\hskip0pt}c}

~~~~~~~~~~~ Texture ~~~~~~~~ \begin{turn}{90} ~~~~~~~~~~~ Shape \end{turn} &
\includegraphics[width=0.14\linewidth]{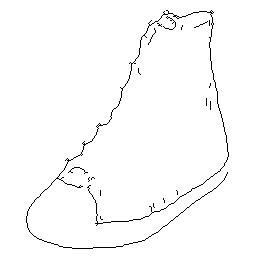} &
\includegraphics[width=0.14\linewidth]{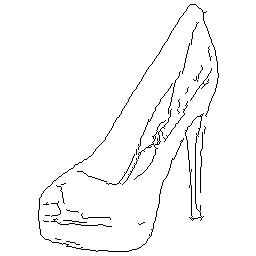} & \includegraphics[width=0.14\linewidth]{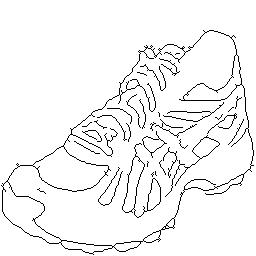} &
\includegraphics[width=0.14\linewidth]{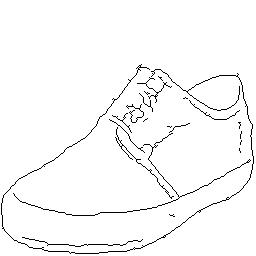} \\

\includegraphics[width=0.14\linewidth]{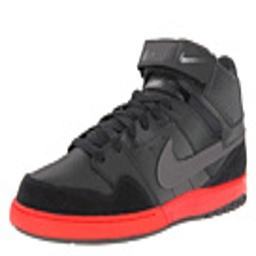} &
\includegraphics[width=0.14\linewidth]{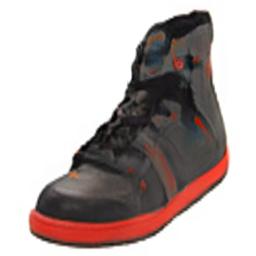} &
\includegraphics[width=0.14\linewidth]{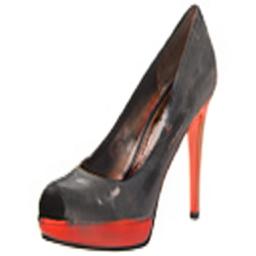} & \includegraphics[width=0.14\linewidth]{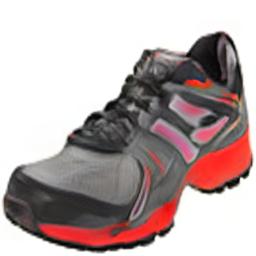} &
\includegraphics[width=0.14\linewidth]{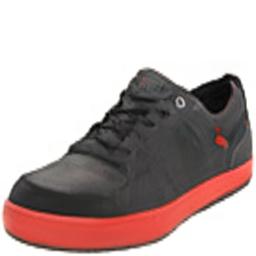} \\

&
\includegraphics[width=0.14\linewidth]{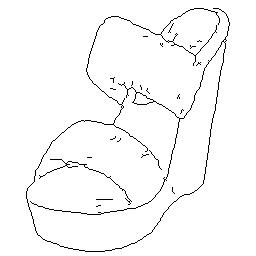} &
\includegraphics[width=0.14\linewidth]{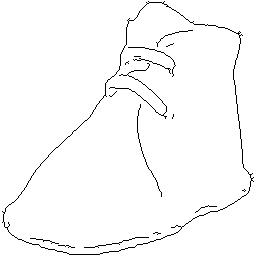} & \includegraphics[width=0.14\linewidth]{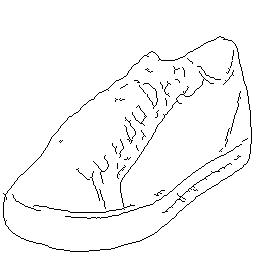} &
\includegraphics[width=0.14\linewidth]{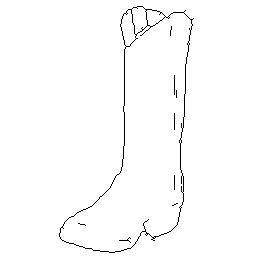} \\

\includegraphics[width=0.14\linewidth]{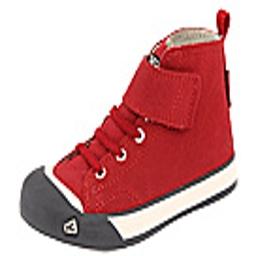} &
\includegraphics[width=0.14\linewidth]{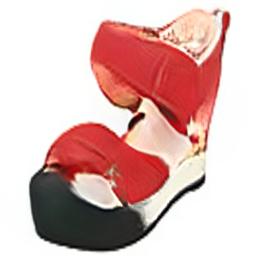} &
\includegraphics[width=0.14\linewidth]{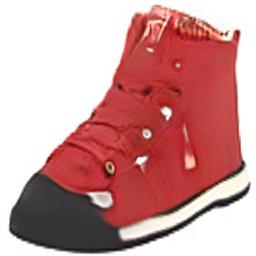} & \includegraphics[width=0.14\linewidth]{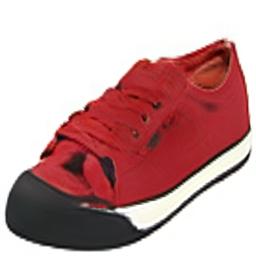} &
\includegraphics[width=0.14\linewidth]{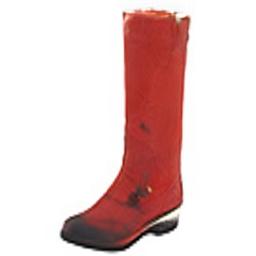} \\

&
\includegraphics[width=0.14\linewidth]{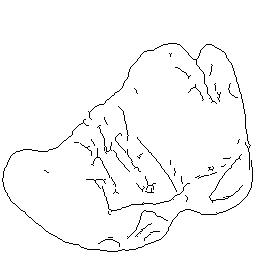} &
\includegraphics[width=0.14\linewidth]{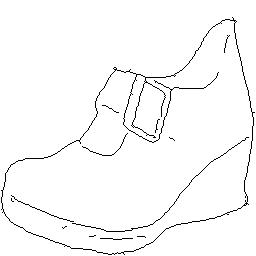} & \includegraphics[width=0.14\linewidth]{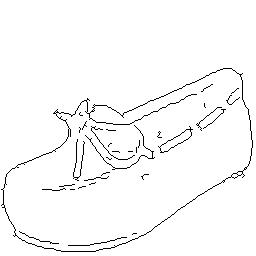} &
\includegraphics[width=0.14\linewidth]{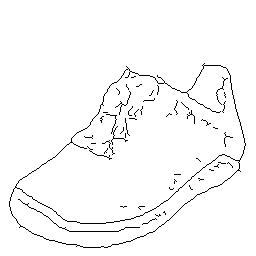} \\

\includegraphics[width=0.14\linewidth]{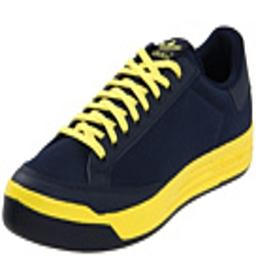} &
\includegraphics[width=0.14\linewidth]{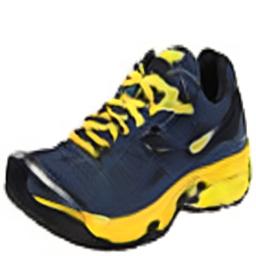} &
\includegraphics[width=0.14\linewidth]{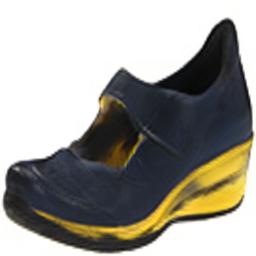} & \includegraphics[width=0.14\linewidth]{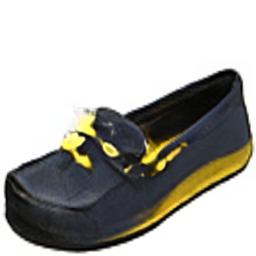} &
\includegraphics[width=0.14\linewidth]{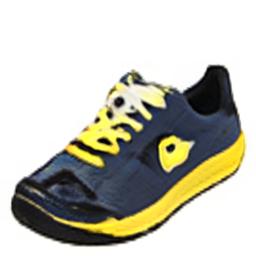} \\

&
\includegraphics[width=0.14\linewidth]{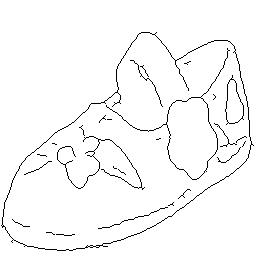} &
\includegraphics[width=0.14\linewidth]{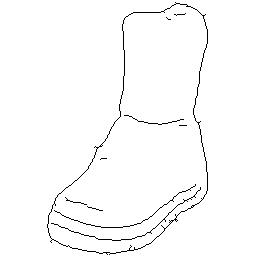} & \includegraphics[width=0.14\linewidth]{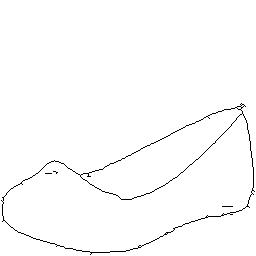} &
\includegraphics[width=0.14\linewidth]{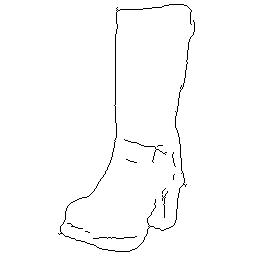} \\

\includegraphics[width=0.14\linewidth]{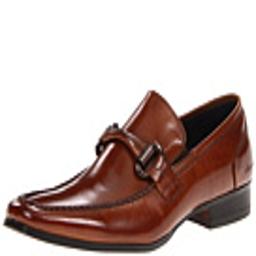} &
\includegraphics[width=0.14\linewidth]{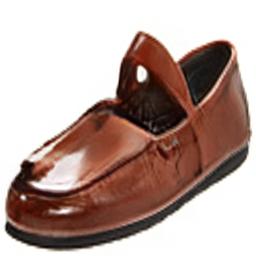} &
\includegraphics[width=0.14\linewidth]{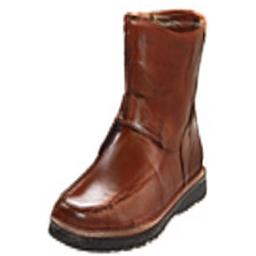} & \includegraphics[width=0.14\linewidth]{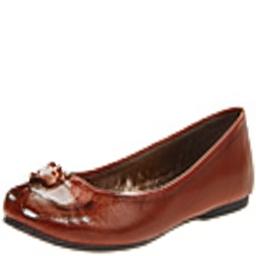} &
\includegraphics[width=0.14\linewidth]{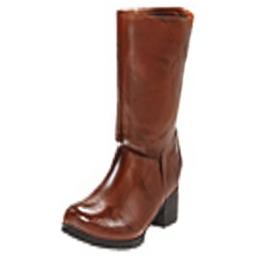} \\

\end{tabular}
\end{center}
\caption{More qualitative results on Edges2Shoes.}
\label{fig:edges2shoes_more_qualitative_a}
\end{figure*}

We provide additional qualitative results on facial age editing (Fig.~\ref{fig:aging_more_qualitative_a}, \ref{fig:aging_more_qualitative_b}, \ref{fig:aging_more_qualitative_c}), identity transfer (Fig.~\ref{fig:celeba_more_qualitative_a}), pose-appearance translation (Fig.~\ref{fig:afhq_more_qualitative_a}, \ref{fig:afhq_more_qualitative_b}), Male-to-Female translation (Fig.~\ref{fig:celebahq_gender_more_qualitative_a}) and Shape-Texture transfer (Fig.~\ref{fig:edges2shoes_more_qualitative_a}).

\subsection{Latent optimization}
\label{app:latent_optimization}
\begin{figure*}
\begin{center}
\includegraphics[width=0.5\linewidth]{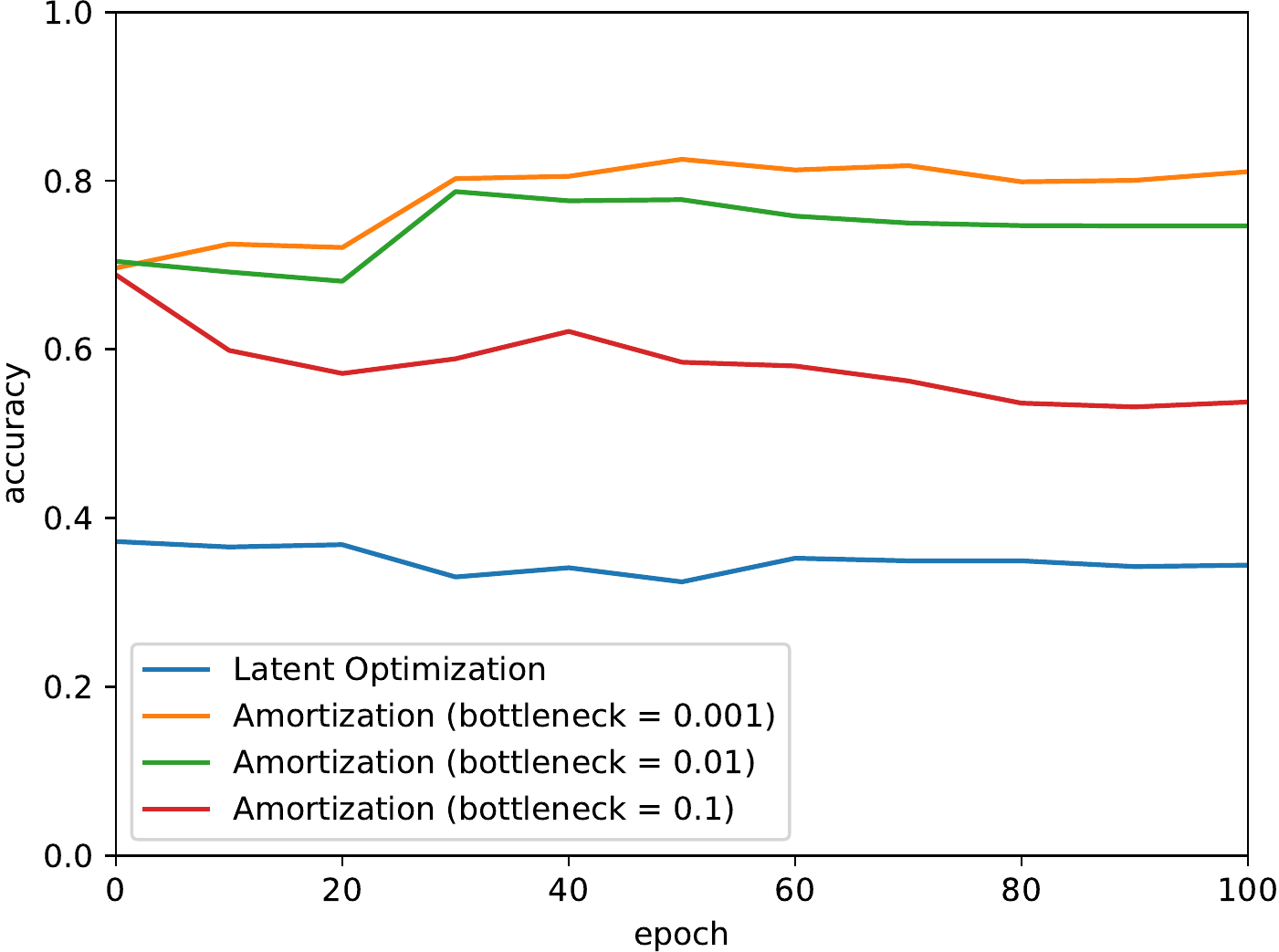} \\
\end{center}
\caption{Evidence for the inductive bias conferred by latent optimization on AFHQ (a validation of the discovery presented in \cite{gabbay2019demystifying}). We plot the accuracy of an auxiliary classifier predicting the labeled attributes from the learned representations of the unlabeled attributes. Latent optimization starts with randomly initialized latent codes and preserves the disentanglement of the labeled and unlabeled representations along the entire training (the accuracy matches a random guess). In contrast, a randomly initialized encoder (amortization) outputs entangled codes. In order to reach disentanglement, the encoder should distillate the information of the labeled attributes during the optimization, which is shown to be unsuccessful in practice.}
\label{fig:inductive_bias}
\end{figure*}

In this work, we opt for learning the representation of the unlabeled uncorrelated attributes ($u^{uncorr}$) using latent optimization, similarly to LORD \cite{gabbay2019demystifying}. Autoencoders assume a parametric model, usually referred to as the encoder, to compute a latent code from an image. On the other hand, we jointly optimize the latent codes and the generator (decoder) parameters. Since the latent codes are learned directly and are unconstrained by a parametric encoder function, our model can recover all the solutions that could be found by an autoencoder, and reach some others.
In order to justify this design choice, we validate the observation presented in \cite{gabbay2019demystifying} stating that latent optimization improves disentanglement and train our disentanglement stage in an amortized fashion using $E_u$. As can be seen in Fig.~\ref{fig:inductive_bias}, amortized training fails to reduce the correlation between the labeled and unlabeled representations. We have experimented with several decay factors ($\lambda_{b}$ in Eq.~\ref{eq:objective_disentanglement}). Although the disentanglement improves as $\lambda_{b}$ increases, the reconstruction gets worse and the model fails to converge with $\lambda_{b} > 0.1$.

\subsection{Visualization of ablation analysis}
\label{app:ablation}
\begin{figure*}[t]
\begin{center}

\begin{tabular}{c@{\hskip2pt}c@{\hskip0pt}c@{\hskip0pt}c@{\hskip0pt}c@{\hskip0pt}c@{\hskip0pt}c}

~~~Appearance \begin{turn}{90} ~~~~~~~Pose \end{turn} & \includegraphics[width=0.11\linewidth]{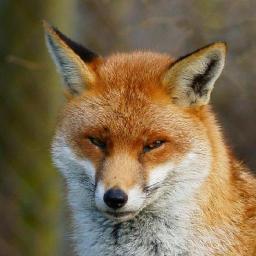} & \includegraphics[width=0.11\linewidth]{figures/ablation/content/1.jpg} &
\includegraphics[width=0.11\linewidth]{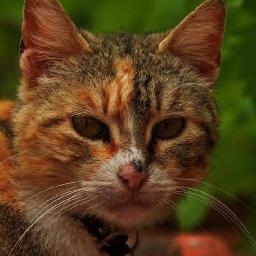} & \includegraphics[width=0.11\linewidth]{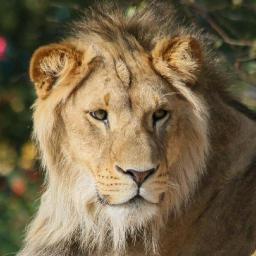} & \\

\includegraphics[width=0.11\linewidth]{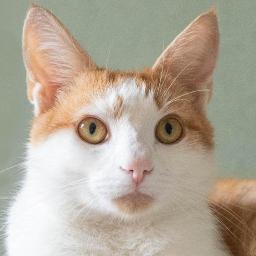} & \includegraphics[width=0.11\linewidth]{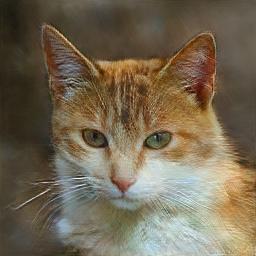} & \includegraphics[width=0.11\linewidth]{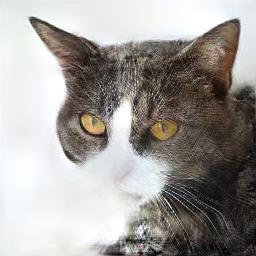} & \includegraphics[width=0.11\linewidth]{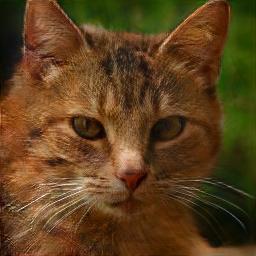} &
\includegraphics[width=0.11\linewidth]{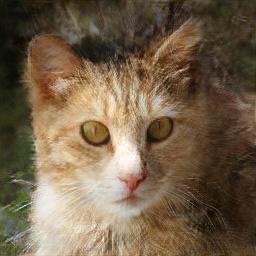} & ~ \rotatebox[origin=r]{270}{w/o $x^{corr}$ ~} \\

& \includegraphics[width=0.11\linewidth]{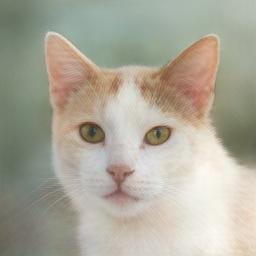} & \includegraphics[width=0.11\linewidth]{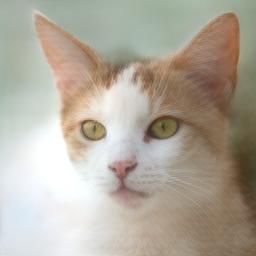} & \includegraphics[width=0.11\linewidth]{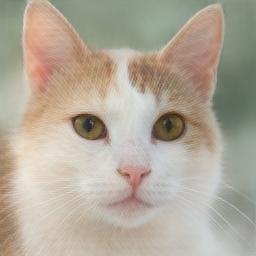} &
\includegraphics[width=0.11\linewidth]{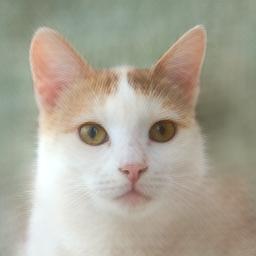} & ~ \rotatebox[origin=r]{270}{w/o adv.~~~} \\

& \includegraphics[width=0.11\linewidth]{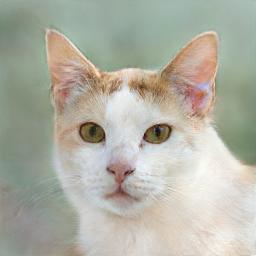} & \includegraphics[width=0.11\linewidth]{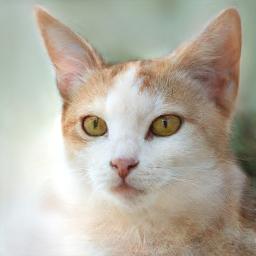} & \includegraphics[width=0.11\linewidth]{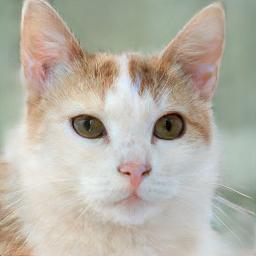} &
\includegraphics[width=0.11\linewidth]{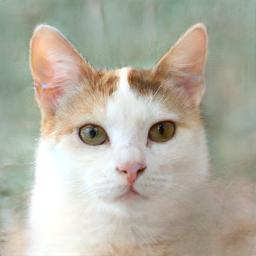} & ~ \rotatebox[origin=r]{270}{\textbf{Ours} ~~~~} \\

\includegraphics[width=0.11\linewidth]{figures/ablation/style/1.jpg} & \includegraphics[width=0.11\linewidth]{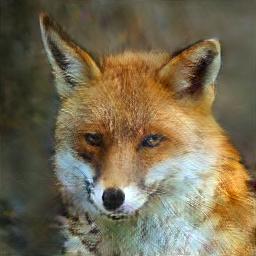} & \includegraphics[width=0.11\linewidth]{figures/ablation/without-style/1-1.jpg} & \includegraphics[width=0.11\linewidth]{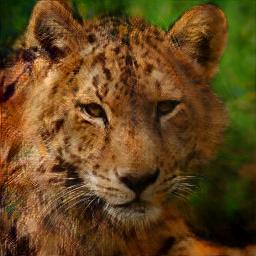} &
\includegraphics[width=0.11\linewidth]{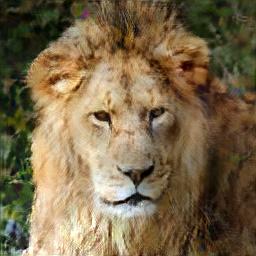} & ~ \rotatebox[origin=r]{270}{w/o $x^{corr}$ ~} \\

& \includegraphics[width=0.11\linewidth]{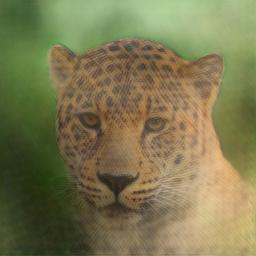} & \includegraphics[width=0.11\linewidth]{figures/ablation/without-gan/1-1.jpg} & \includegraphics[width=0.11\linewidth]{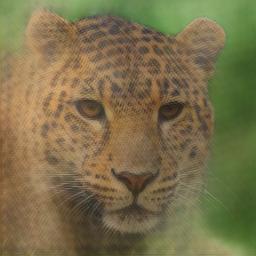} &
\includegraphics[width=0.11\linewidth]{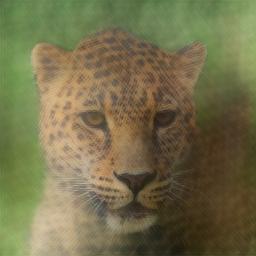} & ~ \rotatebox[origin=r]{270}{w/o adv.~~~} \\

& \includegraphics[width=0.11\linewidth]{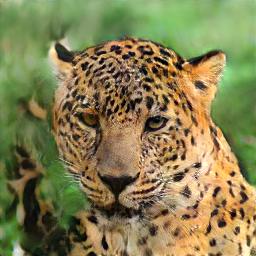} & \includegraphics[width=0.11\linewidth]{figures/ablation/ours/1-1.jpg} & \includegraphics[width=0.11\linewidth]{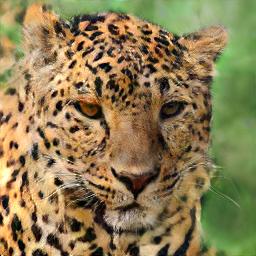} &
\includegraphics[width=0.11\linewidth]{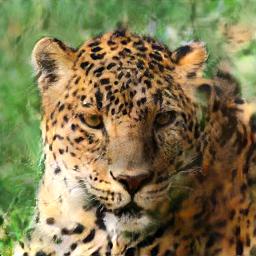} & ~ \rotatebox[origin=r]{270}{\textbf{Ours} ~~~~} \\

\includegraphics[width=0.11\linewidth]{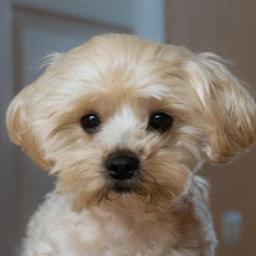} & \includegraphics[width=0.11\linewidth]{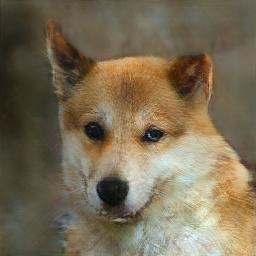} & \includegraphics[width=0.11\linewidth]{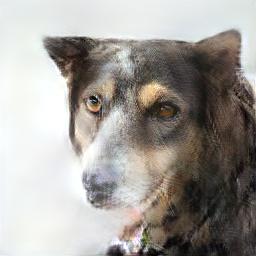} & \includegraphics[width=0.11\linewidth]{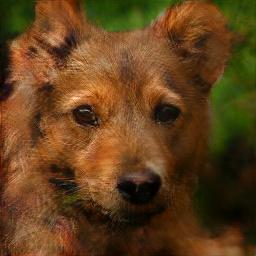} &
\includegraphics[width=0.11\linewidth]{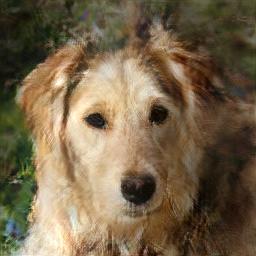} & ~ \rotatebox[origin=r]{270}{w/o $x^{corr}$ ~} \\

& \includegraphics[width=0.11\linewidth]{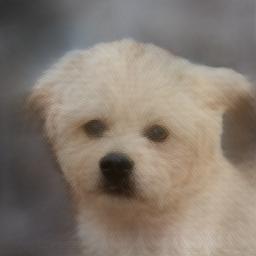} & \includegraphics[width=0.11\linewidth]{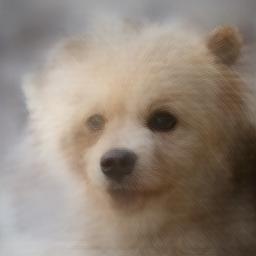} & \includegraphics[width=0.11\linewidth]{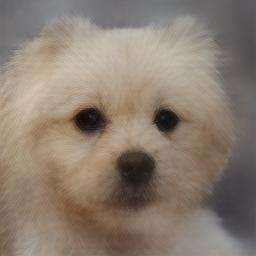} &
\includegraphics[width=0.11\linewidth]{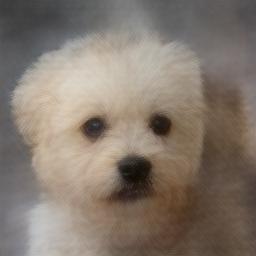} & ~ \rotatebox[origin=r]{270}{w/o adv.~~~} \\

& \includegraphics[width=0.11\linewidth]{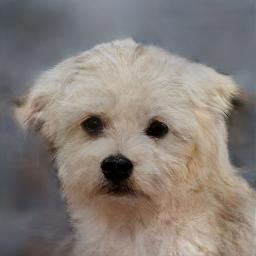} & \includegraphics[width=0.11\linewidth]{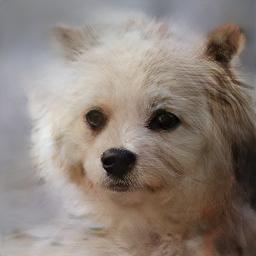} & \includegraphics[width=0.11\linewidth]{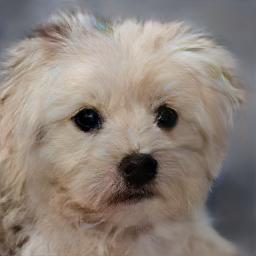} &
\includegraphics[width=0.11\linewidth]{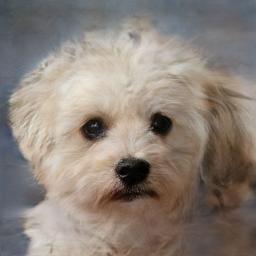} & ~ \rotatebox[origin=r]{270}{\textbf{Ours} ~~~~} \\

\end{tabular}

\end{center}
\caption{Qualitative examples from the ablation analysis; (i) w/o $x^{corr}$: Leaving the correlated attributes intact leads to unreliable and entangled translations. (ii) w/o adv.: Disentanglement is achieved without the adversarial loss, which mostly contributes to the visual fidelity.}
\label{fig:ablation_afhq}
\end{figure*}

\begin{figure*}[t]
\begin{center}

\begin{tabular}{c@{\hskip3pt}c@{\hskip0pt}c@{\hskip0pt}c@{\hskip0pt}c@{\hskip0pt}c}

~~~~~~~~~~~~~~~~~~~~~~~~~~ \begin{turn}{90} ~~~~~~~~~ Pose \end{turn} &
\includegraphics[width=0.15\linewidth]{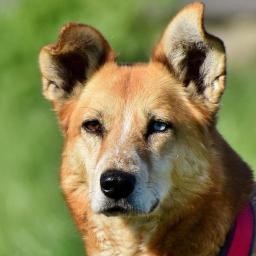} & \includegraphics[width=0.15\linewidth]{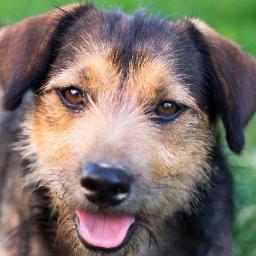} &
\includegraphics[width=0.15\linewidth]{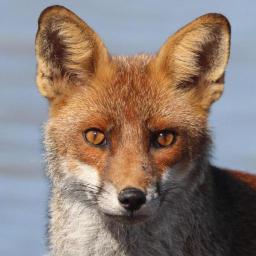} & \includegraphics[width=0.15\linewidth]{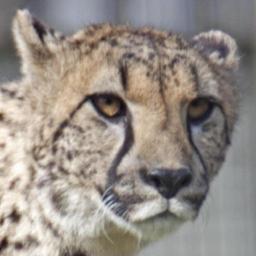} \\

~~~~~ Appearance ~ \begin{turn}{90} ~~~~~~~~ To Cat \end{turn} & \includegraphics[width=0.15\linewidth]{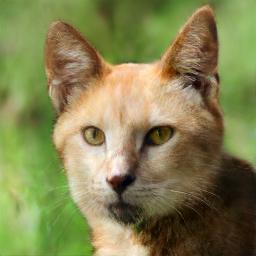} & \includegraphics[width=0.15\linewidth]{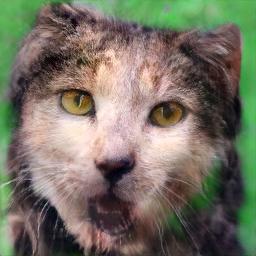} & \includegraphics[width=0.15\linewidth]{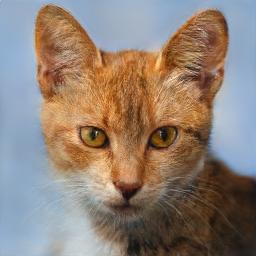} &
\includegraphics[width=0.15\linewidth]{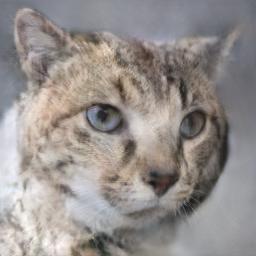} \\

\includegraphics[width=0.15\linewidth]{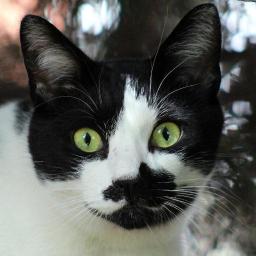} & \includegraphics[width=0.15\linewidth]{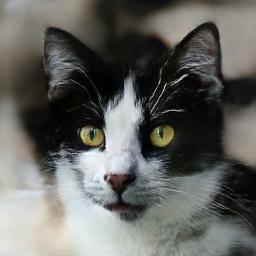} & \includegraphics[width=0.15\linewidth]{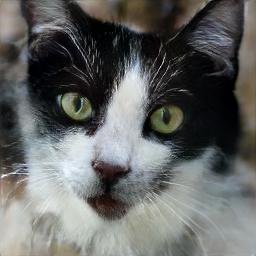} & \includegraphics[width=0.15\linewidth]{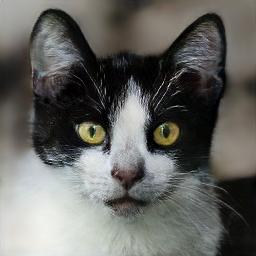} &
\includegraphics[width=0.15\linewidth]{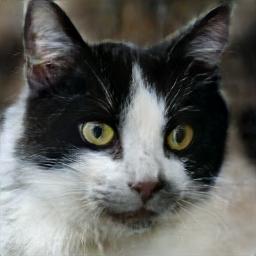} \\

~~~~~ Appearance ~ \begin{turn}{90} ~~~~~~~~ To Dog \end{turn} & \includegraphics[width=0.15\linewidth]{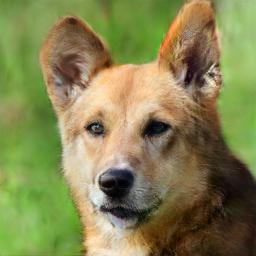} & \includegraphics[width=0.15\linewidth]{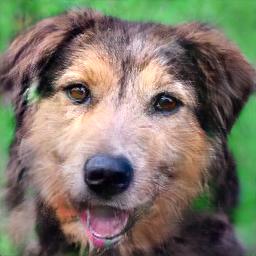} & \includegraphics[width=0.15\linewidth]{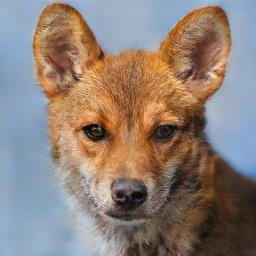} &
\includegraphics[width=0.15\linewidth]{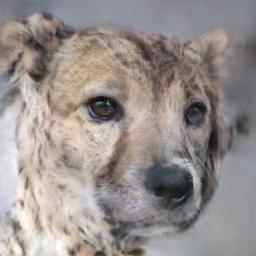} \\

\includegraphics[width=0.15\linewidth]{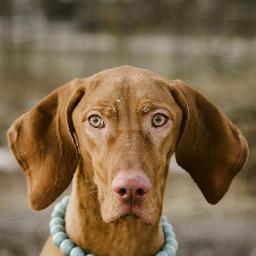} & \includegraphics[width=0.15\linewidth]{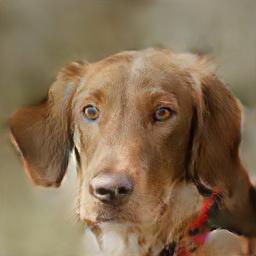} & \includegraphics[width=0.15\linewidth]{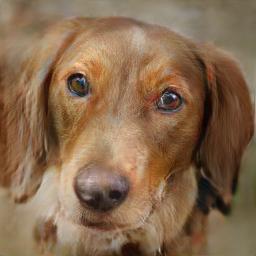} & \includegraphics[width=0.15\linewidth]{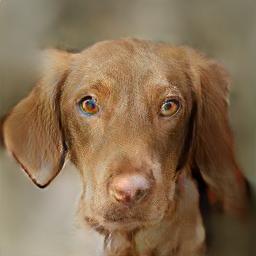} &
\includegraphics[width=0.15\linewidth]{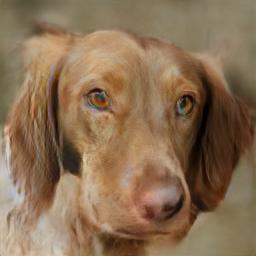} \\

~~~~~ Appearance ~ \begin{turn}{90} ~~~~~~~ To Wild \end{turn} & \includegraphics[width=0.15\linewidth]{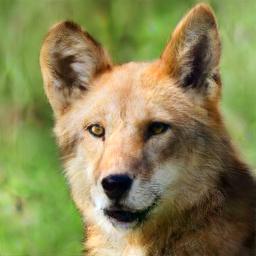} & \includegraphics[width=0.15\linewidth]{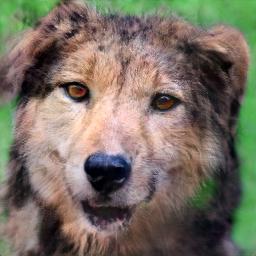} & \includegraphics[width=0.15\linewidth]{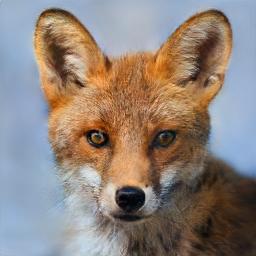} &
\includegraphics[width=0.15\linewidth]{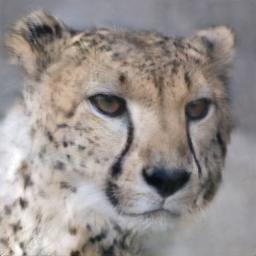} \\

\includegraphics[width=0.15\linewidth]{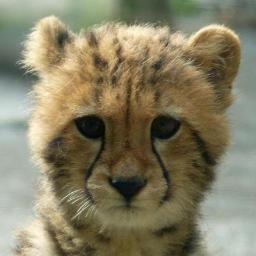} & \includegraphics[width=0.15\linewidth]{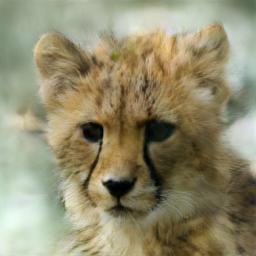} & \includegraphics[width=0.15\linewidth]{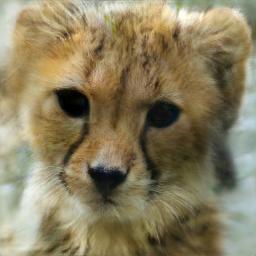} & \includegraphics[width=0.15\linewidth]{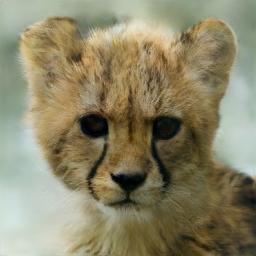} &
\includegraphics[width=0.15\linewidth]{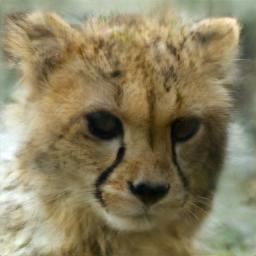} \\

\end{tabular}

\end{center}
\caption{Visualization of the three sets of attributes modeled by our framework. Changing the labeled attribute (e.g. To cat, dog, wild) while leaving the unlabeled correlated attributes intact affects high level semantics of the presented animal, although can generate unreliable translations (e.g. rightmost image in "To Dog"). Guiding the correlated attributes by a reference image allows for specification of the exact target appearance (e.g. breed). The remaining unlabeled and uncorrelated attributes mainly encode the pose of the animal.}
\label{fig:three_axes_afhq}
\end{figure*}

Examples from the ablation analysis are provided in Fig.~\ref{fig:ablation_afhq}.
Visualization of the three sets of attributes modeled by our method is provided in Fig.~\ref{fig:three_axes_afhq}.

\begin{figure*}[t]
\begin{center}

\begin{tabular}{c@{\hskip2pt}c@{\hskip0pt}c@{\hskip0pt}c@{\hskip0pt}c@{\hskip3pt}c}

~~~~ Appearance \begin{turn}{90} ~~~~~~~~~ Pose \end{turn} & \includegraphics[width=0.15\linewidth]{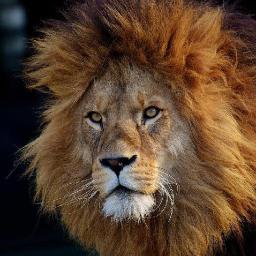} & \includegraphics[width=0.15\linewidth]{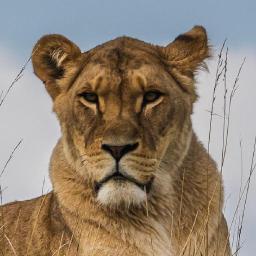} &
\includegraphics[width=0.15\linewidth]{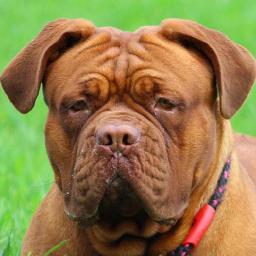} & \includegraphics[width=0.15\linewidth]{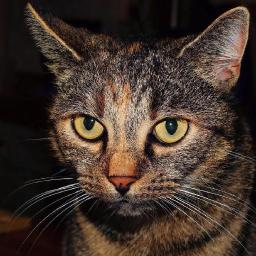} & \\

\includegraphics[width=0.15\linewidth]{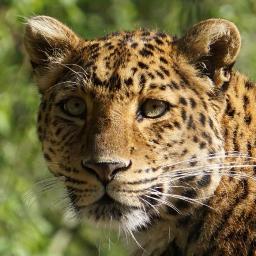} &
\includegraphics[width=0.15\linewidth]{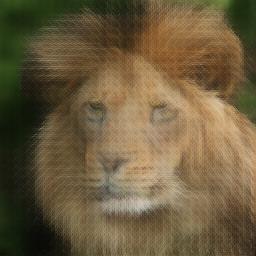} & \includegraphics[width=0.15\linewidth]{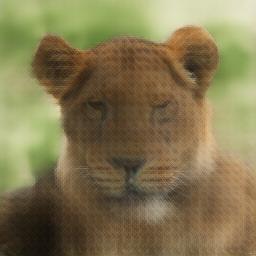} & \includegraphics[width=0.15\linewidth]{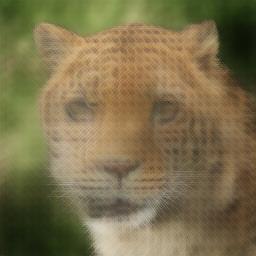} &
\includegraphics[width=0.15\linewidth]{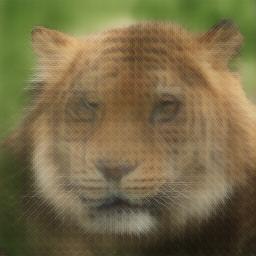} & \rotatebox[origin=r]{270}{LORD~~~~~~~}  \\

& \includegraphics[width=0.15\linewidth]{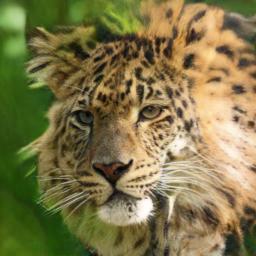} & \includegraphics[width=0.15\linewidth]{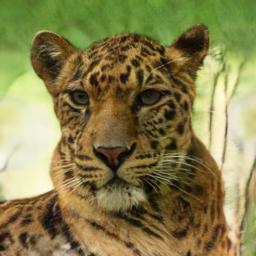} & \includegraphics[width=0.15\linewidth]{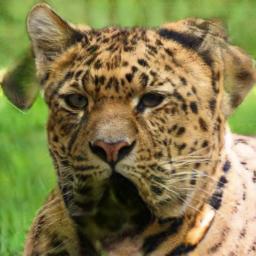} &
\includegraphics[width=0.15\linewidth]{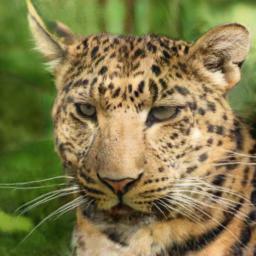} & \rotatebox[origin=r]{270}{FUNIT~~~~~~} \\

& \includegraphics[width=0.15\linewidth]{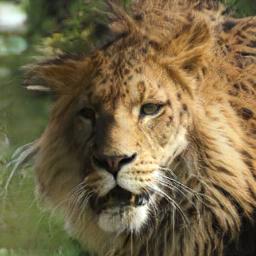} & \includegraphics[width=0.15\linewidth]{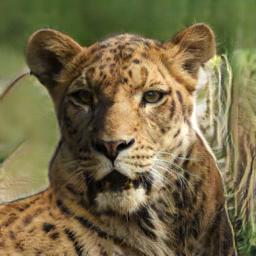} & \includegraphics[width=0.15\linewidth]{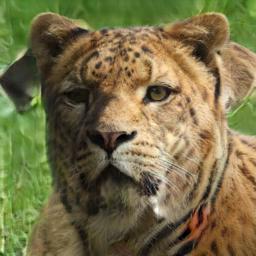} &
\includegraphics[width=0.15\linewidth]{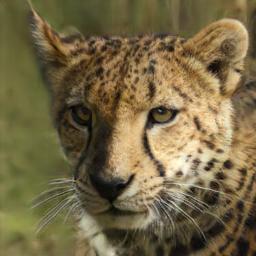} & \rotatebox[origin=r]{270}{StarGAN-v2~~~} \\

& \includegraphics[width=0.15\linewidth]{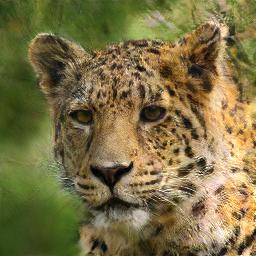} & \includegraphics[width=0.15\linewidth]{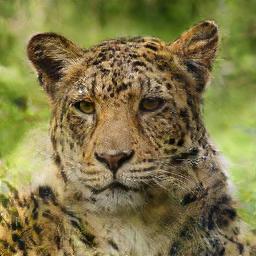} & \includegraphics[width=0.15\linewidth]{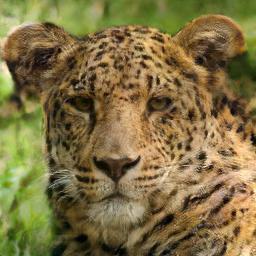} &
\includegraphics[width=0.15\linewidth]{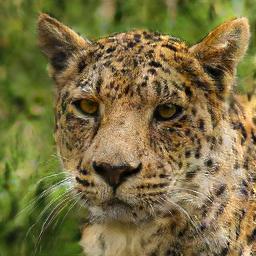} & \rotatebox[origin=r]{270}{\textbf{Ours}~~~~~~~~~}  \\

\end{tabular}

\end{center}
\caption{Comparison against LORD in the presence of correlated attributes (AFHQ). LORD does not distinct between correlated and uncorrelated attributes and can not utilize a reference image e.g. translating the cat into a wild animal is poorly specified and results in a tiger instead of a leopard. Moreover, the generated images exhibit low visual quality which is further improved by our method. FUNIT and StarGAN-v2 leak some of the correlated attributes such as the lion's mane and the dog's facial shape, leading to unreliable translation between species.}
\label{fig:afhq_vs_lord}
\end{figure*}

% \begin{figure*}
% \begin{center}

% \begin{subfigure}{0.4\textwidth}
% \centering
% \includegraphics[width=1\textwidth]{figures/celeba/d/funit.jpg} 
% \caption{FUNIT}
% \vspace{1em}
% \end{subfigure}
% \quad
% \begin{subfigure}{0.4\textwidth}
% \centering
% \includegraphics[width=1\textwidth]{figures/celeba/d/lord.jpg}
% \caption{LORD}
% \vspace{1em}
% \end{subfigure}
% \begin{subfigure}{0.7\textwidth}
% \centering
% \includegraphics[width=1\textwidth]{figures/celeba/d/ours.jpg} 
% \caption{Ours}
% \end{subfigure}
% \quad

% \end{center}
% \caption{More qualitative results on CelebA in the task of translating facial identities (left column) across different unlabeled head poses, expressions and illumination conditions (top row).} 
% \label{fig:celeba_more_qualitative_b}
% \end{figure*}

\begin{table*}
\centering
\caption{Generator architecture based on StyleGAN2. StyleConv and ModulatedConv use the injected latent code composed of $y$, $u^{corr}$, $u^{uncorr}$.}
\label{tab:generator}
\begin{tabular}{ccccc}
    \toprule
    Layer & Kernel Size & Activation & Resample & Output Shape \\
    \midrule
    Constant Input & - & - & - & $4 \times 4 \times 512$ \\
    StyledConv & $3 \times 3$ & FusedLeakyReLU & - & $4 \times 4 \times 512$ \\
    StyledConv & $3 \times 3$ & FusedLeakyReLU & UpFirDn2d & $8 \times 8 \times 512$ \\
    StyledConv & $3 \times 3$ & FusedLeakyReLU & - & $8 \times 8 \times 512$ \\
    StyledConv & $3 \times 3$ & FusedLeakyReLU & UpFirDn2d & $16 \times 16 \times 512$ \\
    StyledConv & $3 \times 3$ & FusedLeakyReLU & - & $16 \times 16 \times 512$ \\
    StyledConv & $3 \times 3$ & FusedLeakyReLU & UpFirDn2d & $32 \times 32 \times 512$ \\
    StyledConv & $3 \times 3$ & FusedLeakyReLU & - & $32 \times 32 \times 512$ \\
    StyledConv & $3 \times 3$ & FusedLeakyReLU & UpFirDn2d & $64 \times 64 \times 512$ \\
    StyledConv & $3 \times 3$ & FusedLeakyReLU & - & $64 \times 64 \times 512$ \\
    StyledConv & $3 \times 3$ & FusedLeakyReLU & UpFirDn2d & $128 \times 128 \times 256$ \\
    StyledConv & $3 \times 3$ & FusedLeakyReLU & - & $128 \times 128 \times 256$ \\
    StyledConv & $3 \times 3$ & FusedLeakyReLU & UpFirDn2d & $256 \times 256 \times 128$ \\
    StyledConv & $3 \times 3$ & FusedLeakyReLU & - & $256 \times 256 \times 128$ \\
    ModulatedConv & $1 \times 1$ & - & - & $256 \times 256 \times 3$ \\
	\bottomrule
\end{tabular}
\end{table*}

\begin{table*}
\centering
\caption{Discriminator architecture based on StyleGAN2.}
\label{tab:discriminator}
\begin{tabular}{ccccc}
    \toprule
    Layer & Kernel Size & Activation & Resample & Output Shape \\
    \midrule
    Input & - & - & - & $256 \times 256 \times 3$ \\
    Conv & $3 \times 3$ & FusedLeakyReLU & - & $256 \times 256 \times 128$ \\
    ResBlock & $3 \times 3$ & FusedLeakyReLU & UpFirDn2d & $128 \times 128 \times 256$ \\
    ResBlock & $3 \times 3$ & FusedLeakyReLU & UpFirDn2d & $64 \times 64 \times 512$ \\
    ResBlock & $3 \times 3$ & FusedLeakyReLU & UpFirDn2d & $32 \times 32 \times 512$ \\
    ResBlock & $3 \times 3$ & FusedLeakyReLU & UpFirDn2d & $16 \times 16 \times 512$ \\
    ResBlock & $3 \times 3$ & FusedLeakyReLU & UpFirDn2d & $8 \times 8 \times 512$ \\
    ResBlock & $3 \times 3$ & FusedLeakyReLU & UpFirDn2d & $4 \times 4 \times 512$ \\
    Concat stddev & $3 \times 3$ & FusedLeakyReLU & UpFirDn2d & $4 \times 4 \times 513$ \\
    Conv & $3 \times 3$ & FusedLeakyReLU & - & $4 \times 4 \times 512$ \\
    Reshape & - & - & - & 8192 \\
    FC & - & FusedLeakyReLU & - & 512 \\
    FC & - & - & - & 1 \\
    
	\bottomrule
\end{tabular}
\end{table*}

\begin{table*}
\centering
\caption{Encoder architecture based on StarGAN-v2. Note that we do not use any domain-specific layers. $D$ is the dimension of $y$, $u^{corr}$, $u^{uncorr}$ respectively.}
\label{tab:encoder}
\begin{tabular}{ccccc}
    \toprule
    Layer & Kernel Size & Activation & Resample & Output Shape \\
    \midrule
    Input & - & - & - & $256 \times 256 \times 3$ \\
    Conv & $3 \times 3$ & - & - & $256 \times 256 \times 64$ \\
    ResBlock & $3 \times 3$ & LeakyReLU ($\alpha = 0.2$) & Avg Pool & $128 \times 128 \times 128$ \\
    ResBlock & $3 \times 3$ & LeakyReLU ($\alpha = 0.2$) & Avg Pool & $64 \times 64 \times 256$ \\
    ResBlock & $3 \times 3$ & LeakyReLU ($\alpha = 0.2$) & Avg Pool & $32 \times 32 \times 256$ \\
    ResBlock & $3 \times 3$ & LeakyReLU ($\alpha = 0.2$) & Avg Pool & $16 \times 16 \times 256$ \\
    ResBlock & $3 \times 3$ & LeakyReLU ($\alpha = 0.2$) & Avg Pool & $8 \times 8 \times 256$ \\
    ResBlock & $3 \times 3$ & LeakyReLU ($\alpha = 0.2$) & Avg Pool & $4 \times 4 \times 256$ \\
    
    Conv & $4 \times 4$ & LeakyReLU ($\alpha = 0.2$) & - & $1 \times 1 \times 256$ \\
    Reshape & - & - & - & 256 \\
    FC & - & - & - & $D$ \\
	\bottomrule
\end{tabular}
\end{table*}

\end{document}